\documentclass[10pt,journal,compsoc]{IEEEtran}
\usepackage{amsmath,amsfonts}
\usepackage{algorithmic}
\usepackage{algorithm}
\usepackage{array}
\usepackage[caption=false,font=normalsize,labelfont=sf,textfont=sf]{subfig}
\usepackage{textcomp}
\usepackage{stfloats}
\usepackage{url}
\usepackage{verbatim}
\usepackage{graphicx}
\usepackage{booktabs}
\usepackage{multicol}
\usepackage{multirow}
\usepackage{color}
\usepackage{xcolor}
\usepackage{colortbl}
\usepackage{hyperref}
\usepackage{balance}
\definecolor{C1}{HTML}{93BFCF}
\definecolor{C2}{HTML}{A0C3D2}
\definecolor{C3}{HTML}{BDCDD6}
\definecolor{C4}{HTML}{EEE9DA}
\definecolor{C5}{HTML}{FFF1DC}
\definecolor{C6}{HTML}{E8D5C4}
\definecolor{C7}{HTML}{EEEEEE}

\ifCLASSOPTIONcompsoc
  \usepackage[nocompress]{cite}
\else
  \usepackage{cite}
\fi
\ifCLASSINFOpdf
\else
\fi
\begin{document}
\title{Self-supervised Learning for Pre-Training 3D Point Clouds: A Survey}

% \IEEEmembership{Graduate Student Member, IEEE}, 
\author{Ben Fei, Weidong Yang, Liwen Liu, Tianyue Luo, Rui Zhang, Yixuan Li, and Ying He% <-this % stops a space
\IEEEcompsocitemizethanks{\IEEEcompsocthanksitem 
\textit{
Corresponding authors: Weidong Yang and Ying He}
\IEEEcompsocthanksitem Ben Fei, Liwen Liu, Tianyue Luo, Rui Zhang, and Weidong Yang are with the School of Computer Science, Fudan University, Shanghai, China, 200433 (e-mails: $\{\text{bfei21}|\text{liwenliu21}|\text{tianyueluo21}|\text{22210240379}\}$@m.fudan.edu.cn;  wdyang@fudan.edu.cn).
\IEEEcompsocthanksitem Yixuan Li is with the Department of Applied Mathematics, the Hong Kong Polytechnic University, Hong Kong SAR, 100872 (e-mail: 22056286g@connect.polyu.hk).
\IEEEcompsocthanksitem Ying He is with the School of Computer Science and Engineering, Nanyang Technological University, Singapore, 639798 (email: yhe@ntu.edu.sg).
}% <-this % stops an unwanted space
% \thanks{Manuscript received April 19, 2005; revised August 26, 2015.}
}

\IEEEtitleabstractindextext{%
% !TEX root = ../bare_jrnl_new_sample4.tex
\begin{abstract}
%Point cloud data has been extensively studied for its excellent accuracy and robustness under a variety of adverse conditions. 
Point cloud data has been extensively studied due to its compact form and flexibility in representing complex 3D structures. The ability of point cloud data to accurately capture and represent intricate 3D geometry makes it an ideal choice for a wide range of applications, including computer vision, robotics, and autonomous driving, all of which require an understanding of the underlying spatial structures. 
% Simultaneously, deep neural networks (DNNs) have demonstrated impressive success in a range of 2D tasks. The convergence of point clouds and DNNs has led to the development of numerous deep point cloud models, which are primarily trained using large-scale, densely labeled point cloud data. 
Given the challenges associated with annotating large-scale point clouds, self-supervised point cloud representation learning has attracted increasing attention in recent years. This approach aims to learn generic and useful point cloud representations from unlabeled data, circumventing the need for extensive manual annotations. In this paper, we present a comprehensive survey of self-supervised point cloud representation learning using DNNs. We begin by presenting the motivation and general trends in recent research. We then briefly introduce the commonly used datasets and evaluation metrics. Following that, we delve into an extensive exploration of self-supervised point cloud representation learning methods based on these techniques. Finally, we share our thoughts on some of the challenges and potential issues that future research in self-supervised learning for pre-training 3D point clouds may encounter.
\end{abstract}

}

\maketitle

\IEEEdisplaynontitleabstractindextext

% \fi
\IEEEpeerreviewmaketitle

% !TEX root = ../bare_jrnl_new_sample4.tex
\section{Introduction}
\label{sec:introduction}
3D point clouds are compact and flexible representations, which offer rich geometric, shape, and scale information. With the rapid advancement of 3D acquisition technology, 3D sensors for capturing point clouds have become increasingly accessible, encompassing various types of 3D scanners, LiDAR, and RGB-D cameras~\cite{xiao2022unsupervised, cui2021deep}. When combined with images, these 3D point cloud data can help machines perceive their surroundings, making them widely used in numerous scene-understanding-related applications such as computer vision, robotics, autonomous driving, remote sensing, and medical treatment~\cite{guo2020deep}.

\begin{figure}
    \centering
    \includegraphics[width=0.9\linewidth]{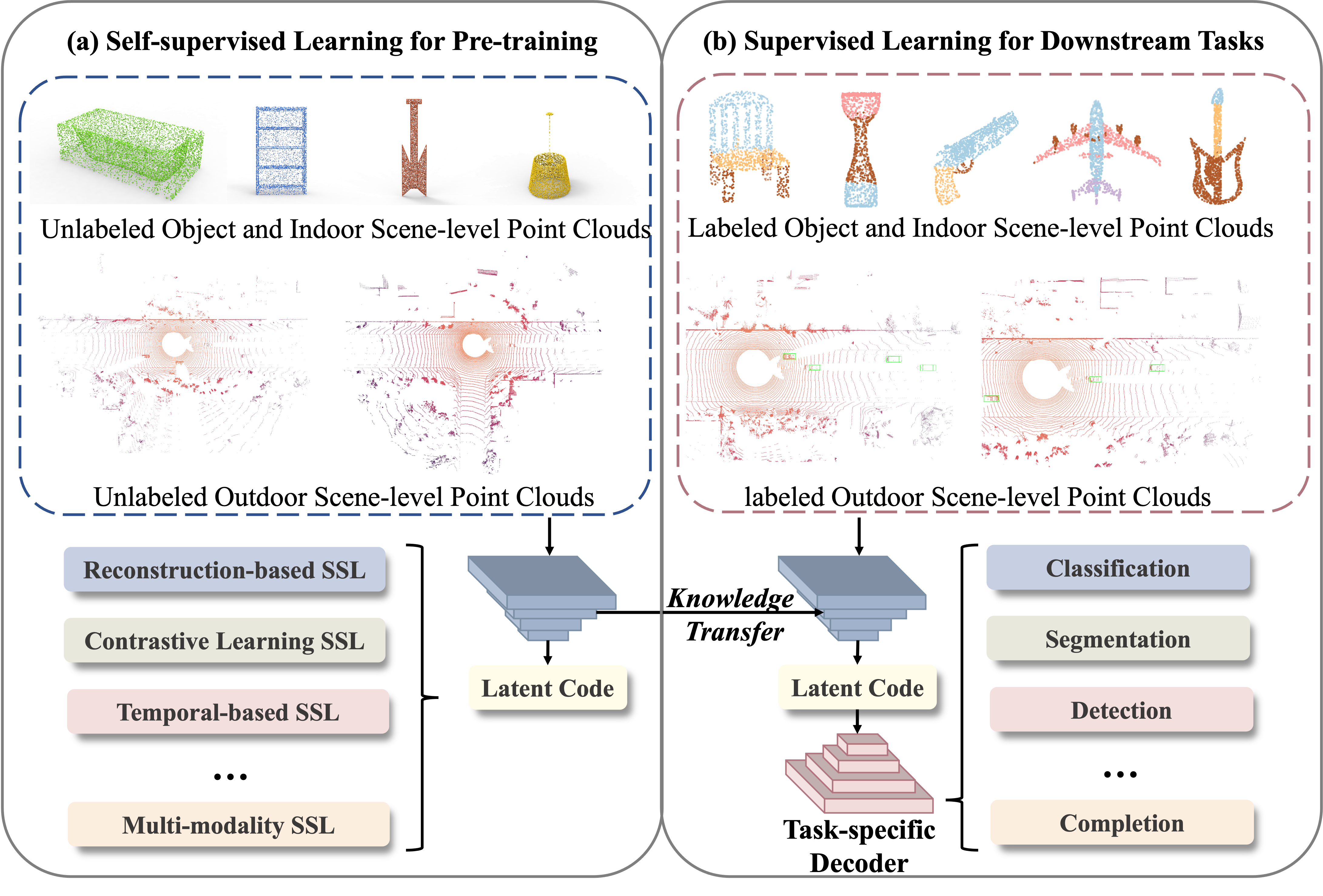}
    \vspace{-0.5cm}
    \caption{The pipeline for \textbf{self-supervised pre-training of 3D point clouds} begins with unlabeled point clouds, followed by pre-training of deep neural networks using self-supervised learning on pretext tasks. The learned point cloud representations are subsequently utilized in various downstream tasks to initialize the network. Finally, the pre-trained networks are fine-tuned using a small amount of labeled task-specific point cloud data to achieve high performance.}
    \label{fig:teaser}
\vspace{-0.3cm}
\end{figure}

As deep neural networks (DNNs) continue to advance, point cloud understanding has gained increasing attention, leading to the development of numerous deep architectures and models in recent years. However, effective training of deep networks typically requires large-scale, human-annotated training data, such as 3D bounding boxes for object detection and pointwise annotations for semantic segmentation. Collecting these annotations can be laborious and time-consuming due to factors such as occlusion, shape variations, and visual inconsistencies between human perception and point cloud display. Consequently, the efficient collection of large-scale annotated point clouds has become a bottleneck in the effective design, evaluation, and deployment of DNNs for various practical tasks. 

To circumvent the time-consuming and expensive data labeling process, numerous self-supervised methods have been proposed to learn visual features from large-scale unlabeled point clouds without relying on any human-generated labels. A popular approach involves designing various pretext tasks for the network to solve. The network can be trained by optimizing the objective function of the pretext tasks and learning features through this process. Various pretext tasks have been proposed for self-supervised learning, including point cloud reconstruction, contrastive learning, and multi-modal learning, among others. 
Pretext tasks share two common properties: (1) The visual features of point clouds must be captured by DNNs to solve the pretext task; and (2) the supervisory signal is generated from the data itself (self-supervision) by exploiting its structure. 

To foster methodological advancements and enable a comprehensive comparison, we review self-supervised learning (SSL) methods for 3D pre-training and provide a unified perspective on comparison and prediction techniques. Our consolidated approach to this problem highlights the differences and similarities among existing methods, potentially inspiring novel solutions. We summarize the contributions of this survey as follows: 
 
\begin{itemize}
    \item \textbf{Unified framework and systematic taxonomy.} We propose a unified framework, based on which, we systematically categorize existing works into two main groups: object and indoor level, and outdoor level. Furthermore, we construct taxonomies of downstream tasks and SSL learning schemes to provide a comprehensive understanding of this field.
    \item \textbf{Comprehensive and up-to-date review.} We provide a comprehensive and timely survey of both classical and cutting-edge 3D pre-training SSL methods. For each type of approach, we offer fine-grained classification, in-depth comparison, and summaries. To the best of our knowledge, our survey presents the first review of SSL specifically focused on pre-training 3D point cloud data.
    % \item \textbf{Abundant resources and applications.} \textcolor{red}{We outline common settings of SSL tasks on 3D pre-training and commonly used datasets of various categories under different settings, laying the stage for the development of future methods.}
    \item \textbf{Outlook on future directions.} We highlight the technical limitations of current research and propose several promising avenues for future work, offering insights from various perspectives to inspire further advancement in this field.
\end{itemize}

\begin{figure*}[htbp]
    \centering
    \includegraphics[width=0.9\linewidth]{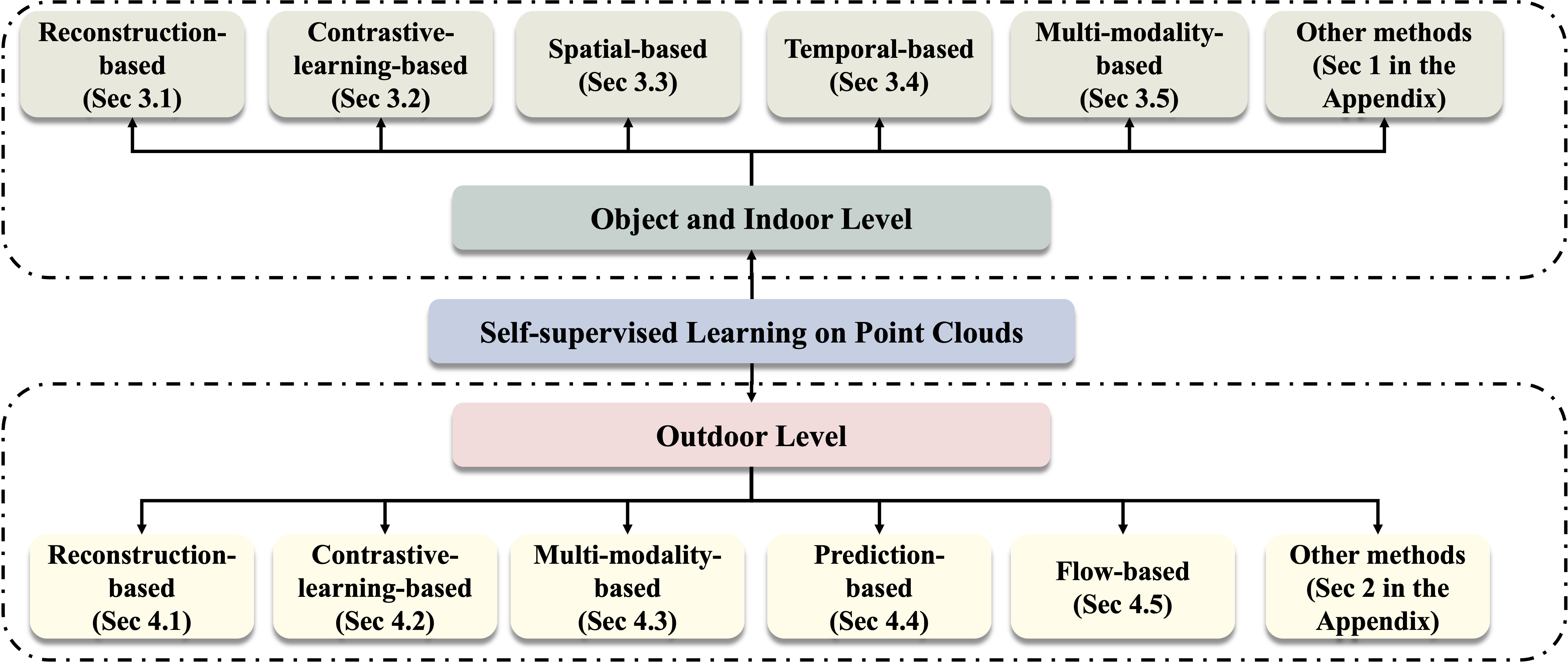}
    \vspace{-0.4cm}
    \caption{Taxonomy of the recent and most relevant SSL-based methods for point cloud pre-training.}
    \label{fig:tax}
\vspace{-0.3cm}
\end{figure*}
% 

% \textcolor{red}{The remaining of the survey is organized as follows: Section~\ref{}...}
The structure of this survey is organized as follows: 
Section~\ref{background} introduces the background knowledge of self-supervised learning for pre-training point clouds, the commonly used datasets, and their characteristics. Section~\ref{object-level} presents a systematic review of the SSL methods for pre-training point clouds at the object and indoor-scene levels, while Section~\ref{outdoor-level} compares and summarizes the methods for the outdoor scene-level data. Finally, Section~\ref{future_work} identifies several promising future directions for self-supervised point cloud pre-training.
% \textbf{Section III (Discussion and Future works)} in the \textbf{Supplementary Material} 

% In Section~\ref{background}, we introduce background knowledge of self-supervised point cloud learning, widely-used datasets, and their characteristics.  
% In Section~\ref{object-level}, we systematically review the SSL methods for point clouds from the object and indoor levels. 
% Section~\ref{outdoor-level} summarizes and compares the SSL methods for point clouds from the outdoor scene level. 
% At last, we figure out several promising future directions for self-supervised point cloud pre-training in Section~\ref{future_work}.

% !TEX root = ../bare_jrnl_new_sample4.tex
% \usepackage{graphicx}
\begin{table*}[htbp]
\centering
\caption{Summary of commonly used datasets for training and evaluations in self-supervised point cloud pre-training studies.}
\vspace{-0.3cm}
\resizebox{\textwidth}{!}{%
\begin{tabular}{>{\columncolor{C7!50}}l>{\columncolor{C6!50}}c>{\columncolor{C5!50}}c>{\columncolor{C4!50}}c>{\columncolor{C3!50}}c>{\columncolor{C2!50}}c>{\columncolor{C1!50}}c}
\toprule[1.5pt]
Dataset        & Year & Samples        & Classes & Type              & Representation       & Label                                \\ \midrule[1pt]
ShapeNet~\cite{chang2015shapenet}      & 2015 & 51,190 objects & 55      & Synthetic object  & Mesh \& LiDAR        & Object/part category label           \\
ShapeNetRender~\cite{afham2022crosspoint} & 2022 & more than 50,000 objects              & 55       & Synthetic object  & RGB \& Mesh \& LiDAR & Object category label                \\
ModelNet40~\cite{wu20153d}     & 2015 & 12,311 objects & 40      & Synthetic object  & Mesh                 & Object category label                \\
ScanObjectNN~\cite{uy2019revisiting}   & 2019 & 2,902 objects  & 15      & Real-world object & Points               & Object category label                \\
ScanNet~\cite{dai2017scannet}        & 2017 & 1,513 scans    & 20      & Indoor scene      & RGB-D \& mesh        & Point category label \& Bounding box \\
SUN RGB-D~\cite{song2015sun}      & 2015 & 5K frames      & 37      & Indoor scene      & RGB-D                & Bounding box                         \\
S3DIS~\cite{armeni20163d}          & 2016 & 272 scans      & 13      & Indoor scene      & RGB-D                & Point category label                 \\
KITTI~\cite{geiger2013vision}          & 2013 & 15K frames     & 8       & Outdoor scene     & RGB \& LiDAR         & Bounding box                         \\
SemanticKITTI~\cite{behley2019semantickitti}  & 2019 & 45K frames     & 28      & Outdoor scene     & LiDAR                & Point category label                 \\
SemanticPOSS~\cite{pan2020semanticposs}   & 2020 & 2K frames      & 14      & Outdoor scene     & LiDAR                & Point category label                 \\
Waymo~\cite{sun2020scalability}          & 2020 & 15K frames     & 23      & Outdoor scene     & LiDAR                & Point category label \& Bounding box \\
nuScene~\cite{caesar2020nuscenes}        & 2019 & 40K frames     & 31      & Outdoor scene     & RGB \& LiDAR         & Bounding box                         \\
ONCE~\cite{mao2021one}           & 2021 & 1M scenes      & 5       & Outdoor scene     & RGB \& LiDAR         & Bounding box    
\\ \bottomrule[1.5pt]
\end{tabular}%
}\label{dataset}
\vspace{-0.3cm}
\end{table*}

% !TEX root = ../bare_jrnl_new_sample4.tex
\vspace{-0.3cm}
\section{Background}
\label{background}
We introduce the relevant terms and concepts in the following sections.
\vspace{-0.3cm}

\subsection{Basic concepts}
\vspace{-0.1cm}

\textbf{3D Point clouds.} A point cloud $\boldsymbol{P}$ is a collection of 3D vectors $\boldsymbol{P} = \{p_1, p_2, \dots, p_L\}$, where each vector can be regarded as a point $p_i = [\boldsymbol{C}_i, \boldsymbol{F}_i]$. $\boldsymbol{C}_i \in \mathbb{R}^{1\times3}$ denotes the 3D coordinates $\left(x_i, y_i, z_i\right)$ of the point, while $\boldsymbol{F}_i$ represents the feature attributes of the point, including RGB values, intensity, normal vector, etc. These attributes are optional and vary depending on the 3D sensors as well as application requirements.

\textbf{Self-supervised learning} is a type of unsupervised learning where the supervision signals are generated from the data itself. In self-supervised learning methods, models are trained on pretext tasks that do not require human-labeled data, enabling them to learn representations that can generalize to downstream tasks.

\textbf{Pre-training} is a commonly-used strategy in deep learning where a model is trained on a large dataset to learn general features or representations, which can then be utilized as a starting point for training on task-specific data. 

\textbf{Transfer learning} refers to the process of transferring knowledge and insights gained from one task, domain or dataset to another. In the context of this survey, transfer learning occurs through pre-training of self-supervised learning, where knowledge is transferred from unlabelled data to various downstream networks. 

\vspace{-0.3cm}
\subsection{Datasets} 
\vspace{-0.1cm}

Various publicly available datasets are utilized to evaluate the performance of pre-trained networks on various downstream tasks. Table~\ref{dataset} provides an overview of some of these datasets for 3D shape classification, object detection and tracking, and segmentation. These datasets have different properties, which are summarized in the table and discussed below. 

\textbf{3D shape classification.} Both synthetic data and real-world data~\cite{uy2019revisiting, dai2017scannet} are commonly used for 3D shape classification tasks. Synthetic datasets, such as ModelNet40~\cite{wu20153d}, and ShapeNet~\cite{chang2015shapenet}, typically consist of complete objects without occlusion or background noise. These datasets are useful for studying the impact of object shape and geometry on classification performance. Real-world datasets, such as ScanObjectNN~\cite{uy2019revisiting}, contain objects with varying degrees of occlusion and background noise. These datasets are more challenging than synthetic datasets and reflect the conditions of real-world applications.

\textbf{Object detection and tracking.} There are two types of datasets frequently used for object detection and tracking: indoor scenes~\cite{dai2017scannet, song2015sun} and outdoor urban scenes~\cite{geiger2012we, kesten2019lyft, sun2020scalability, caesar2020nuscenes, mao2021one}. Indoor scene-level datasets typically consist of point clouds transformed from dense depth maps or sampled from 3D meshes. In contrast, outdoor scene-level datasets are sparser and designed for autonomous driving, with objects that are well-separated spatially.

\textbf{Semantic segmentation.} Two widely used datasets for evaluating the performance of pre-trained networks on semantic segmentation are SemanticKITTI~\cite{geiger2012we,behley2019semantickitti, behley2021towards} and SemanticPOSS~\cite{nekrasov2021mix3d}. Both datasets were collected in outdoor urban environments.

\vspace{-0.2cm}
\subsection{Evaluation Metrics}
\vspace{-0.1cm}

Various metrics have been proposed to evaluate the performance of typical point cloud tasks, such as understanding, segmentation, detection and reconstruction. These metrics provide a quantitative way to compare different methods and models for point cloud processing. 

Overall accuracy (OA) and mean classification accuracy (mAcc) are widely utilized for evaluating 3D shape classification models. OA calculates the average accuracy across all test instances, measuring the proportion of correctly classified shapes to the total number of shapes in the test dataset. It is useful for determining the general performance of a model across all instances without considering class imbalances. In contrast, mAcc, which is the average accuracy over all shape categories, takes into account class imbalances, providing a more comprehensive evaluation of the model's performance in classifying various shape categories. 
%mACC is calculated by first computing the accuracy for each individual category, and then averaging these accuracies across all categories. 

Average precision (AP) is the standard evaluation metric in 3D object detection. It is calculated as the area under the precision-recall curve. Precision measures the proportion of true positive predictions among all positive predictions, while recall captures the proportion of true positive predictions among all actual positive instances. AP takes both false positives and false negatives into account to provide a comprehensive assessment of a model's performance.

In 3D point cloud segmentation, OA, mACC and mean Intersection over Union (mIoU) are the most commonly used performance evaluation metrics. IoU measures the overlap between the predicted and ground-truth segmentation masks by dividing the intersection of the predicted and ground-truth regions by their union. mIoU, which is the average IoU across all classes, accounts for both false positives and false negatives of each class and provides a single value to evaluate the overall performance of a segmentation model.

Mean Average Precision (mAP) is typically used for 3D point cloud instance segmentation evaluation. It is calculated as the average of the maximum precision values for different recall levels, representing the average precision across various object categories. The recall level indicates the percentage of the total number of objects that are detected correctly. Since mAP considers precision and recall simultaneously, it provides a single value for evaluating the overall performance of an instance segmentation model.

Lastly, Chamfer distance (CD)~\cite{fan2017point} and earth mover’s distance (EMD)~\cite{fan2017point} are the most frequently used criteria in 3D reconstruction. CD calculates the minimum distance between each point in one set and the other set, while EMD measures the minimum cost of transforming one point cloud distribution into another.

% !TEX root = ../bare_jrnl_new_sample4.tex
\vspace{-0.2cm}
\section{Object and Indoor Scene-level SSL}
\label{object-level}

Object-level SSL methods mainly focus on pre-training models using individual 3D objects, such as chairs, tables, cars, etc, which are typically associated with semantic labels to provide contextual information about their identity. 
This type of data is commonly used for tasks such as object recognition, detection, and segmentation, aiming at identifying and localizing individual objects within a larger scene.

Indoor scene-level SSL methods, on the other hand, concentrate on pre-training models using entire 3D indoor environments, often containing multiple objects and their spatial arrangements. 
Indoor scene-level data are often associated with semantic labels for objects and architectural elements, such as ``wall'', ``door'', ``window'', and ``floor'', or categorized by functional labels, such as ``kitchen'', ``bedroom'', ``living room'' or ``office''. 
In contrast to object-level SSL methods, indoor scene-level SSL methods often require the input data to be pre-segmented into semantic regions or objects to provide contextual information about the scene. 

\begin{figure}
    \centering
    \includegraphics[width=0.9\linewidth]{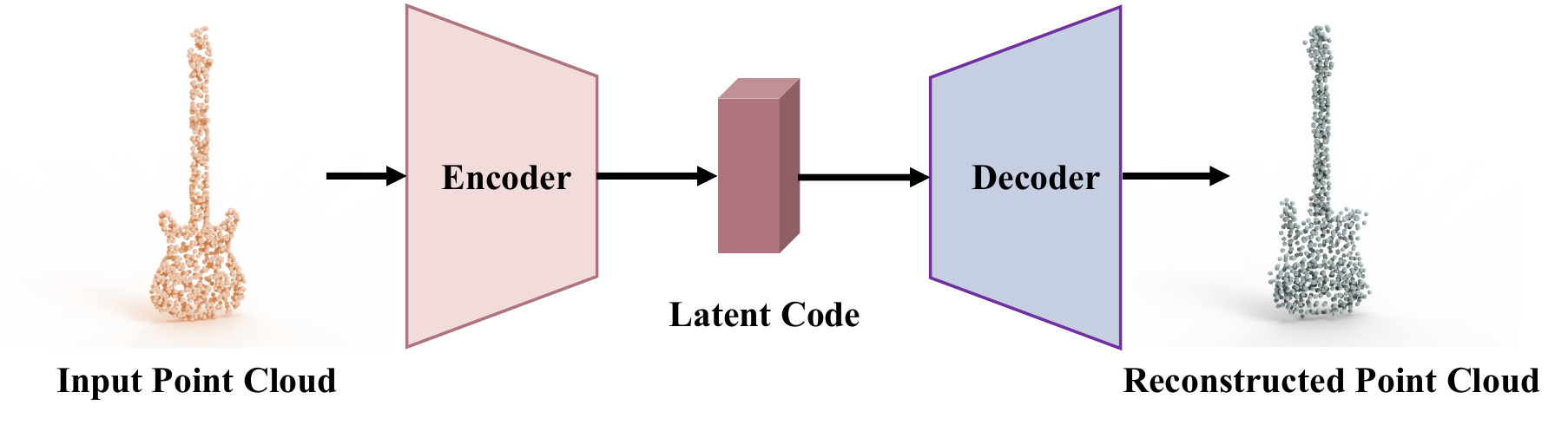}
    \vspace{-0.5cm}
    \caption{Reconstruction-based SSL adopts an encoder-decoder architecture. The Encoder learns to represent a point cloud object by a latent code vector, while the Decoder reconstructs the output object from the latent code. The input point cloud can be either masked or corrupted, leading to two major groups of methods.}
    \label{fig:recontruction}
\vspace{-0.3cm}
\end{figure}

\vspace{-0.3cm}
\subsection{Reconstruction-based SSL}
\vspace{-0.1cm}

Reconstruction-based self-supervised learning methods employ a reconstruction task to enable the network to learn better 3D point cloud representations (see Fig.~\ref{fig:recontruction}). 
They can be broadly classified into two major subgroups depending on the nature of the pretext task: Mask-based and corruption-based. There are also a few methods that do not fit into the two groups, and they are categorized as ``other'' methods. 
%In the following sections, we examine each of these groups in detail.

\vspace{-0.2cm}
\subsubsection{Mask-based Methods}
\vspace{-0.1cm}
Mask-based self-supervised learning methods involve using masks to generate a masked point cloud dataset by randomly sampling different camera viewpoints and masking all the occluded points in each viewpoint. 

Wang et al. proposed the self-supervised pre-training method OcCo~\cite{wang2021unsupervised}, which trains an encode-decoder architecture to reconstruct the complete point cloud from the masked inputs. 
The weights learned by the encoder are utilized as the model initialization for downstream tasks. 
Zhang et al.\cite{zhang2022masked} presented MaskSurf, a self-supervised pre-training method that explicitly considers the local geometrical information of point clouds. 
It adopts a simplified surfel representation (i.e., position and orientation) to enhance point cloud representation, and utilizes the Chamfer distance (CD) and position-indexed normal distance (PIND) as the reconstruction loss of position and orientation to predict the masked surfel in a set-to-set manner. 

% \begin{figure}[ht]
%     \centering
%     \includegraphics[width=\linewidth]{Figures/reconstruction-based-SSL/Point-BERT.pdf}
%     \caption{Illustration of Point-BERT~\cite{yu2022point}. Point-BERT designs a mask point modeling (MPM) task for 3D point clouds by extending the mask language modeling (MLM)~\cite{liu2021self} strategy in BERT to 3D point cloud transformers. Image courtesy of Yu et al.}
%     \label{fig:Point-BERT}
% \end{figure}

Inspired by the bidirectional encoder representations from transformers (BERT)~\cite{tenney2019bert} in natural language processing, Yu et al.\cite{yu2022point} introduced a variant called Point-BERT for 3D point clouds. This method extends the mask language modeling (MLM) strategy in BERT to 3D point cloud transformers by treating the input point cloud and its discrete coding sets as words and sentences in a language. The goal is to design a mask point modeling (MPM) strategy for 3D point clouds based on the MLM strategy in BERT. 
% See  Fig \ref{fig:Point-BERT}. 

Before pre-training, Point-BERT divides the input point cloud into multiple local patches and generates discrete point tokens for each patch using a dVAE tokenizer. 
During pre-training, the method uses a Transformer to predict the masked inputs of randomly masked patches and compares its prediction result with the discrete point tokens of the prediction target to capture advanced semantic knowledge and learn geometric relationships among different patches. The masked tokens are fed into the encoder containing their position information, thereby reducing the difficulty of the reconstruction task.
As the first of its kind, Point-BERT has demonstrated the potential of adopting the masking strategy with BERT in point cloud pre-training, achieving
state-of-the-art performance on various downstream tasks such as point cloud classification, few-shot classification, and part segmentation.
However, it has several limitations: 
Firstly, the use of dVAE tokenizer focuses on the geometric structure of point clouds but ignores the relationship between similar local point patches. Secondly, the entire pre-training process is relatively complex and time-consuming as it requires the preparation of the pre-training dVAE, and heavily relies on data augmentation and contrastive learning. 

Point-BERT encodes local point patches in a point cloud by assigning a unique token ID to each patch. However, when different local patches have similar or identical features, they can be assigned the same token ID, leading to a loss of information and decreased performance. To address this token-ambiguity issue, Fu et al.\cite{fu2022point} proposed the McP-BERT model, which uses improved multi-choice tokens for each local point patch based on probability distribution vectors. This approach avoids the problem of semantically-different patches having the same token IDs caused by strict single-choice constraints. Additionally, the probability distribution vectors are further refined by incorporating high-level semantic relationships learned by the Transformer, which effectively overcomes the problem of semantically-similar patches having different token IDs due to noise interference. 

The Point-BERT model has complex training steps that can be time-consuming and difficult to optimize. To address this issue, Fu et al. proposed POS-BERT, which is a single-stage pre-training method. Replacing the weight-frozen tokenizer used in Point-BERT with a dynamically updated momentum encoder, POS-BERT enables adaptive changes during network training. By using a single-stage pre-training approach, POS-BERT eliminates the need for additional fine-tuning steps, reducing the overall training cost.
In addition, POS-BERT introduces contrastive learning on class tokens between global point clouds and local point clouds obtained by different cropping ratios, which maximizes the class token consistency among point cloud pairs, thereby better learning advanced semantic representation.  

ContrastMPCT~\cite{wang2022self} and MAE3D~\cite{jiang2022masked} are alternative methods to reduce the complexity of Point-BERT by eliminating the additional tokenizer training stage. ContrastMPCT adopts a self-supervised strategy based on contrastive learning and masked autoencoders. It measures the similarity between the predicted and ground truth point clouds using the Chamfer distance and designs two joint loss functions based on JSD and InfoNCE to maximize the global dependence between the input and output tokens for faster model convergence. In contrast, MAE3D directly utilizes Transformers to learn the geometric features of local patches and the contextual relationships among them. It also designs a multi-task reconstruction loss considering both the center point of predicted local patches and point cloud reconstruction for the folding operation.

% \begin{figure}[ht]
%     \centering
%     \includegraphics[width=\linewidth]{Figures/reconstruction-based-SSL/Point-MAE.pdf}
%     \caption{Point-MAE~\cite{pang2022masked} consists of three main parts: a point cloud masking module, an embedding module, and an auto-encoder module. Image courtesy of Pang et al.}
%     \label{fig:Point-MAE}
% \end{figure}

Inspired by the success of the masked auto-encoder (MAE) in 2D computer vision, Pang et al.~\cite{pang2022masked} proposed Point-MAE for 3D point clouds. Their method reduces the model complexity and fixes location information leaking of Point-BERT. 
% As shown in Fig \ref{fig:Point-MAE}, 
This method consists of three main parts: a point cloud masking module, an embedding module, and an auto-encoder module. The point cloud masking module randomly masks divided point clouds to reduce data redundancy caused by an uneven distribution of point clouds. The auto-encoder module uses an asymmetric encoder-decoder structure constructed entirely by standard Transformers, without introducing other networks besides Transformers. Unlike Point-BERT, the method selects a lightweight decoder to process masked tokens, rather than an encoder. The operation of shifting masked tokens to the input of the decoder not only allows the encoder to better learn high-level semantic features of the point cloud, but also increases pre-training efficiency by reducing the complexity of the model and avoiding leaking early location information of masked tokens. 
 
Despite of its simplicity and high efficiency, Point-MAE can only be applied to point clouds of a single resolution, without considering the relationship between its local structure and global shape. To overcome this limitation, Zhang et al. proposed Point-M2AE~\cite{zhang2022point}, which uses a U-Net-like transformer architecture with a multi-scale hierarchical structure to progressively encode and reconstruct the point cloud. This allows the network to learn the multi-scale geometric structure and fine-grained information of 3D shapes.
Before down-sampling the point tokens, Point-M2AE adopts a multi-scale masked strategy by generating consistently visible regions across scales, enabling the network to coherently learn features and avoid information leakage. Moreover, by using skip connections between the encoder and decoder stages, Point-M2AE can supplement the fine-grained information of the encoder in the corresponding stage when up-sampling the point tokens promote local-to-global reconstruction, which helps capture the relationship between the local structure and global shape of the point cloud.

\vspace{-0.2cm}
\subsubsection{Corruption-based Methods} 
\vspace{-0.1cm}
In addition to the mask-based methods, several approaches adopt a corruption-based pretext task, in which point clouds are intentionally corrupted and then recovered, to pre-train the model. 

Xu et al.~\cite{xu2022cp} proposed CP-Net, which utilizes a weight-sharing dual-branch structure to effectively guide self-supervised learning of both structural contour and semantic content of point cloud representations. The assistant branch of the dual-branch structure adds a contour-perturbed augmentation module, which forces it to focus on distinguishing the semantic content of downstream tasks by disturbing the point cloud contour while retaining its content. The other branch learns the ignored high-level semantic content information from the assistant branch, improving the discriminative abilities of point cloud representations by introducing the dual-branch consistency loss. 

Shape self-correction \cite{chen2021shape} employs a shape-disorganizing strategy to destroy certain local shape parts of an object. The corrupted shape and the original shape are then fed into a point cloud network to obtain representations, used to segment points belonging to distorted parts and recover them to restore the shape. 

%Shape self-correction has demonstrated excellent performance in both shape classification and segmentation tasks. 

Zhang et al.\cite{zhang2022point1} proposed Point-DAE, a universal denoising auto-encoder, which was explored under 14 types of corruptions as pretext tasks, including density, noise, and affine transformations. Their findings suggested a linear relationship between task dependencies and performance, with Point-DAE performing best under the affine transformation pretext task, which is more relevant to the classification task. They also introduced a new dataset setting that allows for automatic estimation of the canonical pose, thereby eliminating the implicit class label brought by manually aligned canonical poses of the same category in the pre-training dataset. 

Garg and Chaudhary \cite{garg2022serp} proposed SeRP, which employs point cloud perturbation as a pretext task. The method involves randomly selecting 20 points from an input point cloud of 1024 points and using the nearest neighbors to form a patch for each selected point. Gaussian noise is then applied to correct each patch. SeRP adopts an auto-encoder based on PointNet~\cite{qi2017pointnet} and Transformers to reconstruct the original point cloud. Additionally, 
% to extend SeRP to \textcolor{red}{discrete datasets},\fb{It is described as discrete datasets in the original paper, but I think it is not suitable to use "discrete datasets". May be we can remove it.}, 
an auto-encoder is built to realize vector quantization of discrete representations, extending the SeRP-Transformer.

% \begin{figure}[ht]
%     \centering
%     \includegraphics[width=\linewidth]{Figures/reconstruction-based-SSL/Self-supervised-Sauder.pdf}
%     \caption{Visual example of the 3D jigsaw-based SSL learning method~\cite{sauder2019self}. (a) The input point cloud is uniformly divided into cells along the coordinate axes. (b) The cells are randomly rearranged. (c) The neural network prediction. (d) Points with predicted correct labels (blue) and incorrect labels (red). Images courtesy of Sander and Sievers~\cite{sauder2019self}. \textcolor{red}{given the jigsaw method is not a mainstream method, we may not need to show a figure for it due to page limit} }
%     \label{fig:Self-supervised-Sauder}
% \end{figure}

\vspace{-0.2cm}
\subsubsection{Other Methods} 
\vspace{-0.1cm}
This subsection discusses a few methods that do not belong to either reconstruction- or corruption-based methods.

The 3D jigsaw-based self-supervised learning method, proposed by Sauder and Sievers~\cite{sauder2019self}, involves uniformly dividing the input point cloud into 
$k^3$ voxels along the coordinate axes and 
labeling each point according to the corresponding voxel ID. The voxels are rearranged in random order. The network is then trained to predict the correct label assignment of each point. 
% See Fig.~ \ref{fig:Self-supervised-Sauder}.
This approach not only improves the reconstruction capability of the autoencoder but also has the flexibility to be applied to almost any deep learning models pre-trained with the original point cloud. 
While this approach has shown satisfactory results on ScanObjectNN~\cite{wu20153d} and S3DIS~\cite{armeni20163d}, it may not be able to handle diverse downstream tasks and process large-scale point clouds. In addition, there are discrepancies between randomly arranged synthetic point clouds and realistic generated point clouds, which can result in poorly initialized pre-training weights for downstream tasks.

Achituve et al.\cite{achituve2021self} investigated domain adaptation of SSL on point clouds and introduced a self-supervised task called DefRec, which includes three types of region selection methods. The task involves replacing region points with new points sampled by a Gaussian distribution to achieve deformation, and then training a shared feature encoder to reconstruct the deformed input samples. 
To train labeled samples over the source domain, DefRec employs point cloud mixup (PCM) combined with the MixUp method to replace the standard cross-entropy classification loss, leading to an improvement in classification performance on the target domain. 
% \textcolor{red}{However, it does not explicitly focus on the difference in LiDAR sensor placement.}

Alwala et al.~\cite{alwala2022pre} introduced a multi-stage training method for learning a unified reconstruction model across different object categories, enabling the reconstruction of 3D objects from a single view across hundreds of categories.
In the first stage of training, it uses multi-view renderings of synthetic data to pre-train a basic reconstruction model, helping the model learn correct 3D priors in a weakly-supervised manner. In the second stage of training, this method self-trains category-specific models from various single-view image sets with only foreground mask annotations, obtaining diverse category sets through fine-tuning the initial model. Finally, the adaptive model of each category in the previous training stage is extracted into a unified reconstruction network, leading to a joint model with better generalization. Although this method is effective in reconstructing the global structure of objects, it has limitations in capturing the fine geometric details of those objects. 

Eckart et al.\cite{eckart2021self} developed an SSL method that is adaptable for any DNN architecture producing point-wise classification scores. This method softly segments 3D points into a set of a discrete number of geometric partitions and implicitly parameterizes the latent Gaussian model in these soft partitions. By maximizing the data likelihood associated with the soft partitions generated by the unsupervised point-wise segmentation network, this method promotes learning representations rich in geometric information.

Zhang et al.\cite{zhang2022upsampling} proposed an up-sampling auto-encoder (UAE) that does not  require processing or retrieving negative samples, nor does it depend on any data enhancement techniques. 
UAE takes a low proportion of sub-sampled points as the input to the decoder, and directly reconstructs up-sampled points from the point space, thereby providing a simple and effective up-sampling model that captures high-level semantic information.

Yan et al.~\cite{yan2022implicit} proposed the implicit auto-encoder (IAE), which uses an implicit function as the output surface representation. Compared to the point cloud autoencoder, IAE effectively addresses the sampling variation problem and offers a compact and computationally efficient solution. As a result, it has the potential for handling large-scale real-world point clouds.

\vspace{-0.2cm}
\subsubsection{Challenges and Opportunities}
\vspace{-0.1cm}
Reconstruction-based self-supervised learning methods have emerged as a promising research direction for point cloud pre-training. 
These methods utilize various techniques to learn feature representations from synthetic, object-level point clouds through different generative tasks. 
However, generating scene-level point clouds that provide richer data distributions and broader application potential is a challenging task because of the large number of points, severe occlusions, and complex structures in 3D scenes.
As a result, there has been little research on learning 3D representations from generated scene-level data. 
Nevertheless, this area offers great opportunities for further exploration and development in this area.

\vspace{-0.3cm}
\subsection{Contrastive-learning-based SSL}
\vspace{-0.1cm}
\label{CL-based-methods}
Contrastive learning is a popular self-supervised learning method. It involves constructing positive and negative samples by an auxiliary task and training the model to bring positive sample pairs closer in the embedding space while separating positive samples from negative ones. In contrast to generative methods, contrastive learning does not rely on the details of specific samples, but rather on discriminating positive and negative samples in the embedding space. This feature makes the models easier to optimize and more generalizable. 

One common solution for training the model to distinguish between positive and negative sample pairs is to convert the contrastive learning into a multi-classification problem. This idea was first introduced by Oord et al.~\cite{oord2018representation}, who proposed the InfoNCE loss, which was originally used in 2D image processing. Specifically, given an anchor data $x$, a positive sample $x^+$, and a set of negative samples $\{x^-_0, \ldots, x^-_k\}$, the InfoNCE loss is defined as follows: 
\begin{equation}
\label{eq:infonce_loss}
\mathcal{L}_c = -\log\frac{\exp(x, x^+/\tau)}{\sum_{i=0}^{k}\exp(x, x_i^-/\tau)},
\end{equation}
where $\tau$ is the temperature coefficient that controls the sharpness of the distribution of the similarity scores between the anchor and the positive and negative samples. The InfoNCE loss is similar to the cross-entropy loss~\cite{qi2021self} and the only difference between them is the interpretation of $k$~\cite{he2020momentum}. 
In the cross-entropy loss, $k$ represents the number of categories, while in the InfoNCE loss, $k$ stands for the number of negative samples. InfoNCE loss can be viewed as a ($k$+1)-way classification task whose goal is to assign $x$ to the category where its positive samples $x^+$ belong. 

Recently, extensive research has been performed on contrastive learning methods for images, leading to the development of algorithms for tasks such as image translation~\cite{park2020contrastive,han2021dual}, generation~\cite{kang2020contragan}, segmentation~\cite{chaitanya2020contrastive}, among others. The success of methods like MoCo~\cite{he2020momentum} and SimCLR~\cite{chen2020simple} has demonstrated the potential of contrastive learning for self-supervised representation learning. While applying contrastive learning to 3D point clouds is still a relatively new field, early works~\cite{xie2020pointcontrast, zhang2021self} have explored the use of 2D contrastive learning techniques on transformed point clouds, such as multi-view or depth images. There has been a recent trend toward developing 3D-specific contrastive learning methods that leverage the 3D properties of point clouds to achieve even better performance.

\vspace{-0.2cm}
\subsubsection{View-based Methods}
\begin{figure}[t]
    \centering
    \includegraphics[width=0.9\linewidth]{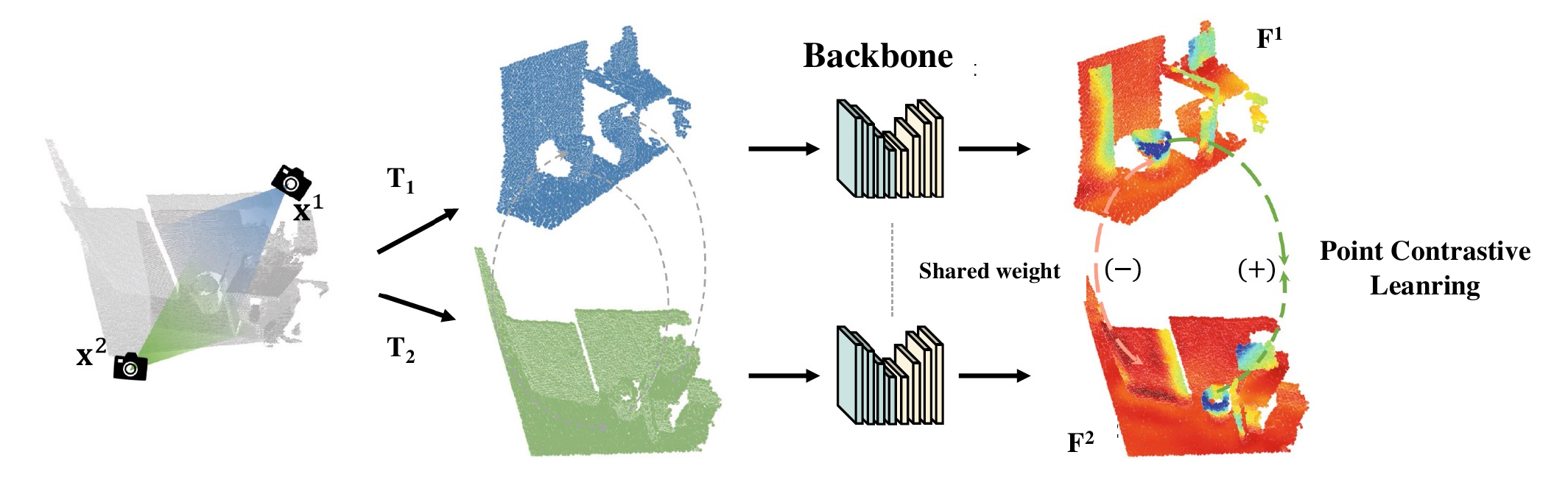}  
    \vspace{-0.5cm}
    \caption{PointContrast~\cite{xie2020pointcontrast}, a notable example of contrastive-learning-based methods, allows the network to learn equivariance with respect to geometric transformations by contrasting points between two transformed views. Image courtesy of Xie et al.~\cite{xie2020pointcontrast}.}
    \label{fig:PointContrast}
\vspace{-0.3cm}
\end{figure}

\vspace{-0.1cm}
Xie et al \cite{xie2020pointcontrast} proposed PointContrast, as shown in Fig.~\ref{fig:PointContrast}, a point-level self-supervised contrastive learning method for multi-view 3D point cloud understanding. PointContrast aims to perform point-level comparisons in two transformed point clouds with different views to capture dense information at the point level. 
There are two loss functions designed for contrastive learning. One is the Hardest-Contrastive loss, which is borrowed from FGCF~\cite {choy2019fully}. The general idea of this loss is to directly minimize the distance between matching point features and maximize the distance between non-matching points. The other is the PointInfoNCE loss, derived from the InfoNCE loss, but focusing on modeling point-level information. Experimental results show that this loss is easier to optimize and more stable than the hardest-contrastive loss~\cite {choy2019fully}.

While PointContrast demonstrates the effectiveness of self-supervised pre-training in a variety of 3D point cloud understanding tasks with context, there is still room for improvement in several areas for improvement. Firstly, 
the method does not utilize spatial contextual information, such as orientation, distance, and relative position, which are critical in many understanding tasks. Second, the scalability of PointContrast is limited by its usage of only 1,024 points for pre-training, meaning that providing more points does not improve performance. Thirdly, PointContrast requires resource-intensive inputs, 
such as the absolute position of the camera, which are not easy to obtain. These issues highlight the need for further research to enhance the scalability, efficiency, and effectiveness of PointContrast.

Zhang et al.\cite{zhang2021self} proposed DepthContrast, which reduces the resource-intensive issue of PointContrast by using only a single-view depth map. 
Employing two feature extractors at the voxel and point levels, respectively, DepthContrast extracts four features from the two augmented inputs and calculates the InfoNCE losses between them in a pairwise fashion, and then aggregates the results. 
DepthConstrast has demonstrated the potential of combining voxel-based and point-based representations to improve 3D point cloud understanding.  

% \begin{itemize}
%     \item \underline{Under-utilization of spatial contextual information.} PointContrast does not utilize spatial contextual information, such as orientation, distance, and relative position, which play a critical role in many understanding tasks. 
% \item \underline{\textcolor{red}{Lack of scalability.}} PointContrast uses only 1,024 points for pre-training, and providing more points does not improve performance. Therefore, there is a pressing need to extend it further to handle large-scale inputs.
% \item \underline{Resource intensive nature.} PointContrast relies on the depth map of the point cloud and the absolute position of the camera to obtain multi-view information, which are both time-consuming and resource-intensive. 
% \end{itemize} 

%\textcolor{blue}{is it necessary to introduce ShapeContext first??} \textcolor{red}{Inspired by ShapeContext\cite{belongie2002shape}}, 
% \textcolor{red}{To address ...}, 
To address the data-efficiency problem in 3D scene understanding,
Hou et al.\cite{hou2021exploring} proposed 
Contrastive Scene Contexts makes utilization of both point-level correspondences and spatial contexts in a scene.
The method partitions the space into inhomogeneous cells based on the relative distance and angle between points and performs contrastive learning in each spatial cell separately. 
It introduces spatial information by sampling negative samples in spatial cells. 
Contrastive Scene Contexts has been shown to outperform PointContrast in semantic segmentation and detection tasks on the S3DIS and ScanNet datasets.

%\textcolor{red}{Apart from formats, the invariance of 3D representations also includes view  invariance, modal invariance between depth and RGB images, etc. }
To systematically and fairly compare different invariants in pre-training, Li and Heizmann ~\cite{li2022closer} proposed a unified contrastive learning framework that leverages the invariances of 3D features, such as perspective-invariance between views of the same scene, modality-invariance between RGB and depth images, and format-invariance between point clouds and voxels. In addition, they introduced a simple and efficient method for jointly pre-training 3D encoders and depth graph encoders. However, pre-training two encoders together does not necessarily guarantee the optimal performance of each individual encoder, indicating the need for further research in this area. 

In addition to investigating point cloud formats, modalities, and view invariance, Chen et al.~\cite{chen2021unsupervised} proposed a multi-level self-supervised learning method for geometric sampling invariant representations. The method aims to learn the intrinsic features of point clouds at various  sampling patterns and densities. To accomplish this, it learns a function $E$ that is invariant over geometric sampling by maximizing the mutual information between different down-sampling results and minimizing the Chamfer Distance between the results of down-sampling and up-sampling and the original point cloud. 

\vspace{-0.2cm}
\subsubsection{Transformer-based Methods}
\vspace{-0.1cm}
The Transformer architecture, introduced by Vaswani et al. in 2017~\cite {vaswani2017attention}, has become very popular in recent years due to its state-of-the-art performance in several NLP tasks, including machine translation and language modeling. One of the main advantages of Transformers is their ability to capture long-range dependencies in data sequences using the self-attention mechanism. Point cloud data, which consists of a set of unordered points, lacks the inherent sequential order found in natural languages. This poses a challenge for traditional DNN architectures, such as convolutional neural networks and recurrent neural networks, which struggle to effectively capture global information from unordered point sets due to their reliance on fixed grid structures or sequential processing. In contrast, the Transformer architecture can operate on the unordered points without the need for any explicit positional encoding, and efficiently capture long-range dependencies between elements by utilizing self-attention mechanisms. As a result, researchers have started exploring the potential of Transformer-based models in 3D point cloud pre-training.

Mask point cloud transformer (MPCT) is a commonly used approach that randomly masks input points and recovers them using Transformer's ability. To optimize the reconstruction performance of MPCT, ContrastMPCT~\cite{wang2022self}, which is a self-supervised pre-training framework based on contrastive learning, computes the contrast loss between the point cloud reconstructed by MPCT and the original point cloud. ContrastMPCT adopts two contrast loss functions: Jensen-Shannon divergence (JSD)-based loss and InfoNCE loss. In contrast to Point-BERT, the contrastive learning design of ContrastMPCT makes it unnecessary to pre-train a ``tokenizer'' and makes it easier to train. 

POS-BERT~\cite{fu2022pos} is another Transformer-based model that introduces contrastive learning on class tokens between global and local point clouds obtained by different cropping ratios, building on the basis of Point-BERT. The contrastive learning approach in POS-BERT maximizes the class token consistency among point cloud pairs, thereby effectively learning high-level semantic representations. In subsequent work, Fu et al.\cite{fu2022distillation} further improved the accuracy in linear classification and several semantic segmentation tasks by leveraging global shape information and the relationship between global shape and local structure through 3D-ViT using knowledge distillation and contrastive learning. 

\vspace{-0.2cm}
\subsubsection{SimSiam-based Methods}
\vspace{-0.1cm}
SimSiam-based methods offer an alternative to InfoNCE-based loss functions commonly used in 3D point cloud contrastive learning. 
While InfoNCE-based methods often require a large number of negative samples to achieve good performance, the SimSiam architecture~\cite{chen2021exploring} only necessitates positive sample pairs during training. In SimSiam-based approaches, two or more Siamese neural networks with shared weights are utilized to separately encode different augmented versions of the same input. These encodings serve to minimize the negative cosine similarity for contrastive learning. Additionally, a prediction module is added to one of the branches, and an asymmetric design with a stop-gradient is employed to prevent training crashes. By leveraging positive sample pairs and shared weights, SimSiam-based methods reduce the amount of data required for training and enhance overall performance in 3D point cloud understanding tasks.

Mei et al. \cite{mei2022unsupervised} introduced ConClu, a self-supervised point cloud pre-training method inspired by contrastive learning and clustering. ConClu reduces the reliance on negative samples in contrastive learning by employing a SimSiam-based architecture. 
Moreover, as ConClu is not tied to specific neural network architectures, it can serve as a general feature extractor, enhancing the performance of various 3D point cloud models. 

Chen et al. \cite{chen20224dcontrast} considered the importance of dynamic movement information of 3D shapes generated while moving through static 3D environments for tasks such as semantic segmentation. They proposed incorporating 4D sequence information and constraints into 3D representation learning by employing contrastive learning under 3D-4D constraints. 
In a similar vein, this method uses a SimSiam-based contrastive learning architecture to compute the contrastive loss under three constraints: spatial correspondences between frames (3D-3D), spatial-temporal correspondences (3D-4D), and dynamic correspondences (4D-4D). This approach enhances time efficiency compared to methods based on InfoNCE loss. However, it requires high memory consumption for pre-training 4D data. 

\vspace{-0.2cm}
\subsubsection{Other Methods}
\vspace{-0.1cm}
While most existing 3D point cloud SLL methods focus on mainstream tasks, such as object classification, semantic segmentation, part segmentation, and object detection tasks, Yang et al.~\cite{yang2021unsupervised} proposed a new task: point cloud object co-segmentation, which aims to segment common objects contained in a collection of point clouds. 
They treated the co-segmentation training problem as an object point sampling problem and designed a self-supervised method combining mutual attention and contrastive learning pre-training framework.
Their method includes two InfoNCE-based contrastive losses, computed at the point and object levels, for contrastive learning within and between point clouds. Moreover, point cloud object co-segmentation can provide pseudo-labeling for object classification tasks, improving performance and making it a valuable and promising direction in the field of 3D point cloud processing.

\vspace{-0.2cm}
\subsubsection{Challenges and Opportunities} 
\vspace{-0.1cm}
Contrastive learning is an emerging technique for pre-training point clouds that presents several challenges and opportunities.

On the one hand, contrastive learning is highly flexible and can be applied to point cloud datasets of various types and sizes. This flexibility enables researchers to compare data from different modalities, thereby improving the model's overall performance. Furthermore, contrastive learning is highly robust to sample imbalance and label noise, as it does not depend on the distribution of data and the accuracy of labels.

On the other hand, a significant challenge associated with contrastive learning based SSL is the high computational cost. This requirement for substantial computing resources and time can result in high training expenses.

% \begin{itemize}
%     \item \textbf{Flexibility:} The contrastive-learning-based method is applicable to point cloud datasets of different types and sizes, and can be compared with data from other modalities to improve model performance.
%     \item \textbf{Robustness:} The contrastive learning pre-training method does not depend on the distribution of data and the accuracy of labels, so it is robust to sample imbalance and label noise.
%     \item \textbf{High computational cost:} The self-supervised point cloud pre-training method based on contrastive learning requires a lot of computing resources and time, leading to the high training cost.
% \end{itemize}

% \begin{figure}[ht]
%     \centering
%     \includegraphics[width=\linewidth]{Figures/contrastive-learning-based/Conclu-pipeline.pdf}
%     \caption{The illustration of Conclu by Mei et al.~\cite{mei2022unsupervised}: it designs a self-supervised point cloud pre-training method based on the ideas of SimSiam and clustering.}
%     \label{fig:Conclu}
% \end{figure}

\vspace{-0.3cm}
\subsection{Spatial-based SSL}
\vspace{-0.1cm}

\begin{figure}[t]
    \centering
    \includegraphics[width=0.9\linewidth]{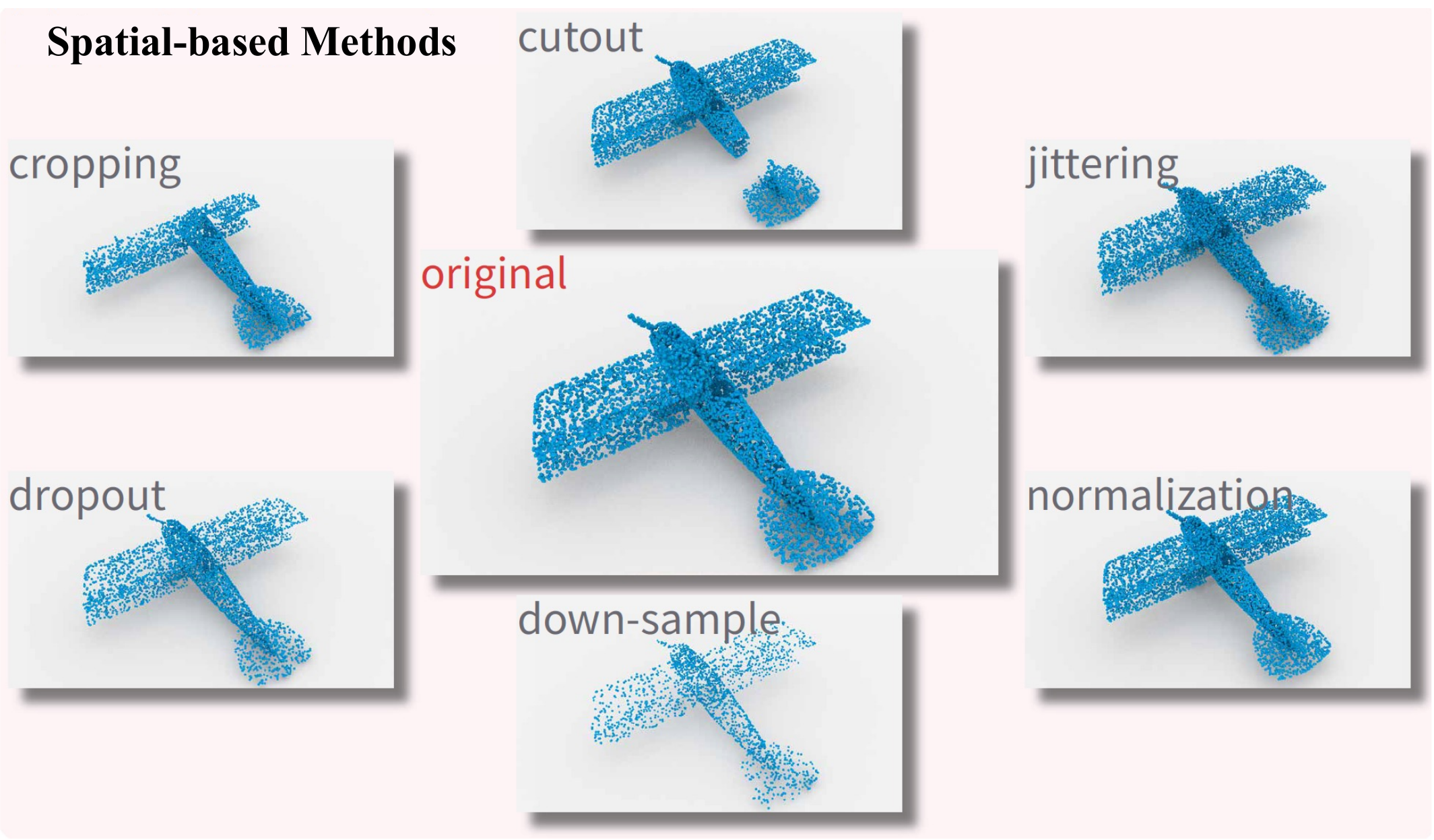}
    \vspace{-0.5cm}
    \caption{Spatial-based SSL methods generate degraded point clouds by applying geometric transformations and subsequently pre-train models through the process of recovering the original spatial information. This approach leverages the rich geometric context inherent in point cloud data for effective representation learning. Images courtesy of Huang et al.~\cite{huang2021spatio}. }
    \label{spatial}
\vspace{-0.3cm}
\end{figure}

% Due to the difficulty of constructing 3D point cloud datasets and the complexity of 3D scene understanding tasks, training pre-trained models with low annotation cost and high portability has become a hot research topic nowadays. 
Spatial-based SSL methods harness the abundant geometric information inherent in point clouds to develop pretext tasks. 
Figure~\ref{spatial} illustrates several typical geometric transformations, such as cropping, cutout, jittering, dropout, down-sampling, and normalization.
By employing the recovery process of these spatially degraded point clouds, models can be effectively pre-trained, taking advantage of the rich spatial context within the data. 

\vspace{-0.2cm}
\subsubsection{Rotation-based methods}
\vspace{-0.1cm}
Among the various spatial methods, rotation-based techniques are the most widely used pre-task for self-supervised pre-training. These methods have been studied in the context of images~\cite{zhai2019s4l, hendrycks2019using}.
Compared with image-oriented rotation methods, rotation operations on point clouds are more fitting to the modality of point cloud data.

Huang et al.\cite{huang2021spatio} employed Spatial-Temporal Representation Learning (STRL) to process time-dependent frames as input alongside spatial data augmentation, achieving good results in synthetic, indoor, and outdoor datasets. Their spatial data augmentation techniques include random crop, random cutout, random jittering, random drop-out, down-sampling, and normalization. 
These spatial augmentations transform the input by altering the local geometry of the point cloud, enabling STRL to learn a more effective spatial structure representation. 
Sun et al.~\cite{sun2021canonical} introduced a capsule network that employs a self-supervised approach for pre-training object-centric representations using random rotation. This method has demonstrated advantages for tasks such as 3D point cloud reconstruction, canonicalization, and unsupervised classification. 
Poursaeed  et al.~\cite{poursaeed2020self} introduced an orientation estimation-based pre-training method that utilizes a rotation approach to obtain rich learned features from unlabelled data. 

As the use of 3D point clouds in security-critical applications continues to grow~\cite{sun2021adversarially}, ensuring the adversarial attack robustness of 3D deep learning models has become a crucial concern. There are three representative SSL proxy tasks, namely 3D rotation prediction, 3D jigsaw, and autoencoding, which aim to improve the models' resilience to adversarial attacks.

%The novel self-supervised learning method (MD) proposed by Sun et al.\cite{sun2022self} (mixing and separation, i.e., after mixing the input shapes and making the model to separate the mixed shapes).
%The advantages are twofold: (1) point cloud datasets are smaller than image datasets. The mixing operation leads to larger training samples.
%(2) The disentangling operation enables the model to mine the geometric prior knowledge.
% \subsubsection{Correction-based Methods} 

\vspace{-0.2cm}
\subsubsection{Cluster-based Methods}
% \vspace{-0.1cm}
Cluster-based SSL methods capture both local geometric structures and global spatial relationships within point clouds. For instance, the self-labeled three-dimensional recognition (SL3D) framework\cite{cendra2022sl3d} not only addresses two coupled goals of clustering and learning feature representation but is also capable of solving various 3D recognition tasks. Another example is joint learning of multi-task models~\cite{hassani2019unsupervised}, which defines three self-supervised tasks including clustering, reconstruction, and self-supervised classification. 

\vspace{-0.2cm}
\subsubsection{Recognition-based Methods}
\vspace{-0.1cm}
Recognition-based SSL approaches utilize the inherent spatial structure of point clouds to learn meaningful representations through tasks such as point cloud classification, segmentation, or object recognition, without relying on manual annotations. These methods aim to leverage the spatial information of point clouds to design recognition-based pre-tasks that can improve the quality of learned representations.

For instance, Sharama and Kaul\cite{sharma2020self} proposed the cover tree method for few-shot learning (FSL) in 3D point cloud pre-training. This method involves designing two self-supervised pre-training tasks and using the cover tree to hierarchically partition the point cloud into subsets that lie within balls of varying radii at each level of the cover tree. The fully-trained self-supervised network’s point embeddings are then input to the downstream task’s network, such as classification and segmentation.

Rao et al.~\cite{rao2022pointglr} extended the self-supervised structural representation learning approach to more complex 3D scenes and demonstrated its good generalization ability and robustness. Their method learns point cloud representations through bidirectional reasoning between local geometries at different abstraction hierarchies and the global structure.

Sun et al.~\cite{sun2022unsupervised} proposed a random block detection pre-task for SSL pre-training the detection model. Specifically, this method involves sampling random blocks from the original point clouds, which are then fed into the Transformer decoder. Subsequently, the Transformer is trained by detecting the locations of these blocks. In this way, the pre-trained detection model outperformed the train-from-scratch detection model on the challenging ScanNetV2 dataset.

Yamada et al.~\cite{yamada2022point} proposed the PC-FractalDB pre-training model based on fractal geometry. 
The PC-FractalDB is automatically built by defining a fractal category by utilizing variance threshold and instance augmentation with FractalNoiseMix.
A 3D fractal scene is generated by randomly selecting 3D fractal models and translating these from the origin.
This method directly acquires feature representation for 3D object detection in the pre-training stage and assists in fine-tuning when the dataset is limited to a small number of training data and annotations.

%The proposed fractal geometry-based point cloud fractal database (PC-FractalDB) pre-training model~\cite{yamada2022point}, directly enables the acquisition of feature representation for 3D object detection in the pre-training phase. PC-FractalDB pre-training assists in fine-tuning when the dataset is limited to the number of training data and annotations.
% The PC-FractalDB can also suggest an effective approach to the difficulties in constructing 3D point cloud datasets.

\vspace{-0.2cm}
\subsubsection{Spatial Mapping-based Methods}
% \vspace{-0.1cm}
In addition to the aforementioned methods, there exist spatial mapping-based approaches that transform irregular 3D point clouds into regular representations that can be more easily processed by conventional 2D networks. Zhang et al.~\cite{zhang2023flattening} proposed Flattening-net, which converts irregular 3D point clouds into a complete and regular 2D representation known as point geometry images, allowing for direct application of conventional 2D networks to the transformed data.
They also developed RegGeoNet~\cite{zhang2022reggeonet} for handling large-scale point clouds in real-world applications. RegGeoNet utilizes global anchor embedding to produce a global parameterization of downsampled sparse anchors. It then adopts a local patch embedding module to generate the local parameterization of patches centered at the anchor positions.

\vspace{-0.2cm}
\subsubsection{Challenges and Opportunities} 
\vspace{-0.1cm}
Point cloud data contains rich spatial information, such as the distance between points and normal vectors, which can provide more features for learning and improve the representation learning ability of the model. Therefore, SSL methods that use spatial information can learn the structure and characteristics of the point cloud, enhancing the model's generalization ability.

However, point density and distribution can influence the extraction and utilization of point cloud spatial information, requiring pre-processing and normalization of the input point cloud data. Additionally, to process point cloud data with complex geometries, advanced processing methods may be necessary.

\vspace{-0.3cm}
\subsection{Temporal-based SSL}
\vspace{-0.1cm}

\begin{figure}[t]
    \centering
    \includegraphics[width=0.9\linewidth]{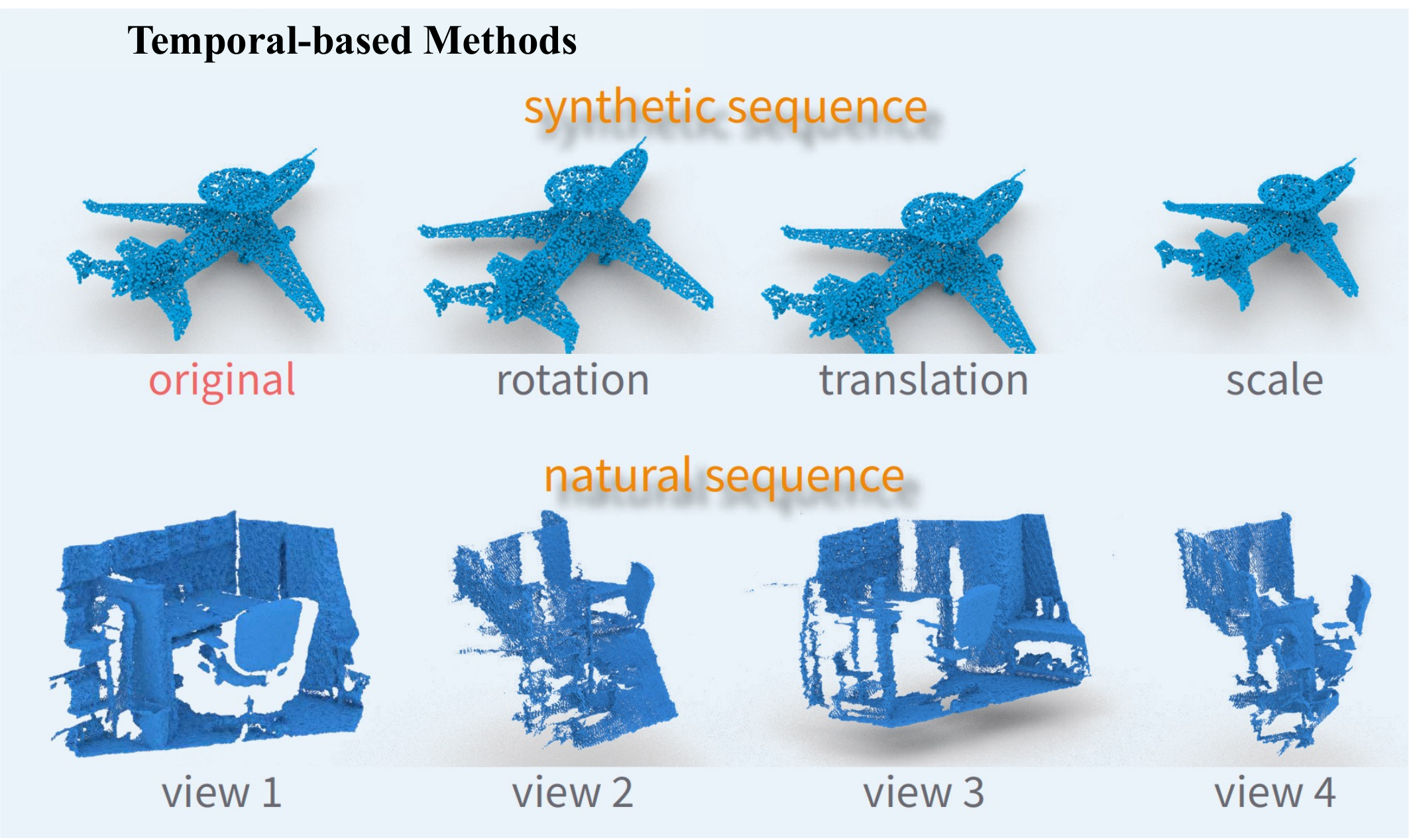}
    \vspace{-0.5cm}
    \caption{Illustration of temporal-based methods. 
    For synthetic data (row 1), the original input can be augmented by rotation, translation, and scaling to emulate the viewpoint change. 
    For natural sequences (row 2), two frames are sampled with a natural viewpoint change in depth sequences as the input pair.
    The temporal difference between the inputs enables models to capture the randomness and invariance across different viewpoints.
    Images courtesy of Huang et al.~\cite{huang2021spatio}. }
    \label{temporal}
\vspace{-0.3cm}
\end{figure}

Temporal-based SSL methods emphasize the use of inherent temporal information present in sequences or artificially generated transformations. Point cloud sequences consist of continuous point cloud frames, analogous to video data. 
Examples include indoor point cloud sequences converted from RGB-D video frames and LiDAR sequential data comprised of successive point cloud scans. These point cloud sequences hold a wealth of temporal information, which can be harnessed by designing pretext tasks for self-supervised learning and employing the extracted data as supervisory signals to train networks. 
The resulting learned representations can be effectively transferred to a variety of downstream tasks.

Huang et al.~\cite{huang2021spatio} introduced the spatio-temporal representation learning (STRL) framework. STRL adapts the bootstrap your own latent (BYOL) approach from 2D to 3D vision, extracting both spatial and temporal representations from 3D shapes. In particular, it regards two adjacent point cloud frames as positive pairs and minimizes the mean squared error between the learned feature representations of these sample pairs. 

Contrastive learning, which has been extensively explored as detailed in Section~\ref{CL-based-methods}, can also be considered an effective means to learn temporal information.
Specifically, frame-level contrastive learning methods aim to align continuous frames and learn their representations.
Chen et al.~\cite{chen20224dcontrast} utilized synthetic 3D shapes moving within a static 3D environment to create dynamic scenes and pairs of samples in temporal order. Then, They  carried out contrastive learning to learn 3D representations for dynamic understanding. The pre-trained model can be effectively transferred to downstream 3D semantic scene understanding tasks.

SSL pre-training utilizing temporal context structures has proven to be effective in 3D computer vision tasks. 
This direction holds promise; however, further exploration and development are necessary to better capture and exploit temporal context information. 

On the one hand, utilizing temporal information in point cloud sequences allows models to learn object motion and changes, enhancing the model's robustness and generalization capabilities. For time-critical applications, employing temporal information in point clouds can improve the accuracy of object recognition and tracking for dynamic objects.

On the other hand, compared to SSL methods based on single-frame point cloud data, using temporal information in point clouds necessitates a significant amount of point cloud sequence data for model training. Additionally, incorporating temporal information in the self-supervised point cloud pre-training methods requires modeling and processing point cloud sequence data, which can be computationally demanding. In practical applications, point cloud sequence data may be subject to noise interference or missing data, which could negatively impact the model's learning effectiveness.

% \begin{itemize}
%     \item The model can leverage the temporal information in point cloud sequences to learn object motion and changes, improving the model's robustness and generalization ability.
%     \item For real-time applications, using temporal information in point clouds can improve the accuracy of object recognition and tracking for dynamic objects.
%     \item Compared to traditional self-supervised pre-training methods based on images, using temporal information in point clouds can more accurately reflect object motion and changes.
%     \item \textbf{Requires a large amount of point cloud sequence data:} Compared to self-supervised pre-training methods based on single-frame point cloud data, using temporal information in point clouds requires a large amount of point cloud sequence data to train the model.
%     \item \textbf{Noise and missing data in temporal sequences:} In practical applications, point cloud sequence data may be subject to noise interference or have missing data, which can affect the model's learning effectiveness.
%     \textbf{High computational costs:} Using temporal information in the self-supervised point cloud pre-training methods requires modeling and processing point cloud sequence data, which can be computationally intensive.
% \end{itemize}
\vspace{-0.3cm}
\subsection{Multi-modality SSL}
\vspace{-0.1cm}

\begin{figure}
    \centering
    \includegraphics[width=0.9\linewidth]{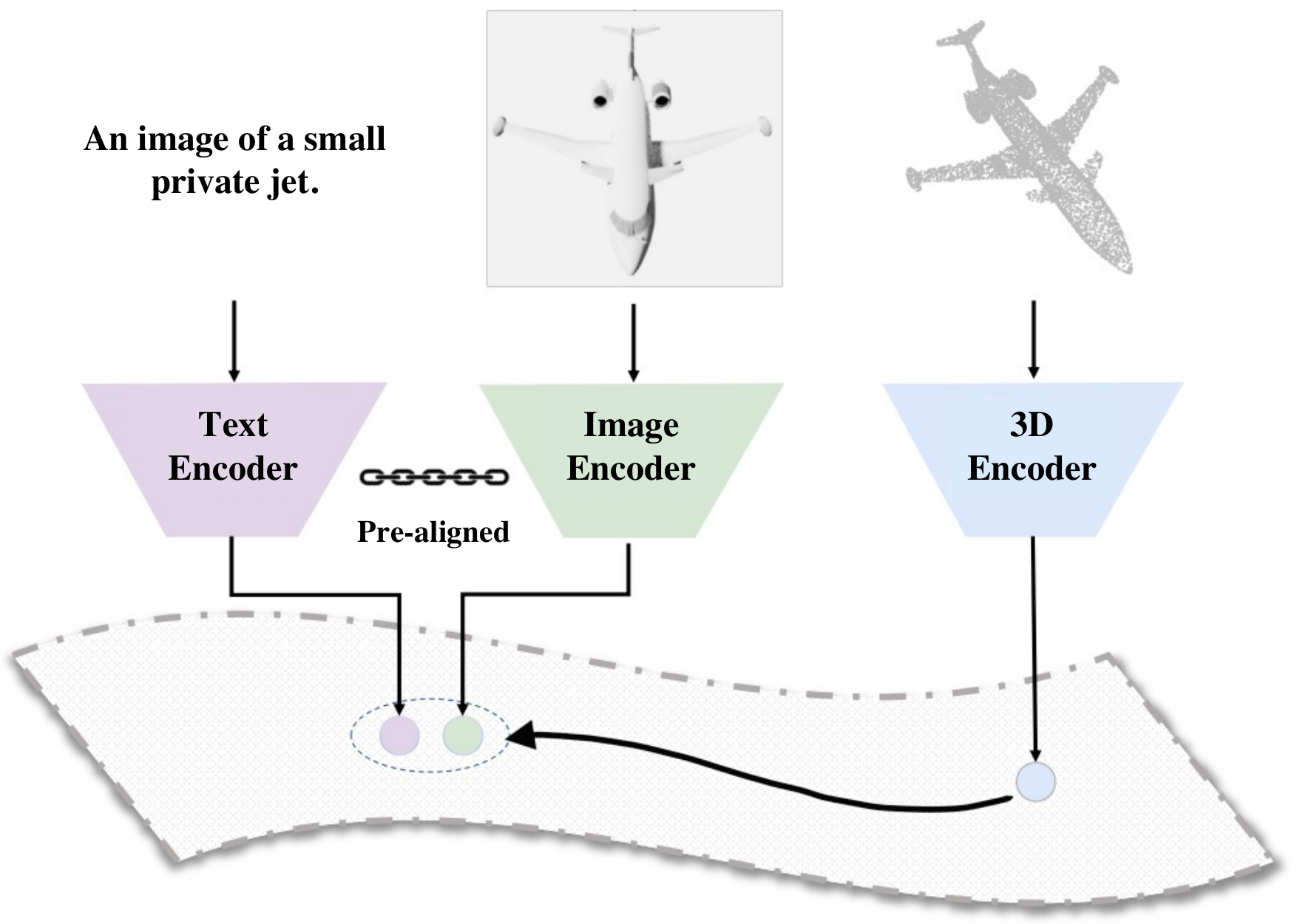}
    \vspace{-0.5cm}
    \caption{Multi-modality SSL methods align features from images, texts, and point clouds for point cloud understanding. Additional modalities, such as depth images, can also be used for alignment. Image courtesy of Xue et al.~\cite{xue2022ulip}.}
    \label{multi-modality}
\vspace{-0.3cm}
\end{figure}

Multi-modality learning aims to leverage the correlation across different modalities, such as images, texts, and point clouds (Figure.~\ref{multi-modality}). 
The advantages of these approaches include the ability to leverage complementary information from multiple sources, robustness to missing or noisy data in any one modality, and improved generalization to new environments.

\vspace{-0.2cm}
\subsubsection{Single-view Methods}
\vspace{-0.1cm}

2D images have been shown to complement 3D point clouds in numerous studies, making them the most commonly used modality for cross-modal learning of point clouds. Furthermore, self-supervised cross-modal pre-training learning between 2D and 3D is believed to be a promising approach to understanding 3D point clouds. 

Li et al. \cite{li2022closer} proposed a unified framework for systematically comparing various invariants in different pre-training strategies by jointly pre-training two distinct encoders. This framework unifies the comparison of different formats and network structures in a contrastive manner, incorporating multi-modal approaches between RGB images and point clouds. Janda et al.~\cite{janda2022self} took this step further by proposing a multi-modal pre-training method using only single scan point clouds and corresponding images. This approach treats learned features from images as the target of contrastive loss to pre-train the 3D model. 

Sun et al.~\cite{sun2022self} proposed an SSL method dubbed mixing and disentangling (MD) for point clouds, addressing the issue of limited data in open-source datasets. During the pre-training process, MD combines two original point clouds from the training set and then requires the model to reconstruct the original shapes based on the corresponding 2D projections from the mixed point cloud. In this way, the model learns geometric prior knowledge related to the shape.

In addition to global cross-modal learning between images and point clouds, some studies have sought to exploit the contribution of local correspondence, such as pixel-point correspondence, for transferring 2D local knowledge to 3D. 
For example, Zhou et al.~\cite{zhou2022pointcmc} proposed PointCMC, a novel SSL method for modeling cross-modal multi-scale correspondences without the need for cumbersome reconstruction steps. PointCMC consists of three components: a local-to-local (L2L) module, a local-to-global (L2G) module, and a global-to-global (G2G) module, which together enable comprehensive modeling of both local and global correspondences between point clouds and images. 
Notably, the L2L module, based on a two-branch local attention block, can directly learn local correspondence, and the local-to-global correspondence is investigated for the first time in this approach. 

\vspace{-0.2cm}
\subsubsection{Multi-view Methods}
\vspace{-0.1cm}
In contrast to single-view methods, multi-view approaches aim to enhance the robustness and generalization of learned representations by incorporating information from multiple views of point clouds. Furthermore, distinct views of the same scene captured from various angles offer diverse insights, improving the overall understanding of the environment.

For instance, Jing et al.\cite{jing2021self} introduced a multi-modality and multi-view SSL method to jointly learn features from 2D images and 3D point clouds. They demonstrated that two types of constraints can serve as self-supervised signals: cross-modality correspondence and cross-view correspondence. 

Zhang et al.\cite{zhang2022learning} proposed I2P-MAE learn 3D representations from a 2D pre-trained model through a masked auto-encoder. I2P-MAE first trains the 2D model to extract multi-view visual features from the input point cloud and then implements two image-to-point strategies: the 2D-guided masking strategy and the 2D-semantic reconstruction strategy for the 2D feature map and 2D visual features, respectively. These strategies help transfer knowledge from 2D to 3D domains. 
The 2D-guided masking strategy directs semantically important point tokens to remain visible to the encoder. 
In contrast, the 2D-semantic reconstruction strategy guides the reconstruction of multi-view 2D features from visible tokens. 
By reconstructing both 3D coordinates and 2D semantics, I2P-MAE can obtain better 3D representation, effectively transferring the knowledge from 2D pre-training to 3D pre-training. 
Compared with other pre-training methods requiring multiple scans from different 3D views for comparative learning, I2P-MAE has higher scalability. 

% \begin{figure}[ht]
%     \centering
%     \includegraphics[width=\linewidth]{Figures/multi-modality/CM-CV-pipeline.pdf}
%     \caption{The framework of CM-CV proposed by Jing et al.\cite{jing2021self}: It proposed a framework for self-supervised 2D and 3D feature learning by cross-modality and cross-view correspondences.}
%     \label{fig:CM-CV}
% \end{figure}

Inspired by auto-encoding transformations (AET)~\cite{zhang2019aet}, Gao et al.~\cite{gao2021self} introduced the idea of ``transformation equivariant'' to 3D point cloud understanding and proposed a multi-view transformation equivariant representation (TER) learning method 
 called MV-TER. Specifically, MV-TER transforms the 3D point cloud and then decodes the 3D transformation on its corresponding projection image, compelling the model to learn the intrinsic 3D object representation. The ``transformation equivariant'' approach discards labeling information, making MV-TER broadly applicable and scalable.

% Yamada et al.~\cite{yamada2022point} introduced the concept of a multi-viewpoint cloud, which represents a 3D point as a set of features extracted from multiple views. \textcolor{red}{This approach aims to achieve better performance in 3D classification, retrieval, and semantic segmentation. }

Tran et al.~\cite{tran2022self} proposed a multi-view pre-training method that employs local pixel-point-level correspondence loss and global image-point cloud-level loss jointly as supervised signals. Initially, they trained an image encoder using a set of object views, capturing the features of the object from various perspectives. Then, they performed global knowledge transfer and point-level knowledge transfer by minimizing the distance between global features and local features. 

Since different forms of 2D rendered images bring different information, Zhang et al.~\cite{zhang2022self} proposed PointVST, where the view-specific cross-modal translation is devised to convert a 3D point cloud to diverse forms of rendered images, including silhouette, contour, and depth images. Combined with a point-wise visibility mask, the supervision of these types of rendered images is utilized to pre-train the model.

\vspace{-0.2cm}
\subsubsection{Knowledge Transfer Methods} 
\vspace{-0.1cm}
By bridging the gap between 2D and 3D modalities, it is possible to leverage successful 2D pre-trained models for learning  effective representations in the 3D point cloud domain. This approach can partially address the limitations in 3D pre-training data currently faced by researchers. 

Rather than using features from different modalities for correspondence, some studies enable the transfer of 2D pre-trained knowledge to 3D by transforming point clouds into 2D images. As a result, large-scale pre-trained models from the 2D domain can be directly applied in the 3D domain. 

For instance, Wang et al.\cite{wang2022p2p} presented a novel point-to-pixel prompting strategy that transforms 3D point clouds into 2D color images, enabling the utilization of large-scale 2D pre-trained models for learning 3D point cloud representations. They also conducted extensive experiments on various 2D models to investigate the performance of different architecture designs on point cloud understanding.

Qian et al.\cite{qian2022pix4point} proposed an improved network PViT based on a standard Transformer, as well as a framework  called Pix4Point, which uses image pre-training to enhance the performance of standard Transformer models for point cloud understanding. PViT uses progressive point patch embedding as the tokenizer and adds feature propagation with global representation as the decoder. The Pix4Point framework directly exploits a pre-trained Transformer in the image domain to enhance the understanding of downstream point clouds through the use of the tokenizer and the decoder. 

\vspace{-0.2cm}
\subsubsection{Text-assisted Methods}
\vspace{-0.1cm}
In recent years, vision-language models have gained significant attention, including models such as CLIP~\cite{radford2021learning} and Stable Diffusion~\cite{rombach2022high}. 

CLIP learns transferable visual features with natural language supervision. For zero-shot classification of ``unseen'' categories, CLIP exploits the pre-trained correlation between vision and language to perform open-vocabulary recognition. As a pioneer work using CLIP, Zhang et al \cite{zhang2022pointclip} proposed PointCLIP, which demonstrated that CLIP, pre-trained through large-scale image-text pairs in 2D, can be generalized to the 3D domain. By designing an inter-view adapter to incorporate global features from multiple views, PointCLIP can effectively fuse few-shot 3D knowledge into 2D models. This approach allows for the fine-tuning of the adapter to achieve significant performance improvement. 

% \begin{figure}[ht]
%     \centering
%     \includegraphics[width=\linewidth]{Figures/multi-modality/P2P-pipeline.pdf}
%     \caption{The pipeline of P2P~\cite{wang2022p2p}. P2P transfers a point cloud into a color image and exploits the ability of 2D pre-trained models for 3D understanding.}
%     \label{fig:P2P}
% \end{figure}

% \begin{figure}[ht]
%     \centering
%     \includegraphics[width=\linewidth]{Figures/multi-modality/PointCLIP-pipeline.pdf}
%     \caption{The Pipeline of PointCLIP~\cite{zhu2022pointclip}. PointCLIP projects the point cloud onto multi-view depth maps and conducts 3D recognition via CLIP pre-trained in 2D.}
%     \label{fig:PointCLIP}
% \end{figure}
Although PointCLIP has demonstrated its effectiveness under various few-shot settings, it still suffers from limitations in zero-shot tasks~\cite{zhu2022pointclip}:
(1) Sparse visual projection. The simple projection of 3D point clouds into sparse depth maps with single depth values in PointCLIP may not adequately represent real-world images. This discrepancy can introduce interference to the CLIP image encoder and limit its effectiveness in zero-shot tasks, as it struggles to capture and understand the full complexity of the 3D point clouds. (2) Na\"ive textual prompting.
PointCLIP uses relatively simple textual prompts compared to the original CLIP model. By adding only a few domain-specific words, the prompts might not capture the overall shape and structure of the 3D point cloud, potentially limiting the performance of the model.

Zhu et al \cite{zhu2022pointclip} proposed PointCLIP V2 to address the limitations of PointCLIP in zero-shot settings. The main improvements introduced by PointCLIP V2 are a 4-step projection module and an optimized prompt selection strategy. 
The former aims to reduce the visual discrepancies between the generated depth map and real-world images, allowing the CLIP image encoder to capture and understand the 3D point cloud more effectively. The latter leverages large-scale language models like GPT-3\cite{brown2020language} for better capturing the properties and characteristics of the 3D point clouds, enhancing the model's performance in zero-shot tasks. 

Huang et al.\cite{huang2022frozen} proposed EPCL that takes advantage of the frozen CLIP model to train a point cloud model directly without the need for 3D pre-training or 2D-3D data matching. The method uses a specially designed tokenizer to weakly align 2D and 3D features, further refining the alignment through the CLIP model to narrow the gap between 2D images and 3D point clouds.

Xue et al.\cite{xue2022ulip} proposed ULIP, a method that learns unified representations of three modalities (2D images, text and 3D point clouds) to enhance the understanding of 3D models. The main challenge addressed by ULIP is the lack of accessible triplet data. To overcome this issue, ULIP employs a two-step approach. It first pre-trains a common vision-language feature space using large-scale image-text pairs. This pre-training process helps the model to capture meaningful and transferable visual and textual features. After establishing the vision-language feature space, ULIP aligns a small number of automatically synthesized point cloud triplets into the pre-aligned visual-language feature space. This step allows ULIP to integrate any 3D architectures and support a variety of cross-modal downstream tasks. 

\vspace{-0.2cm}
\subsubsection{Challenges and Opportunities}
\vspace{-0.1cm}
Despite the advancements of SSL pre-training methods based on  multi-modality, several challenges still remain. First, 
collecting large-scale multi-modal datasets can be challenging due to the need for diverse sensors or acquisition devices. 
Second, accurate alignment of multi-modal data is crucial for ensuring meaningful correspondence between different modalities. This can be difficult to achieve, especially for large-scale datasets. Third, designing models that can effectively handle and fuse information from multi-modalities is a complex task. 

The utilization of multi-modal data presents at least two opportunities for self-supervised pre-training methods. 
1) Enhanced representation: Multi-modal data can provide complementary information, leading to better representation and alignment of features. This helps design a range of self-supervised pre-training tasks that are more transferable across multiple modalities. 2) Improved generalization: Multi-modal data contains more scene information and noise interference, which can help train models with better generalization performance. 

\vspace{-0.3cm}
\section{Outdoor Scene-level SSL}
\vspace{-0.1cm}
\label{outdoor-level}

The primary distinction between indoor-level and outdoor-level SSL stems from the complexity and sparsity of the point cloud data. 
Indoor-level SSL focuses on environments with relatively less variability and higher point cloud density, such as rooms, buildings, or other enclosed spaces. Outdoor-level SSL, on the other hand, deals with more complex and dynamic environments like streets, forests, and urban landscapes, where point clouds are typically sparser. The sparsity of outdoor point clouds compared to object- and indoor scene-level data results in a scarcity of semantic information as there may be only a few points representing an object or category. Moreover, the perception of outdoor scene-level point clouds is often considered as an open-set problem due to various unseen categories, making the task more challenging.

Autonomous driving systems typically rely on LiDAR data for outdoor scenes, which are sparse and lack color information~\cite{nunes2022segcontrast}. While unlabeled LiDAR data is easily obtainable~\footnote{For example, an autonomous car can collect about 200,000 frames of point clouds within just 8 hours~\cite{yin2022proposalcontrast}.}, labeled data is expensive to produce. This presents a significant challenge for building perceptual models in autonomous driving that rely on large-scale labeled 3D data~\cite{min2022voxel}. Consequently, recent works have focused on leveraging self-supervised learning on large amounts of unlabeled 3D data to improve the performance of downstream tasks in autonomous driving.

The above-mentioned challenges make pre-training on outdoor scene-level point clouds a non-trivial endeavor.
Nevertheless, certain methods, such as prediction- and flow-based methods, have been developed to align with the intrinsic characteristics of outdoor scene-level point clouds.

%\textcolor{red}{briefly discuss the main difference between indoor-level and outdoor-level and explain why such difference matters}

% However, labeling 3D data is expensive and time-consuming.
% Therefore, many scholars have conducted research related to the self-supervised learning of large amounts of unlabeled 3D data in autonomous driving. 
\vspace{-0.3cm}
\subsection{Reconstruction-based SSL}
\vspace{-0.1cm}

\begin{figure}[t]
    \centering
    \includegraphics[width=\linewidth]{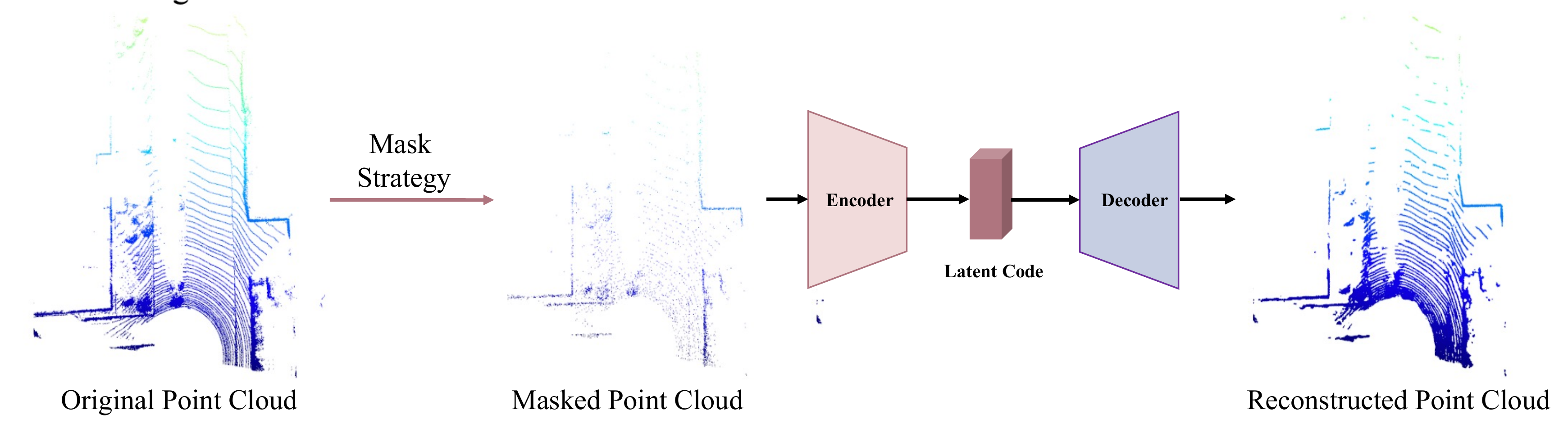}
    \vspace{-0.8cm}
    \caption{Illustration of outdoor scene-level reconstruction-based methods. The auto-encoder architecture is used to reconstruct masked point clouds. The methods can be classified into two groups based on the representation of 3D data. Images courtesy of Min et al.~\cite{min2022voxel}. }
    \label{scene-reconstruction}
\vspace{-0.3cm}
\end{figure}

Similar to object- and indoor scene-level data, the exploration of reconstruction-based self-supervised pre-training has become an important area for outdoor scene-level data (as illustrated in Figure~\ref{scene-reconstruction}). However, the sparsity of outdoor scene-level point clouds presents challenges to 3D reconstruction. To address the difficulties in the direct processing of large-scale point clouds, voxel-based and Bird's Eye View (BEV)-based reconstruction methods have emerged as effective ways to tackle these challenges.

\vspace{-0.2cm}
\subsubsection{Voxel-based Methods} 
\vspace{-0.1cm}

Voxels are a commonly used representation in outdoor scenes and can be highly efficient for processing large-scale point cloud datasets. They allow for operations on a fixed-size grid instead of processing each individual point, which is especially useful given the sparsity of outdoor scene-level point clouds~\cite{min2022voxel}. 
% Additionally, \textcolor{red}{voxel representations preserve the spatial relationships between points within the same voxel}
% \fb{the voxel representation of point clouds is capable of preserving the spatial relationships between points within each voxel.}
% , making it possible to encode local geometric features and relationships that are important for many 3D tasks. 
Therefore, using voxel-based masked auto-encoders for pre-training large-scale point clouds is a plausible solution to this problem.
% At present, the main representations of 3D point clouds are point-based and voxel-based representations, and it is relatively simple to reconstruct the features of large-scale point clouds\cite{min2022voxel}. Meanwhile, voxel-based point cloud representations are very popular, but the Transformer architecture cannot handle a large number of unmasked voxels of large-scale point clouds, so reacquiring new learned features from the input point clouds through special encoders and decoders, and pre-training large-scale point clouds using Masked Auto-Encoder provide a plausible solution to this problem.

Two VoxelMAE methods have been proposed to improve 3D perception for autonomous driving~\cite{min2022voxel, hess2022masked}. In ~\cite{min2022voxel}, Min et al. adopted a range-aware random masking strategy and designed a binary voxel classification task, demonstrating that masking autoencoders can enhance 3D perception in autonomous driving.
In~\cite{hess2022masked}, Hess et al. pre-trained a Transformer-based 3D object detection backbone to recover obscured voxels and distinguish between free and occluded voxels, leading to improvements in 3D object detection performance.
%, leading to a 1.75 mAP and 1.05 NDS improvement in 3D object detection (OD) performance.

% \begin{figure}[ht]
%     \centering
%     \includegraphics[width=\linewidth]{Figures/reconstruction-based-SSL/Voxel-MAE-1.pdf}
%     \caption{----.}
%     \label{fig:Voxel-MAE-1}
% \end{figure}

% \begin{figure}[ht]
%     \centering
%     \includegraphics[width=\linewidth]{Figures/reconstruction-based-SSL/Voxel-MAE-2.pdf}
%     \caption{----.}
%     \label{fig:Voxel-MAE-2}
% \end{figure}

% \begin{figure*}[t]
%     \centering
%     \subfloat[\footnotesize{Voxel-MAE by Min et al.~\cite{min2022voxel}}]{
%     % \label{fig:local_feature}
%     \includegraphics[width=0.5\textwidth]{Figures/reconstruction-based-SSL/Voxel-MAE-1.pdf}}\subfloat[\footnotesize{Voxel-MAE by Hess et al.~\cite{hess2022masked}}]{
%     % \label{fig:local_feature}
%     \includegraphics[width=0.5\textwidth]{Figures/reconstruction-based-SSL/Voxel-MAE-2.pdf}}
%     \caption{The comparison of two Voxel-MAEs.}
%     \label{fig:Voxel-MAE}
% \end{figure*}

Krispel et al.~\cite{krispel2022maeli} introduced a Masked AutoEncoder for LiDAR point clouds (MAELi) that intuitively leverages the sparsity of LiDAR point clouds in both the encoder and decoder during the reconstruction process. 
MAELi distinguishes between empty and non-empty voxels and employs a novel masking strategy that targets LiDAR’s inherent spherical projection.
However, the irregular shape of MAE still poses challenges in large-scale 3D point cloud exploration. 
Yang et al.~\cite{yang2022gd} proposed a generative decoder (GD-MAE) to automatically merge the surroundings in a hierarchical fusion manner and recover the occluded geometric knowledge. GD-MAE achieves comparable accuracy on the Waymo dataset even with only 20$\%$ of labeled data and has been tested on several large-scale benchmarks. 

There are several works that adopt point cloud completion strategies, integrating object-level point cloud completion methods in the process.
For example, Ren et al.~\cite{ren2022self} proposed TraPCC, a self-supervised point cloud completion method that takes advantage of vehicle symmetry and similarity to create a vehicle memory bank using continuous point cloud frames. By concentrating on both local geometric details and global shape features of the input for point cloud completion, TraPCC achieves impressive performance on the KITTI and nuScenes datasets, even without the need for complete data as supervision. 
%The TraPCC pre-trained model demonstrates its effectiveness in 3D detection applications.
In point clouds obtained from autonomous driving scenarios, points of objects might be missing due to long distances and occlusions. Xie et al.~\cite {xie2022masked} proposed PCMAE that adopts a PC-Mask strategy, which can effectively recover partial objects from external occlusion and signal miss.
Through this pre-task, PCMAE improves the feature representation of 3D object detection backbones for long-distance and occluded objects through SSL.
% PCMAE ultimately achieves faster object detection with SECOND (Sparsely Embedded Convolutional Detection) and Part-A2-Net (Part-aware and Aggregate Neural Network) models in terms of convergence speed.

Contrasting with the object-level completion strategy employed for outdoor scene-level point clouds, some research focuses on utilizing semantic scene completion to reconstruct the entire scene directly.
Alexandre et al.~\cite{boulch2022also} proposed ALSO, a method that introduces a novel pre-task based on surface reconstruction. In this approach, occupancies are used for semantic scene completion.

Given that occupancy networks~\cite{mescheder2019occupancy} have been extensively studied in past several years and are widely used in companies like Tesla~\cite{shi2023grid}, it might be a promising way to pre-train 3D large-scale models based on occupancy prediction and transfer them to downstream tasks.
% This pre-task has a very simple formulation, making it easy to implement and widely applicable to a large range of 3D sensors.

\vspace{-0.2cm}
\subsubsection{BEV-based Methods} 
% \vspace{-0.1cm}
The Bird's Eye View (BEV) representation is a popular way to represent point clouds for 3D object detection and scene understanding tasks. 
It projects the point cloud onto a 2D grid from a top-down view, preserving the 3D point cloud's spatial information in a 2D format. This allows for easier processing of point clouds using 2D deep neural networks.
Several reconstruction-based methods have been designed using the BEV representation. 
For example, BEV-MAE~\cite{lin2022bev} introduces a BEV-guided mask strategy to guide the 3D encoder to learn feature representation in a BEV perspective and avoid the complicated design of the decoder during pre-training.
Moreover, a trainable point token is proposed to maintain a consistent receptive field size of the 3D encoder when fine-tuning for masked point cloud inputs.
% learn feature representation and employs learnable point labeling and point density prediction to learn location information.
% achieving improved results on different 3D object detectors from both Waymo and nuScenes. 
A more recent approach TPVformer~\cite{huang2023tri} uses the BEV-derived TPV (Top-View Projection) representation, which might be a promising and novel strategy for MAE.

% As a result, in the 1$\%$ labeled data, BEV-MAE exceeds the \textit{ab initio} training baseline of 6.05 APH on LEVEL 1 and 5.46 APH on LEVEL 2, demonstrating its potential in using large amounts of unlabeled data.
\vspace{-0.2cm}
\subsubsection{Challenges and Opportunities}
\vspace{-0.1cm}

Reconstruction-based self-supervised point cloud pre-training methods can efficiently obtain pre-trained models by reconstruction from large-scale point clouds collected by autonomous vehicles, thereby improving the perception ability of autonomous vehicles.
This method can effectively utilize the geometric information of the 3D scene and handle unordered and irregular point cloud data, making the feature representations of the pre-trained model more robust.
Due to the diversity and variability of autonomous driving scenes, this method can gradually adapt the pre-trained model to different scenes and situations, improving the generalization performance of the model.

However, reconstruction-based self-supervised pre-training methods require large computational and storage costs since they need to process a large number of point cloud data.
Because of the noise and missing data in the point cloud collection and reconstruction process, the feature representations of the pre-trained model may be disturbed, leading to a decline in the model performance.
This method relies on high-quality scene reconstruction models. 
Therefore, for some difficult scene reconstruction scenarios, it may be difficult to obtain high-quality pre-training models.
\vspace{-0.3cm}
\subsection{Contrastive Learning-based SSL}
\vspace{-0.1cm}
Compared to object or indoor scene-level point clouds, outdoor scene-level point clouds have larger sizes, noise, sparsity, complex weather, and lighting conditions. 
These factors can affect the effectiveness of self-supervised pre-training methods based on contrastive learning.
Therefore, the application of contrastive learning-based SSL in outdoor scene-level point clouds requires improvement based on the characteristics of outdoor scenes to improve the performance and generalization ability of the pre-trained models.
In this section, we mainly focus on contrastive learning at the outdoor scene-level point clouds.

\begin{figure}[t]
    \centering
    \includegraphics[width=\linewidth]{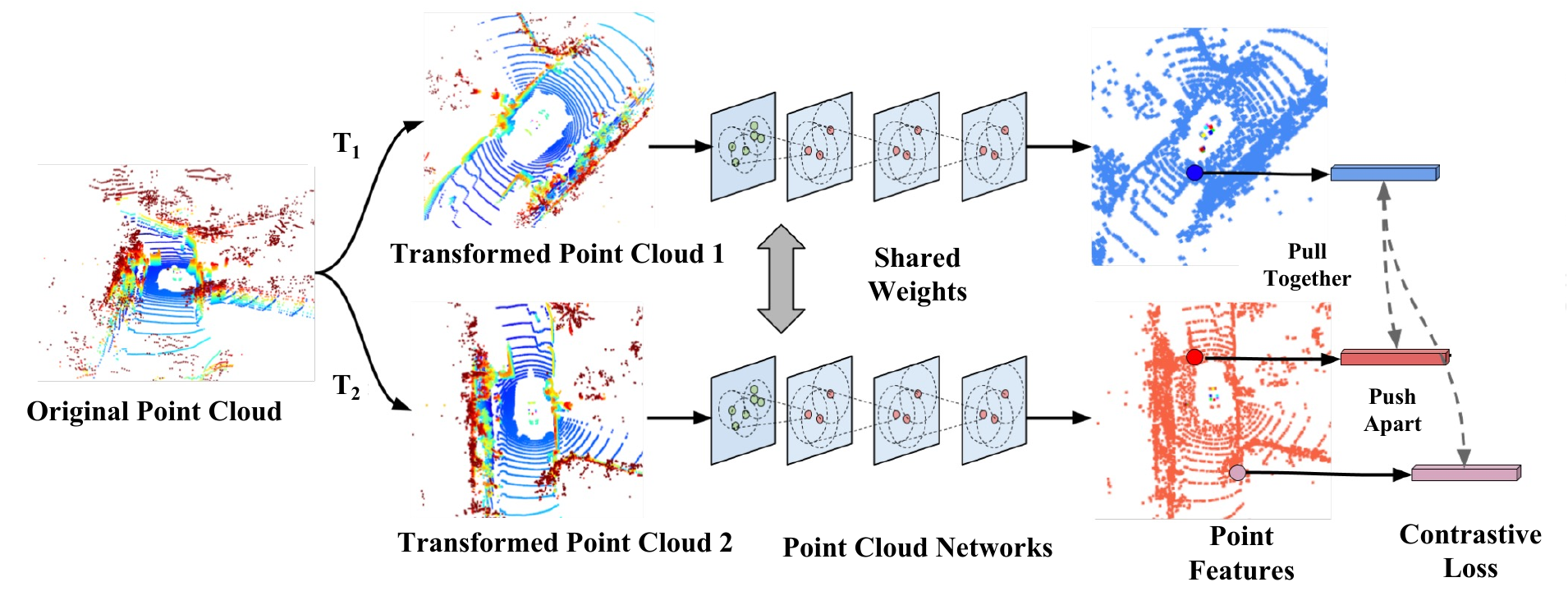}
    \vspace{-0.8cm}
    \caption{Illustration of scene-level contrastive learning-based methods. 
    The original point cloud undergoes a transformation to produce two transformed point clouds.
    These transformed point clouds are then fed into the shared-weight networks to obtain the point features.
    Aligning these features serves as the pre-task for contrastive-learning-based SSL.
    Images courtesy of Shi et al.~\cite{shi2022self}.}
    \label{contrast-scene}
\vspace{-0.3cm}
\end{figure}

\vspace{-0.2cm}
\subsubsection{View-based Methods}
\vspace{-0.1cm}
A series of existing self-supervised models like PointContrast~\cite{xie2020pointcontrast} and DepthContrast~\cite{zhang2021self} fail to be directly applied in outdoor autonomous driving scenarios for 3D object detection due to their statical partial view setting and lack of semantic information~\cite{liang2021exploring}. 
Therefore, Liang et al.~\cite{liang2021exploring} proposed the self-supervised learning framework, dubbed GCC-3D that integrates geometry-aware contrast and clustering harmonization without static partial views setting. 
Injecting the prior that spatially close voxels in point cloud tend to have similar local geometric structures, GCC-3D utilizes geometric distance to guide voxel-wise feature learning to alleviate the "class collision" problem inherent in hard labeling strategy. 
In the Harmonized Instance Clustering module, GCC-3D first generates pseudo instances by exploiting sequential information to localize moving objects. 
It further aggregates the voxel features obtained in the Geometric-Aware Contrast module as instance embedding to encode contextual semantic information. 
On this basis, a harmonization term is used to force different views of pseudo instances to be consistent with clustering prototype centers. 
% GCC-3D achieved state-of-the-art performance on many popular 3D object detection datasets, i.e., 67.29$\%$ mAP on Waymo and 57.3$\%$ mAP on nuScenes.
Moreover, Li et al.~\cite{li2022simipu} first studied a pre-training method SimIPU for outdoor multi-modal datasets of contrastive learning and devised a multi-modal contrastive learning pipeline composed of an intra-modal spatial perception and an inter-modal feature interaction to learn spatial-aware visual representations. 
However, it only focuses on the visual representations of spatial perception but ignores the semantic information. 

\vspace{-0.2cm}
\subsubsection{Region-based Methods}
\vspace{-0.1cm}
Scene-level contrastive learning methods are prone to loss of local details, and voxel-level methods fail to produce complete object representations due to limited receptive field and overemphasis on fine-grained features. 
Region-based methods~\cite{nunes2022segcontrast, yin2022proposalcontrast} are the trade-off methods mentioned above, making them more suitable for 3D object detection and semantic segmentation tasks in outdoor autonomous driving scenarios. 

To boost the performance of downstream semantic segmentation tasks, SegContrast pre-training~\cite{nunes2022segcontrast} first extracts class-agnostic segments from the point cloud and applies a segment-wise contrastive loss over the augmented pair of the sample class-agnostic segment to learn more contextualized information. 
Compared with other contrastive learning works, SegContrast exhibits significant superiority when using fewer labels, i.e., 1$\%$, producing a robust and fine-grained feature representation and can be well transferable between different datasets. 
For 3D object detection, Yin et al.~\cite{yin2022proposalcontrast} devised a two-stage proposal-level SSL framework called ProposalContrast, which learns point cloud representation by contrasting region proposals. 
Different from the convention in 2D SSL methods, ProposalContrast adopts spherical proposals instead of bounding box proposals considering the enlarged space in 3D scenarios. 
In the regional proposal encoding module, cross-attention is used to explicitly capture the geometric relations among points inside each proposal. 
Then, two pretext tasks, inter-proposal discrimination, and inter-cluster separation are optimized jointly to better meet the need for 3D object detection. 
% The framework is first pre-trained on Waymo Open Dataset and then transferred to KITTI for fine-tuning. 
% Experimental results show that ProposalContrast outperforms several concurrent SSL methods, i.e., DepthContrast~\cite{zhang2021self}, PointContrast~\cite{xie2020pointcontrast}, GCC-3D~\cite{liang2021exploring}, and STRL~\cite{huang2021spatio}. 

\vspace{-0.2cm}
\subsubsection{Multi-view Methods} 
\vspace{-0.1cm}
Contrastive learning highly relies on corresponding views generated by anchor data using methods like data augmentation or extracting from different timestamps. 
However, views produced by such methods are either too similar or hard to find correct correspondences.
The other challenge is unable to produce views that differ enough but still share abundant common semantic information, hampering the model performance in downstream tasks\cite{chen2022co}. 
To overcome these limitations, Chen et al. \cite{chen2022co} utilized DAIR-V2X dataset to build views from both the vehicle side and the infrastructure side as contrastive samples.

\vspace{-0.2cm}
\subsubsection{Combining with Other Pre-tasks} 
\vspace{-0.1cm}
Although contrastive learning has made significant progress, relying solely on it for pre-training may not be sufficient, as the learned representations may not capture task-related information effectively~\cite{wang2022rethinking}. To address this limitation, Chen et al.~\cite{chen2022co} introduced a shape context prediction task that reconstructs local distribution, providing more relevant information to improve downstream 3D detection tasks.
Furthermore, Shi et al.~\cite{shi2022self} demonstrated that pre-training with only contrastive loss could negatively impact the accuracy of object heading estimation. To counter this issue, they combined contrastive learning with a set of geometric pre-tasks, specifically observation angle difference recognition and relative scaling recognition. These additional pre-tasks help enhance the model's performance by capturing more informative representations, ultimately leading to improved results in downstream tasks.

% Experiments on the nuScenes dataset show that this method improves the average precision and the object heading accuracy at the same time.

\vspace{-0.2cm}
\subsubsection{Challenges and Opportunities}
\vspace{-0.1cm}
A main benefit of outdoor scene-level contrastive learning is that it can be applied to various types of sensors used in autonomous driving, such as LiDAR and RGB-D data. This capability improves the compatibility and versatility of the pre-trained models for different sensor types.

However, there are several challenges associated with contrastive learning-based self-supervised pre-training methods. First, these methods require careful tuning of hyperparameters, such as the contrastive loss margin and batch size. Finding the optimal settings can be time-consuming and computationally expensive.
Second, contrastive learning methods may not fully capture the semantic information of the scene, potentially limiting the model's ability to perform high-level scene understanding and decision-making tasks.

% \begin{itemize}
%     \item Contrastive learning-based self-supervised point cloud pre-training methods can be applied to various types of autonomous driving sensors such as LiDAR and RGB-D data, which can improve the compatibility and versatility of the pre-training model for different types of sensors.
%     \item Contrastive learning-based self-supervised point cloud pre-training methods require careful tuning of hyperparameters such as the contrastive loss margin and batch size, which can be time-consuming and computationally expensive.
%     \item The quality of the pre-training data can affect the performance of the model, and it can be challenging to obtain diverse and representative point cloud data for pretraining.
%     \item The contrastive learning-based self-supervised point cloud pre-training method may not fully capture the semantic information of the scene, which can limit the model's ability to perform high-level scene understanding and decision-making tasks.
% \end{itemize}

\vspace{-0.3cm}
\subsection{Multi-modality SSL}
\vspace{-0.1cm}
\begin{figure}[t]
    \centering
    \includegraphics[width=0.9\linewidth]{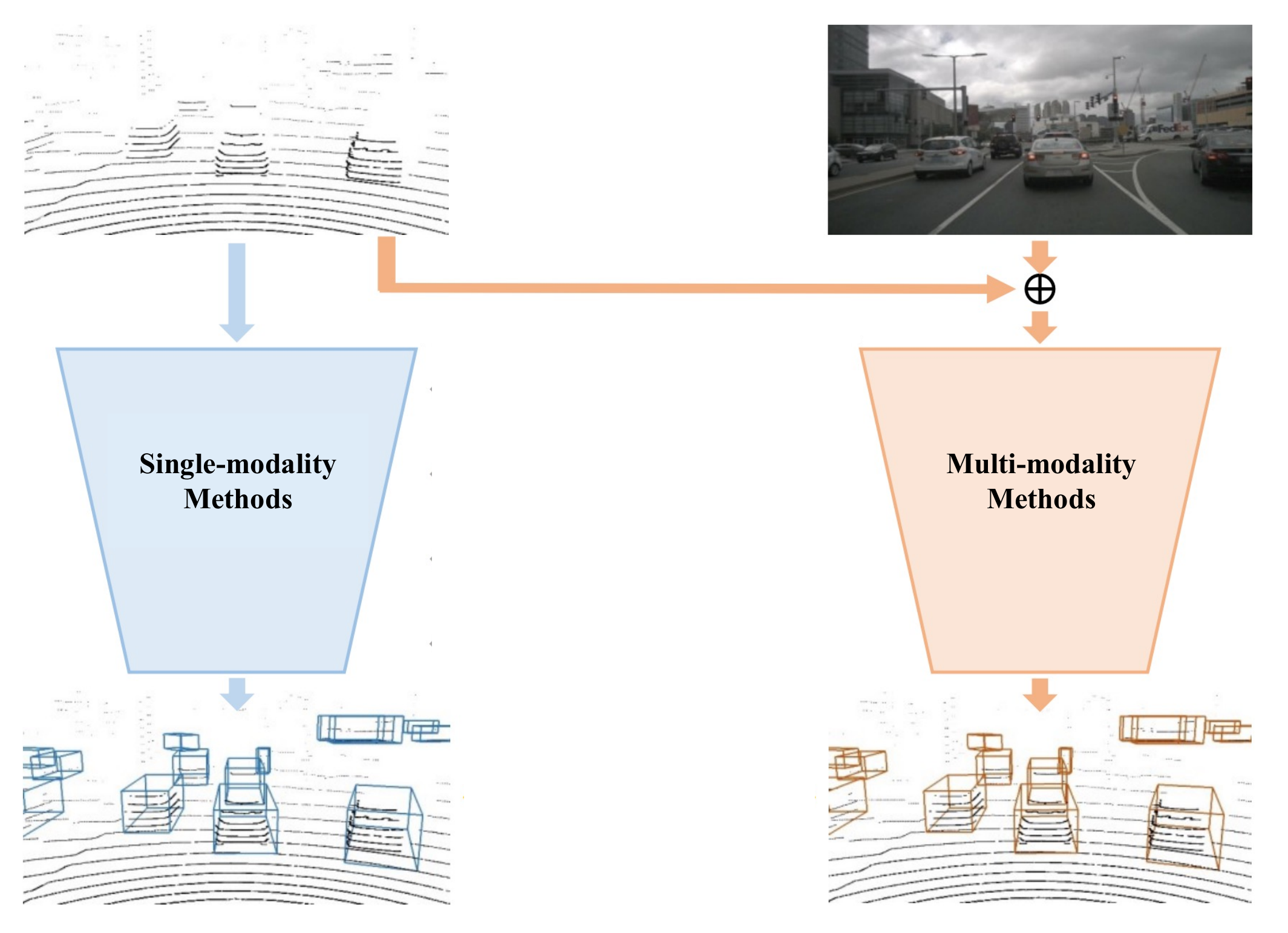}
    \vspace{-0.5cm}
    \caption{Comparison of multi-modality methods with single-modality methods. 
    Single-modality methods rely solely on point clouds for self-supervised learning, while multi-modality methods integrate images or range images to enhance SSL. 
    Images courtesy of Zheng et al.~\cite{zheng2022boosting}. }
    \label{multi-modality-scene}
\vspace{-0.3cm}
\end{figure}

The sparsity of point clouds increases with distance due to laser beam divergence, making it very difficult to predict the boundaries and semantic classes of small and distant objects. Combining multiple sensors such as LiDAR and cameras can provide complementary information that enhances the overall robustness of autonomous driving systems. The use of high-resolution 2D images from cameras allows the system to better handle small and distant objects, which are challenging to detect and classify with LiDAR data alone (Figure~\ref{multi-modality-scene}). However, the acquisition and processing of multi-modality data for high-quality data fusion are extremely tedious. While higher precision can often be attained, multi-modality detectors inevitably sacrifice inference efficiency to process the extra modality~\cite{zheng2022boosting}.

To address the challenge of efficiently leveraging multi-modal data, Zheng et al.~\cite{zheng2022boosting} presented a one-stage framework called S2M2-SSD, which combines four different levels of knowledge distillation to guide the single-modal network to generate simulated multi-modal features. This method only takes multi-modal data as input in the training stage, thus capable of achieving excellent performance with high efficiency, precision, and robustness during inference. 
In the S2M2-SSD framework, response distillation and sparse-voxel distillation are incorporated with a crucial response mining strategy to concentrate on important responses and avoid irrelevant background information for improved computational efficiency. 
For objects with sparse points or small sizes, which may be difficult to represent accurately with voxel features, S2M2-SSD adopts a voxel-to-point distillation technique that transforms coarse-grained voxel features into fine-grained point features through interpolation. Following the transformation, the method performs point-wise distillation to further enhance the network's ability of detection. 
Finally, S2M2-SSD ensures instance-level consistency in deep-layer features by incorporating an instance distillation process, which learns deep-layer BEV features in the Non-Maximum Suppression (NMS)-filtered bounding boxes. 
When tested on the nuScenes dataset, S2M2-SSD outperforms other single-modality 3D detectors and even exceeds the baseline LiDAR-image detector on the nuScenes detection score (NDS) metrics.

%\subsubsection{Challenges and Opportunities}
While the success of S2M2-SSD demonstrates the potential benefits of incorporating multi-modal data during the training process, there are still several technical challenges for outdoor scene-level point clouds that need to be addressed. For example, integrating and processing data from different sensors can be difficult, as they often have diverse data formats, sampling rates, and resolutions. Furthermore, aligning multi-modal data to a common coordinate system is essential yet poses its own set of challenges, particularly in real-world scenarios characterized by dynamic and unstructured environments.

\vspace{-0.3cm}
\subsection{Prediction-based SSL}
\vspace{-0.1cm}

\begin{figure}[t]
    \centering
    \includegraphics[width=0.9\linewidth]{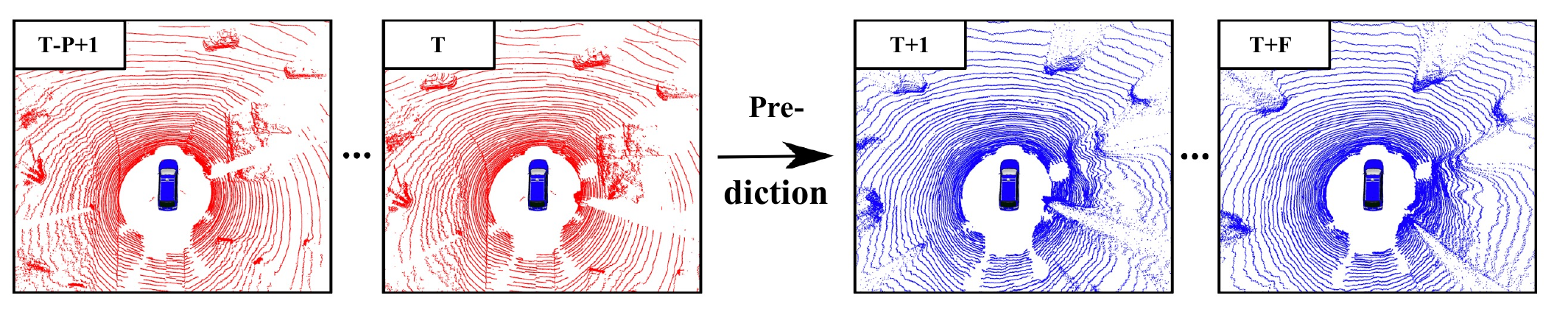}
    \vspace{-0.5cm}
    \caption{Illustration of prediction-based methods. 
    Given a sequence of $P$ past point clouds (left in red) at time $T$, the objective is to predict the $F$ future scans (right in blue).
    The prediction of future scans can serve as a pre-task for self-supervised learning.
    Images courtesy of Mersch et al.~\cite{mersch2022self}. }
    \label{prediction}
\vspace{-0.3cm}
\end{figure}

Point cloud prediction assists vehicles in improving their decision-making for tasks such as path planning and collision avoidance. Since the ground truth is inherently provided in the subsequent frame of the LiDAR scan, it can be trained in an SSL manner without the need for costly labeling, making it a promising method for autonomous driving applications. 

Range image- and vision-based prediction approaches have been extensively studied for predicting future point clouds from a sequence of past LiDAR scans.
Methods such as those in \cite{weng2021inverting} and \cite{lu2021monet} utilize RNN to model the temporal correlations, while methods in \cite{hoermann2018dynamic}, \cite{song20192d}, \cite{wu2020motionnet}, and \cite{toyungyernsub2021double} focus on estimating voxelized point clouds.

\vspace{-0.2cm}
\subsubsection{RNN-based Methods}
\vspace{-0.1cm}
One example of RNN-based methods involves using a multi-layer RNN to predict the next point in a sequence of points created by a fast space-filling curve (Morton-order curve). 
The final RNN state, also known as Morton features~\cite{thabet2020self}, has been found to be general and exhibits improved performance in semantic segmentation. Furthermore, these features can be transferred from self-supervised network to other large-scale datasets, such as vKITTI.

\vspace{-0.2cm}
\subsubsection{Range Image-based Methods}
\vspace{-0.1cm}
Mersch et al. ~\cite{mersch2022self} proposed an auto-encoder architecture using 3D convolutions to jointly process spatial and temporal information and predict the full-scale point clouds without the necessity for voxelization.
Before being fed into the encoder-decoder CNN, past point clouds are first projected into 2D range images and then concatenated as a spatial-temporal tensor. 
To main details and spatial consistency, both skip connections and horizontal circular padding are employed during convolutions.

% \fb{i will check this}
\vspace{-0.2cm}
\subsubsection{Combining with Other Pre-tasks}
\vspace{-0.1cm}
As mentioned in Section~\ref{CL-based-methods}, contrastive learning alone is unable to capture task-related information for downstream tasks\cite{chen2022co}. 
To address this limitation, a shape context prediction task is introduced in the CO$^{\wedge}$3 framework\cite{chen2022co}. 
Reconstructing the entire scene with voxel-level representation is challenging, hence, the goal of this pre-training task is to predict the local distribution of each voxel in the point cloud using a shape context descriptor. 
As a result, CO{$^{\wedge}$}3 can incorporate more task-relevant information and achieve excellent transferability across different datasets.

\vspace{-0.2cm}
\subsubsection{Challenges and Opportunities}
\vspace{-0.1cm}

Predictive-based SSL methods are suitable for outdoor autonomous driving, as point cloud frames are continuous and predictable in these environments. 
However, relatively few studies have been carried out in this area.

Predictive-based SSL can be further improved by adopting point-level prediction. Current voxel-level prediction methods result in the loss of information, but a point-level prediction might be more computationally complicated than a voxel-level prediction, necessitating a trade-off between the two variants.

As simulation engines continue to advance rapidly, obtaining simulated continuous point cloud frames has become more accessible. Pre-training models on these simulated point clouds using predictive-based self-supervised learning could be a promising avenue for future research.

% The summaries and discussion are listed as follows:
% \begin{itemize}
%     \item Predictive-based self-supervised learning is suitable for outdoor autonomous driving, which owns intrinsic characteristics that point cloud frames are continuous and predictable. However, relatively few studies have been carried out in this aspect.
%     \item Predictive-based self-supervised learning can be further promoted by point-level prediction, while the current voxel-level prediction will result in the loss of information. But the point-level prediction might be more computationally complicated than a voxel-level prediction, determining the trade-off between two variants.
%     \item With the rapid development of simulation engines, simulated continuous point cloud frames are more easily to be obtained. Pre-training the model on these simulated point clouds with predictive-based self-supervised learning might be a promising way.
% \end{itemize}

\vspace{-0.3cm}
\subsection{Flow-based SSL}
\vspace{-0.1cm}
\begin{figure}[t]
    \centering
    \includegraphics[width=0.9\linewidth]{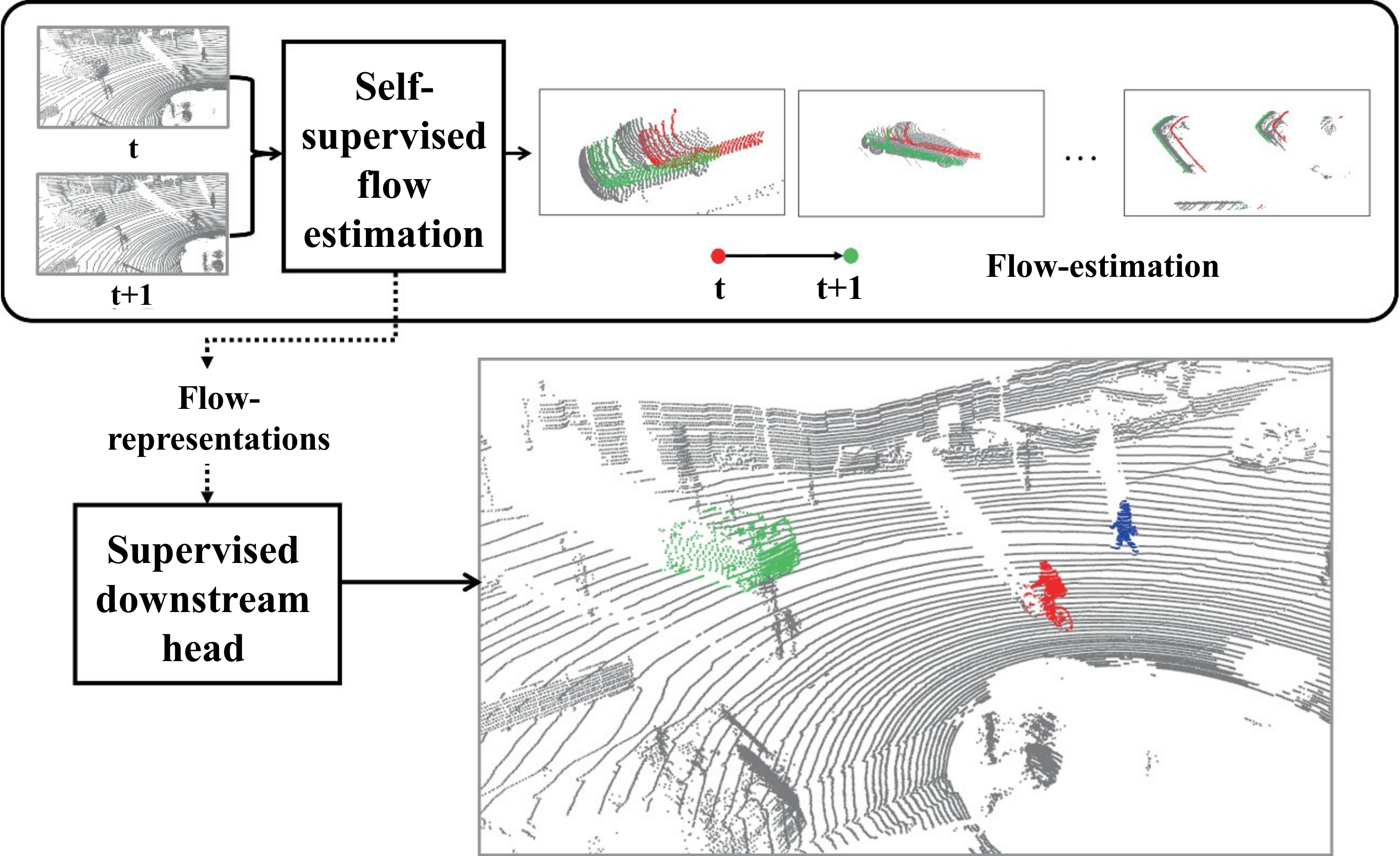}
    \vspace{-0.4cm}
    \caption{Illustration of flow-based methods.
    The flow estimation task can be considered as a pre-task for self-supervised learning, and the pre-trained model can be readily transferred to supervised downstream tasks.
    Images courtesy of Erccelik et al.~\cite{erccelik20223d}. }
    \label{flow}
\vspace{-0.5cm}
\end{figure}

Scene flow refers to the relative motion of each 3D point in a temporal sequence of point clouds. Scene flow estimation is a significant topic in the field of autonomous driving, as it supports safe planning and navigation by helping autonomous vehicles perceive the actions of surrounding entities. 

\vspace{-0.2cm}
\subsubsection{Scene Flow Estimation}
\vspace{-0.1cm}
Most state-of-the-art scene flow estimation methods are trained with synthetic data and then fine-tuned on real-scene datasets due to the limited amount of labeled real-world data. 
However, it has been suggested in~\cite{liu2019flownet3d} that significant improvements can be fulfilled by training on real-world data from the domain of the target application. 
Motivated by this, Mittal et al.~\cite{mittal2020just} devised a self-supervised training approach for scene flow estimation that incorporates two self-supervised losses on unlabeled datasets. 
% The endpoint error used as the evaluation metric in supervised loss is unavailable in self-supervised learning for the lack of ground truth flow. 
% Therefore, ~\cite{mittal2020just} replaced such loss with nearest neighbor loss, which refers to the mean Euclidean distance between the estimated point and its nearest neighbor. 
The nearest neighbor loss is utilized to measure the mean Euclidean distance between the estimated point and its nearest neighbor. 
To avoid network degeneration, this method also introduces a cycle consistency loss, which equals the error between the original point and the predicted point after the reverse flow. 
Moreover, errors produced in the estimated reverse flow can lead to structural distortions in the transformed point cloud, so anchoring points are added to help stabilize the transformed cloud and retain useful structural information. 
Results show that this method matches the performance of supervised methods and even surpasses them on a smaller annotated dataset. 

Although the method proposed by~\cite{mittal2020just} achieves competitive performance against supervised methods, it only uses 3D point coordinates as the measure in point-wise matching, overlooking other discriminative measures such as colors and surface normals~\cite{li2021self}. 
Additionally, the matching is operated in an unconstrained condition, which may lead to a degeneration solution, i.e., a many-to-one problem. 
To address these issues, Li et al.~\cite{li2021self} formulated the scene flow task as an optimal transportation problem under the constraint of one-to-one matching, considering coordinates, colors, and surface normals to compute the matching cost. 
It generates pseudo labels by deriving an optimal assignment matrix with the Sinkhorn algorithm~\cite{altschuler2017near}. 
However, the label generation process neglects the local relation among neighboring points and is prone to producing conflicting labels. 
Thus, a random walk module is also introduced to strengthen the local consistency. 
The models trained on the generated pseudo labels achieve state-of-the-art performance on the FlyingThings3D and KITTI datasets.

Most previous works~\cite{mittal2020just, li2021self, baur2021slim, kittenplon2021flowstep3d, pontes2020scene, tishchenko2020self, wu2020pointpwc} follow the point-wise matching between two consecutive point clouds to generate pseudo labels. 
However, these point-matching strategies focus only on local similarities and fail to capture the latent structural motions of objects, resulting in local inconsistency in pseudo labels~\cite{li2022rigidflow}. 
Based on the assumption that scene flow can be divided into a series of rigid motions of individual parts, Li et al.~\cite{li2022rigidflow} developed a piecewise rigid motion estimation method, called RigidFlow. It first applies an over-segmentation approach to decompose the point cloud into supervoxels and predict the rigid flow for each supervoxel. It then forms the pseudo labels for the entire scene flow using the generated pseudo labels for local regions. 
% On FlyingThing3D with no occlusion and KITTI, RigidFlow beats five self-supervised methods, i.e., Ego-motion~\cite{tishchenko2020self}, PointPWC-Net~\cite{wu2020pointpwc}, SLIM~\cite{baur2021slim}, Self-Point-Flow~\cite{li2021self} and FlowStep3D~\cite{kittenplon2021flowstep3d}.

The existence of occlusions challenges the accuracy of scene flow estimation, as corresponding points are likely to be masked in the target point cloud. Therefore, it has been proposed by~\cite{zhao2020maskflownet,saxena2019pwoc, ouyang2021occlusion} to exclude the occluded areas before the cost volume construction, improving the performance at the cost of harming the flow approximation accuracy for the occluded areas. 
To overcome this dilemma, 3D-OGFlow~\cite{ouyang2021occlusion} merges two networks across all layers to conduct flow estimation and learn the occlusions simultaneously. 
For finer scene flow, an occlusion-weighted cost volume layer is added to each level, constructing cost volumes for the occluded and non-occluded regions separately, and then aggregating them in an occlusion-weighted manner.

\vspace{-0.2cm}
\subsubsection{Flow for Detection and Segmentation}
\vspace{-0.1cm}
Scene flow estimation can be leveraged as a pre-training method to improve the performance of downstream tasks related to autonomous driving. 
Yurtsever et al.~\cite{erccelik20223d} used learned scene flow and motion representation to guide the 3D object detection tasks. 
Drawing inspiration from~\cite{mittal2020just}, they adopted the cycle consistency approach to train the self-supervised backbone network and scene flow head based on FlowNet3D~\cite{liu2019flownet3d}. 
The learned motion representations aid downstream 3D detectors in recognizing objects based on their moving patterns.
Subsequently, the pre-trained backbone and the 3D detection head are fine-tuned on a smaller labeled dataset. 
% On KITTI and nuScenes datasets, pre-trained Point-GNN~\cite{shi2020point}, CenterPoint~\cite{yin2021center} and PointPillars~\cite{lang2019pointpillars} all exceed their baseline models. 

% \subsubsection{Flow for Segmentation}
Scene flow estimation has also been introduced into the field of object segmentation by Song and Yang~\cite{song2022ogc} through their self-supervised method OGC. OGC utilizes learned dynamic motion patterns as supervision signals to train the network to simultaneously segment multiple objects in a single forward pass. 
Different from traditional motion segmentation methods that take sequential point clouds as input, OGC incorporates multi-object rigid consistency and object shape invariance into the loss functions for high-quality segmentation.

\vspace{-0.2cm}
\subsubsection{Challenges and Opportunities}
\vspace{-0.1cm}
Flow-based self-supervised point cloud pre-training methods can effectively capture the temporal information of point clouds collected by autonomous driving vehicles, enhancing performance in motion prediction and dynamic scene understanding. Additionally, these methods can generate synthetic training data for pre-training models, increasing  the model's robustness to various dynamic and complex driving scenarios.

However, flow-based methods necessitate precise motion estimation and registration of point clouds, which can be challenging in scenarios involving high-speed and complex motion. These methods also rely on the accuracy of the optical flow estimation algorithm, which can be affected by factors such as occlusion, lighting, and sensor noise.
Furthermore, flow-based methods may not fully capture the semantic information of the scene, potentially limiting the model's capability to perform high-level scene understanding and decision-making tasks.

% \begin{itemize}
%     \item Flow-based self-supervised point cloud pre-training methods can effectively capture the temporal information of point cloud data collected by autonomous driving vehicles, which can improve the performance of motion prediction and dynamic scene understanding.
%     \item Flow-based self-supervised point cloud pre-training methods can generate synthetic training data for pre-training models, which can improve the model's robustness to various dynamic and complex driving scenarios.
%     \item Flow-based self-supervised point cloud pre-training methods can be applied to both LiDAR and RGB-D data, which can improve the compatibility and versatility of the pre-training model for different types of autonomous driving sensors.
%     \item Flow-based self-supervised point cloud pre-training methods require accurate motion estimation and registration of point cloud data, which can be challenging in some scenarios with high-speed and complex motion.
%     \item The flow-based self-supervised point cloud pre-training method relies on the accuracy and reliability of the optical flow estimation algorithm, which can be affected by various factors such as occlusion, lighting, and sensor noise.
%     \item The flow-based self-supervised point cloud pre-training method may not fully capture the semantic information of the scene, which can limit the model's ability to perform high-level scene understanding and decision-making tasks.
% \end{itemize}

% !TEX root = ../bare_jrnl_new_sample4.tex
\vspace{-0.3cm}
\section{Discussions and Future Directions}
\vspace{-0.1cm}
\label{future_work}

Self-supervised learning has shown great potential for pre-training point cloud data. However, several challenges remain because of the complex structures and diverse tasks of point clouds. In this section, we will discuss these challenges and potential directions for future research.

\vspace{-0.3cm}
\subsection{Unified Backbone Design}
\vspace{-0.1cm}
One of the main reasons why deep learning has achieved great success in NLP and 2D computer vision is the standardization of architectures such as BERT and GPT in NLP and VGG and ResNet in 2D computer vision. The use of a unified backbone design greatly facilitates knowledge transfer between various datasets and tasks. However, for 3D point clouds, although various 3D architectures have been designed in the past few years, the development of similar unified 3D backbones is far from fully explored.
Most current backbone models are significantly different from each other (as shown in Tables I and II in the Supplementary Material), hindering the development of 3D point cloud networks in terms of scalable design, efficient deployment in various practical applications, etc. In 3D vision, the development of common backbones as ubiquitous as BERT, GPT, and ResNet is vital for the advancement of 3D point cloud networks, including self-supervised point cloud representation learning. 

\vspace{-0.3cm}
\subsection{High-quality Pre-training Datasets}
\vspace{-0.1cm}
Most existing self-supervised pre-training datasets were originally collected for supervised learning tasks, as shown in Tables I and II in the Supplementary Material. 
However, due to the time-consuming and laborious nature of point cloud annotation, these datasets are severely limited in terms of data amount and diversity and are not suitable for self-supervised point cloud representation learning, typically requiring a large number of point clouds with diverse and abundant information. 
These limitations of existing datasets largely explain the negligible performance improvement of self-supervised pre-training. To address this critical problem, there is a pressing need to collect sufficiently diverse large-scale high-quality point cloud datasets that cover object- and scene-level data.

\vspace{-0.3cm}
\subsection{Standardized Downstream Tasks}
\vspace{-0.1cm}
Standardization of downstream tasks is essential for evaluating the effectiveness of self-supervised pre-training. At present, there are issues with the setting of downstream tasks: (1) While there is a unified standard for downstream tasks at the object level for synthetic objects (ModelNet40), real object classification (ScanObjectNN), few-shot classification (Few-shot ModelNet40), and synthetic object segmentation (ShapeNetPart), these tasks are difficult to generalize to other datasets (see Figure I-VII in Appendix). 
Therefore, there is a need to collect datasets that help pre-trained models to transfer to real scenes.  
(2) For indoor-level detection (ScanNet) and segmentation (S3DIS \& SUN RGB-D), there is a lack of a unified model and framework (see Figure VIII-XI in the Appendix). 
(3) Downstream tasks for the outdoor scenes are more diverse, there is an urgent need for a unified evaluation standard to ensure a fair comparison.

\vspace{-0.3cm}
\subsection{Further Promotion of Self-supervised Pre-training for Scene-level Tasks}
\vspace{-0.1cm}
Most of the existing research in point cloud processing has focused on object-level point clouds (Table I in the Supplementary Material). However, there have been some pioneering studies exploring self-supervised pre-training on scene-level point clouds. For examples, Voxel-MAE~\cite{min2022voxel} and BEV-MAE~\cite{lin2022bev} are two methods that pre-train deep neural networks on scene-level point clouds to improve various downstream tasks such as 3D object detection and 3D instance segmentation. 

Previous work has shown that learned self-supervised representations can generalize effectively across domains and tasks (Table II in the Supplementary Material). 
Therefore, scene-level point cloud self-supervised learning, as a new direction, has great potential in various applications and deserves more attention. 
However, there are still challenges associated with network architectures and datasets that need to be addressed.
Additionally, the high cost of labeling autonomous driving scenes has led to the exploration of self-supervised pre-training as a solution to reducing annotation efforts and facilitating transfer learning to new scenarios. 

% We foresee more useful research on unsupervised learning on scene-level point clouds.

\vspace{-0.3cm}
\subsection{Integration of Multi-modal Data}
\vspace{-0.1cm}
Self-supervised pre-training methods can be extended to incorporate multi-modal data such as images, videos, and audio to improve the robustness and accuracy of the models. Nowadays, multi-modality SSL methods can be divided into two main streams.
On the one hand, multi-model data can be utilized to exploit multi-modal features, which are aligned by contractive learning, distillation, and attention. 
Although ULIP~\cite{xue2022ulip} leverages three modalities (images, texts, and point clouds), it is still possible to explore a unified framework in which texts, images, depth images, RGB-D images, and point clouds, and even meshes are all considered.
On the other hand, 2D pre-trained knowledge can be transferred to point cloud understanding, such as CLIP~\cite{zhu2022pointclip} or 2D transformer~\cite{vaswani2017attention}. 
To bridge the gap between images and point clouds, text-to-3D generation models are also promising future works. A pioneering work in this direction is Point-E~\cite{nichol2022point}, whose knowledge might be applied to pre-training point clouds.

\vspace{-0.3cm}
\subsection{Incorporation of Temporal Information}
\vspace{-0.1cm}
With the increasing availability of unlabeled sequences of point clouds from autonomous vehicles and intelligent robots, self-supervised pre-training methods can be devised to utilize the temporal information of point cloud data to enhance the model's ability to capture dynamic scenes and improve the performance of motion-related tasks.
While most existing SSL works focus on static point clouds, point cloud streams can provide rich temporal information used as useful supervision signals. Therefore, there is a need for more effective pre-tasks that can learn temporal information from unlabelled sequential point cloud frames, and we anticipate the development of such methods in the future.

\vspace{-0.3cm}
\subsection{Incorporation of Higher-level Semantic Information} 
\vspace{-0.1cm}
The incorporation of higher-level semantic information into point cloud SSL is a potential research direction. 
For example, semantic segmentation and object detection techniques can be used to annotate point clouds, and these labels can be used as supervision signals for pre-training tasks. 
As a pioneering method, the conditional completion in Relationship-Based Point Cloud Completion~\cite{zhao2021relationship} can produce pairwise scenes with better spatial relationships, which might be utilized as a pre-task for self-supervised pre-training methods.
More recently, methods such as SAM~\cite{kirillov2023segment} and SegGPT~\cite{wang2023seggpt} have been proposed to generate point cloud annotations.
Additionally, existing deep learning frameworks can be utilized to extract semantic information from point clouds and use it for pre-training tasks. 
These methods can further improve the performance of point cloud representation learning while providing richer semantic information for downstream tasks.
In general, self-supervised pre-training methods can also be designed to incorporate higher-level semantic information such as object attributes and relationships to improve the model's ability to perform high-level scene understanding and decision-making tasks.

% !TEX root = ../bare_jrnl_new_sample4.tex
\vspace{-0.3cm}
\section{conclusion}
\vspace{-0.1cm}
Despite recent successes in natural language processing and computer vision, self-supervised learning applied to point cloud data remains an emerging field with significant challenges to be addressed. Existing methods employ self-supervision in neural networks, such as contrastive learning, predictive learning, or multi-modality learning. 

This paper provides a comprehensive and up-to-date overview of self-supervised learning methods based on deep neural networks. We summarize existing SSL methods and provide a unified review of them in terms of datasets, evaluation metrics, and performance comparisons, while also discussing the challenges and potential future directions for the field. We also provide a comparative summary of methods, which can benefit researchers in the 3D computer vision community.

\ifCLASSOPTIONcaptionsoff
  \newpage
\fi



%\bibliographystyle{IEEEtran}


% \end{thebibliography}
\vspace{-0.3cm}
\bibliographystyle{IEEEtran}\balance
\bibliography{ref_abb}

\begin{thebibliography}{100}
\providecommand{\url}[1]{#1}
\csname url@samestyle\endcsname
\providecommand{\newblock}{\relax}
\providecommand{\bibinfo}[2]{#2}
\providecommand{\BIBentrySTDinterwordspacing}{\spaceskip=0pt\relax}
\providecommand{\BIBentryALTinterwordstretchfactor}{4}
\providecommand{\BIBentryALTinterwordspacing}{\spaceskip=\fontdimen2\font plus
\BIBentryALTinterwordstretchfactor\fontdimen3\font minus
  \fontdimen4\font\relax}
\providecommand{\BIBforeignlanguage}[2]{{%
\expandafter\ifx\csname l@#1\endcsname\relax
\typeout{** WARNING: IEEEtran.bst: No hyphenation pattern has been}%
\typeout{** loaded for the language `#1'. Using the pattern for}%
\typeout{** the default language instead.}%
\else
\language=\csname l@#1\endcsname
\fi
#2}}
\providecommand{\BIBdecl}{\relax}
\BIBdecl

\bibitem{xiao2022unsupervised}
A.~Xiao, J.~Huang, D.~Guan, and S.~Lu, ``Unsupervised representation learning
  for point clouds: A survey,'' \emph{arXiv:2202.13589}, 2022.

\bibitem{cui2021deep}
Y.~Cui, R.~Chen, W.~Chu, L.~Chen, D.~Tian, Y.~Li, and D.~Cao, ``Deep learning
  for image and point cloud fusion in autonomous driving: A review,''
  \emph{IEEE TITS}, vol.~23, no.~2, pp. 722--739, 2021.

\bibitem{guo2020deep}
Y.~Guo, H.~Wang, Q.~Hu, H.~Liu, L.~Liu, and M.~Bennamoun, ``Deep learning for
  3d point clouds: A survey,'' \emph{IEEE TPAMI}, vol.~43, no.~12, pp.
  4338--4364, 2020.

\bibitem{chang2015shapenet}
A.~X. Chang, T.~Funkhouser, L.~Guibas, P.~Hanrahan, Q.~Huang, Z.~Li,
  S.~Savarese, M.~Savva, S.~Song, H.~Su \emph{et~al.}, ``Shapenet: An
  information-rich 3d model repository,'' \emph{arXiv:1512.03012}, 2015.

\bibitem{afham2022crosspoint}
M.~Afham, I.~Dissanayake, D.~Dissanayake, A.~Dharmasiri, K.~Thilakarathna, and
  R.~Rodrigo, ``Crosspoint: Self-supervised cross-modal contrastive learning
  for 3d point cloud understanding,'' in \emph{CVPR}, 2022, pp. 9902--9912.

\bibitem{wu20153d}
Z.~Wu, S.~Song, A.~Khosla, F.~Yu, L.~Zhang, X.~Tang, and J.~Xiao, ``3d
  shapenets: A deep representation for volumetric shapes,'' in \emph{CVPR},
  2015, pp. 1912--1920.

\bibitem{uy2019revisiting}
M.~A. Uy, Q.-H. Pham, B.-S. Hua, T.~Nguyen, and S.-K. Yeung, ``Revisiting point
  cloud classification: A new benchmark dataset and classification model on
  real-world data,'' in \emph{ICCV}, 2019, pp. 1588--1597.

\bibitem{dai2017scannet}
A.~Dai, A.~X. Chang, M.~Savva, M.~Halber, T.~Funkhouser, and M.~Nie{\ss}ner,
  ``Scannet: Richly-annotated 3d reconstructions of indoor scenes,'' in
  \emph{CVPR}, 2017, pp. 5828--5839.

\bibitem{song2015sun}
S.~Song, S.~P. Lichtenberg, and J.~Xiao, ``Sun rgb-d: A rgb-d scene
  understanding benchmark suite,'' in \emph{CVPR}, 2015, pp. 567--576.

\bibitem{armeni20163d}
I.~Armeni, O.~Sener, A.~R. Zamir, H.~Jiang, I.~Brilakis, M.~Fischer, and
  S.~Savarese, ``3d semantic parsing of large-scale indoor spaces,'' in
  \emph{CVPR}, 2016, pp. 1534--1543.

\bibitem{geiger2013vision}
A.~Geiger, P.~Lenz, C.~Stiller, and R.~Urtasun, ``Vision meets robotics: The
  kitti dataset,'' \emph{IJRR}, vol.~32, no.~11, pp. 1231--1237, 2013.

\bibitem{behley2019semantickitti}
J.~Behley, M.~Garbade, A.~Milioto, J.~Quenzel, S.~Behnke, C.~Stachniss, and
  J.~Gall, ``Semantickitti: A dataset for semantic scene understanding of lidar
  sequences,'' in \emph{ICCV}, 2019, pp. 9297--9307.

\bibitem{pan2020semanticposs}
Y.~Pan, B.~Gao, J.~Mei, S.~Geng, C.~Li, and H.~Zhao, ``Semanticposs: A point
  cloud dataset with large quantity of dynamic instances,'' in \emph{IEEE
  IV}.\hskip 1em plus 0.5em minus 0.4em\relax IEEE, 2020, pp. 687--693.

\bibitem{sun2020scalability}
P.~Sun, H.~Kretzschmar, X.~Dotiwalla, A.~Chouard, V.~Patnaik, P.~Tsui, J.~Guo,
  Y.~Zhou, Y.~Chai, B.~Caine \emph{et~al.}, ``Scalability in perception for
  autonomous driving: Waymo open dataset,'' in \emph{CVPR}, 2020, pp.
  2446--2454.

\bibitem{caesar2020nuscenes}
H.~Caesar, V.~Bankiti, A.~H. Lang, S.~Vora, V.~E. Liong, Q.~Xu, A.~Krishnan,
  Y.~Pan, G.~Baldan, and O.~Beijbom, ``nuscenes: A multimodal dataset for
  autonomous driving,'' in \emph{CVPR}, 2020, pp. 11\,621--11\,631.

\bibitem{mao2021one}
J.~Mao, M.~Niu, C.~Jiang, H.~Liang, J.~Chen, X.~Liang, Y.~Li, C.~Ye, W.~Zhang,
  Z.~Li \emph{et~al.}, ``One million scenes for autonomous driving: Once
  dataset,'' \emph{arXiv:2106.11037}, 2021.

\bibitem{geiger2012we}
A.~Geiger, P.~Lenz, and R.~Urtasun, ``Are we ready for autonomous driving? the
  kitti vision benchmark suite,'' in \emph{CVPR}.\hskip 1em plus 0.5em minus
  0.4em\relax IEEE, 2012, pp. 3354--3361.

\bibitem{kesten2019lyft}
R.~Kesten, M.~Usman, J.~Houston, T.~Pandya, K.~Nadhamuni, A.~Ferreira, M.~Yuan,
  B.~Low, A.~Jain, P.~Ondruska \emph{et~al.}, ``Lyft level 5 av dataset 2019,''
  \emph{https://level5. lyft. com/dataset}, vol.~1, p.~3, 2019.

\bibitem{behley2021towards}
J.~Behley, M.~Garbade, A.~Milioto, J.~Quenzel, S.~Behnke, J.~Gall, and
  C.~Stachniss, ``Towards 3d lidar-based semantic scene understanding of 3d
  point cloud sequences: The semantickitti dataset,'' \emph{IJRR}, vol.~40, no.
  8-9, pp. 959--967, 2021.

\bibitem{nekrasov2021mix3d}
A.~Nekrasov, J.~Schult, O.~Litany, B.~Leibe, and F.~Engelmann, ``Mix3d:
  Out-of-context data augmentation for 3d scenes,'' in \emph{3DV}.\hskip 1em
  plus 0.5em minus 0.4em\relax IEEE, 2021, pp. 116--125.

\bibitem{fan2017point}
H.~Fan, H.~Su, and L.~J. Guibas, ``A point set generation network for 3d object
  reconstruction from a single image,'' in \emph{CVPR}, 2017, pp. 605--613.

\bibitem{wang2021unsupervised}
H.~Wang, Q.~Liu, X.~Yue, J.~Lasenby, and M.~J. Kusner, ``Unsupervised point
  cloud pre-training via occlusion completion,'' in \emph{CVPR}, 2021, pp.
  9782--9792.

\bibitem{zhang2022masked}
Y.~Zhang, J.~Lin, C.~He, Y.~Chen, K.~Jia, and L.~Zhang, ``Masked surfel
  prediction for self-supervised point cloud learning,''
  \emph{arXiv:2207.03111}, 2022.

\bibitem{tenney2019bert}
I.~Tenney, D.~Das, and E.~Pavlick, ``Bert rediscovers the classical nlp
  pipeline,'' \emph{arXiv:1905.05950}, 2019.

\bibitem{yu2022point}
X.~Yu, L.~Tang, Y.~Rao, T.~Huang, J.~Zhou, and J.~Lu, ``Point-bert:
  Pre-training 3d point cloud transformers with masked point modeling,'' in
  \emph{CVPR}, 2022, pp. 19\,313--19\,322.

\bibitem{fu2022point}
K.~Fu, M.~Yuan, and M.~Wang, ``Point-mcbert: A multi-choice self-supervised
  framework for point cloud pre-training,'' \emph{arXiv:2207.13226}, 2022.

\bibitem{wang2022self}
D.~Wang and Z.-X. Yang, ``Self-supervised point cloud understanding via mask
  transformer and contrastive learning,'' \emph{IEEE RAL}, vol.~8, no.~1, pp.
  184--191, 2022.

\bibitem{jiang2022masked}
J.~Jiang, X.~Lu, L.~Zhao, R.~Dazeley, and M.~Wang, ``Masked autoencoders in 3d
  point cloud representation learning,'' \emph{arXiv:2207.01545}, 2022.

\bibitem{pang2022masked}
Y.~Pang, W.~Wang, F.~E. Tay, W.~Liu, Y.~Tian, and L.~Yuan, ``Masked
  autoencoders for point cloud self-supervised learning,'' in
  \emph{ECCV}.\hskip 1em plus 0.5em minus 0.4em\relax Springer, 2022, pp.
  604--621.

\bibitem{zhang2022point}
R.~Zhang, Z.~Guo, P.~Gao, R.~Fang, B.~Zhao, D.~Wang, Y.~Qiao, and H.~Li,
  ``Point-m2ae: multi-scale masked autoencoders for hierarchical point cloud
  pre-training,'' \emph{arXiv:2205.14401}, 2022.

\bibitem{xu2022cp}
M.~Xu, Z.~Zhou, H.~Xu, Y.~Wang, and Y.~Qiao, ``Cp-net: Contour-perturbed
  reconstruction network for self-supervised point cloud learning,''
  \emph{arXiv:2201.08215}, 2022.

\bibitem{chen2021shape}
Y.~Chen, J.~Liu, B.~Ni, H.~Wang, J.~Yang, N.~Liu, T.~Li, and Q.~Tian, ``Shape
  self-correction for unsupervised point cloud understanding,'' in \emph{ICCV},
  2021, pp. 8382--8391.

\bibitem{zhang2022point1}
Y.~Zhang, J.~Lin, R.~Li, K.~Jia, and L.~Zhang, ``Point-dae: Denoising
  autoencoders for self-supervised point cloud learning,''
  \emph{arXiv:2211.06841}, 2022.

\bibitem{garg2022serp}
S.~Garg and M.~Chaudhary, ``Serp: Self-supervised representation learning using
  perturbed point clouds,'' \emph{arXiv preprint arXiv:2209.06067}, 2022.

\bibitem{qi2017pointnet}
C.~R. Qi, H.~Su, K.~Mo, and L.~J. Guibas, ``Pointnet: Deep learning on point
  sets for 3d classification and segmentation,'' in \emph{CVPR}, 2017, pp.
  652--660.

\bibitem{sauder2019self}
J.~Sauder and B.~Sievers, ``Self-supervised deep learning on point clouds by
  reconstructing space,'' \emph{NeurIPS}, vol.~32, 2019.

\bibitem{achituve2021self}
I.~Achituve, H.~Maron, and G.~Chechik, ``Self-supervised learning for domain
  adaptation on point clouds,'' in \emph{WACV}, 2021, pp. 123--133.

\bibitem{alwala2022pre}
K.~V. Alwala, A.~Gupta, and S.~Tulsiani, ``Pre-train, self-train, distill: A
  simple recipe for supersizing 3d reconstruction,'' in \emph{CVPR}, 2022, pp.
  3773--3782.

\bibitem{eckart2021self}
B.~Eckart, W.~Yuan, C.~Liu, and J.~Kautz, ``Self-supervised learning on 3d
  point clouds by learning discrete generative models,'' in \emph{CVPR}, 2021,
  pp. 8248--8257.

\bibitem{zhang2022upsampling}
C.~Zhang, J.~Shi, X.~Deng, and Z.~Wu, ``Upsampling autoencoder for
  self-supervised point cloud learning,'' \emph{arXiv:2203.10768}, 2022.

\bibitem{yan2022implicit}
S.~Yan, Z.~Yang, H.~Li, L.~Guan, H.~Kang, G.~Hua, and Q.~Huang, ``Implicit
  autoencoder for point cloud self-supervised representation learning,''
  \emph{arXiv:2201.00785}, 2022.

\bibitem{oord2018representation}
A.~v.~d. Oord, Y.~Li, and O.~Vinyals, ``Representation learning with
  contrastive predictive coding,'' \emph{arXiv:1807.03748}, 2018.

\bibitem{qi2021self}
Z.~Qi, S.~Wang, C.~Su, L.~Su, Q.~Huang, and Q.~Tian, ``Self-regulated learning
  for egocentric video activity anticipation,'' \emph{IEEE TPAMI}, 2021.

\bibitem{he2020momentum}
K.~He, H.~Fan, Y.~Wu, S.~Xie, and R.~Girshick, ``Momentum contrast for
  unsupervised visual representation learning,'' in \emph{CVPR}, 2020, pp.
  9729--9738.

\bibitem{park2020contrastive}
T.~Park, A.~A. Efros, R.~Zhang, and J.-Y. Zhu, ``Contrastive learning for
  unpaired image-to-image translation,'' in \emph{ECCV}.\hskip 1em plus 0.5em
  minus 0.4em\relax Springer, 2020, pp. 319--345.

\bibitem{han2021dual}
J.~Han, M.~Shoeiby, L.~Petersson, and M.~A. Armin, ``Dual contrastive learning
  for unsupervised image-to-image translation,'' in \emph{CVPR}, 2021, pp.
  746--755.

\bibitem{kang2020contragan}
M.~Kang and J.~Park, ``Contragan: Contrastive learning for conditional image
  generation,'' \emph{NeurIPS}, vol.~33, pp. 21\,357--21\,369, 2020.

\bibitem{chaitanya2020contrastive}
K.~Chaitanya, E.~Erdil, N.~Karani, and E.~Konukoglu, ``Contrastive learning of
  global and local features for medical image segmentation with limited
  annotations,'' \emph{NeurIPS}, vol.~33, pp. 12\,546--12\,558, 2020.

\bibitem{chen2020simple}
T.~Chen, S.~Kornblith, M.~Norouzi, and G.~Hinton, ``A simple framework for
  contrastive learning of visual representations,'' in \emph{ICML}.\hskip 1em
  plus 0.5em minus 0.4em\relax PMLR, 2020, pp. 1597--1607.

\bibitem{xie2020pointcontrast}
S.~Xie, J.~Gu, D.~Guo, C.~R. Qi, L.~Guibas, and O.~Litany, ``Pointcontrast:
  Unsupervised pre-training for 3d point cloud understanding,'' in
  \emph{ECCV}.\hskip 1em plus 0.5em minus 0.4em\relax Springer, 2020, pp.
  574--591.

\bibitem{zhang2021self}
Z.~Zhang, R.~Girdhar, A.~Joulin, and I.~Misra, ``Self-supervised pretraining of
  3d features on any point-cloud,'' in \emph{ICCV}, 2021, pp. 10\,252--10\,263.

\bibitem{choy2019fully}
C.~Choy, J.~Park, and V.~Koltun, ``Fully convolutional geometric features,'' in
  \emph{ICCV}, 2019, pp. 8958--8966.

\bibitem{hou2021exploring}
J.~Hou, B.~Graham, M.~Nie{\ss}ner, and S.~Xie, ``Exploring data-efficient 3d
  scene understanding with contrastive scene contexts,'' in \emph{CVPR}, 2021,
  pp. 15\,587--15\,597.

\bibitem{li2022closer}
L.~Li and M.~Heizmann, ``A closer look at invariances in self-supervised
  pre-training for 3d vision,'' in \emph{ECCV}.\hskip 1em plus 0.5em minus
  0.4em\relax Springer, 2022, pp. 656--673.

\bibitem{chen2021unsupervised}
H.~Chen, S.~Luo, X.~Gao, and W.~Hu, ``Unsupervised learning of geometric
  sampling invariant representations for 3d point clouds,'' in \emph{ICCV},
  2021, pp. 893--903.

\bibitem{vaswani2017attention}
A.~Vaswani, N.~Shazeer, N.~Parmar, J.~Uszkoreit, L.~Jones, A.~N. Gomez,
  {\L}.~Kaiser, and I.~Polosukhin, ``Attention is all you need,''
  \emph{NeurIPS}, vol.~30, 2017.

\bibitem{fu2022pos}
K.~Fu, P.~Gao, S.~Liu, R.~Zhang, Y.~Qiao, and M.~Wang, ``Pos-bert: Point cloud
  one-stage bert pre-training,'' \emph{arXiv:2204.00989}, 2022.

\bibitem{fu2022distillation}
K.~Fu, P.~Gao, R.~Zhang, H.~Li, Y.~Qiao, and M.~Wang, ``Distillation with
  contrast is all you need for self-supervised point cloud representation
  learning,'' \emph{arXiv:2202.04241}, 2022.

\bibitem{chen2021exploring}
X.~Chen and K.~He, ``Exploring simple siamese representation learning,'' in
  \emph{CVPR}, 2021, pp. 15\,750--15\,758.

\bibitem{mei2022unsupervised}
G.~Mei, X.~Huang, J.~Liu, J.~Zhang, and Q.~Wu, ``Unsupervised point cloud
  pre-training via contrasting and clustering,'' in \emph{ICIP}.\hskip 1em plus
  0.5em minus 0.4em\relax IEEE, 2022, pp. 66--70.

\bibitem{chen20224dcontrast}
Y.~Chen, M.~Nie{\ss}ner, and A.~Dai, ``4dcontrast: Contrastive learning with
  dynamic correspondences for 3d scene understanding,'' in \emph{ECCV}.\hskip
  1em plus 0.5em minus 0.4em\relax Springer, 2022, pp. 543--560.

\bibitem{yang2021unsupervised}
C.-K. Yang, Y.-Y. Chuang, and Y.-Y. Lin, ``Unsupervised point cloud object
  co-segmentation by co-contrastive learning and mutual attention sampling,''
  in \emph{ICCV}, 2021, pp. 7335--7344.

\bibitem{huang2021spatio}
S.~Huang, Y.~Xie, S.-C. Zhu, and Y.~Zhu, ``Spatio-temporal self-supervised
  representation learning for 3d point clouds,'' in \emph{ICCV}, 2021, pp.
  6535--6545.

\bibitem{zhai2019s4l}
X.~Zhai, A.~Oliver, A.~Kolesnikov, and L.~Beyer, ``S4l: Self-supervised
  semi-supervised learning,'' in \emph{ICCV}, 2019, pp. 1476--1485.

\bibitem{hendrycks2019using}
D.~Hendrycks, M.~Mazeika, S.~Kadavath, and D.~Song, ``Using self-supervised
  learning can improve model robustness and uncertainty,'' \emph{NeurIPS},
  vol.~32, 2019.

\bibitem{sun2021canonical}
W.~Sun, A.~Tagliasacchi, B.~Deng, S.~Sabour, S.~Yazdani, G.~E. Hinton, and
  K.~M. Yi, ``Canonical capsules: Self-supervised capsules in canonical pose,''
  \emph{Advances in Neural Information Processing Systems}, vol.~34, pp.
  24\,993--25\,005, 2021.

\bibitem{poursaeed2020self}
O.~Poursaeed, T.~Jiang, H.~Qiao, N.~Xu, and V.~G. Kim, ``Self-supervised
  learning of point clouds via orientation estimation,'' in \emph{3DV}.\hskip
  1em plus 0.5em minus 0.4em\relax IEEE, 2020, pp. 1018--1028.

\bibitem{sun2021adversarially}
J.~Sun, Y.~Cao, C.~B. Choy, Z.~Yu, A.~Anandkumar, Z.~M. Mao, and C.~Xiao,
  ``Adversarially robust 3d point cloud recognition using self-supervisions,''
  \emph{NeurIPS}, vol.~34, pp. 15\,498--15\,512, 2021.

\bibitem{cendra2022sl3d}
F.~J. Cendra, L.~Ma, J.~Shen, and X.~Qi, ``Sl3d: Self-supervised-self-labeled
  3d recognition,'' \emph{arXiv:2210.16810}, 2022.

\bibitem{hassani2019unsupervised}
K.~Hassani and M.~Haley, ``Unsupervised multi-task feature learning on point
  clouds,'' in \emph{ICCV}, 2019, pp. 8160--8171.

\bibitem{sharma2020self}
C.~Sharma and M.~Kaul, ``Self-supervised few-shot learning on point clouds,''
  \emph{NeurIPS}, vol.~33, pp. 7212--7221, 2020.

\bibitem{rao2022pointglr}
Y.~Rao, J.~Lu, and J.~Zhou, ``Pointglr: Unsupervised structural representation
  learning of 3d point clouds,'' \emph{IEEE TPAMI}, 2022.

\bibitem{sun2022unsupervised}
M.~Sun, X.~Huang, Z.~Sun, Q.~Wang, and Y.~Yao, ``Unsupervised pre-training for
  3d object detection with transformer,'' in \emph{PRCV}.\hskip 1em plus 0.5em
  minus 0.4em\relax Springer, 2022, pp. 82--95.

\bibitem{yamada2022point}
R.~Yamada, H.~Kataoka, N.~Chiba, Y.~Domae, and T.~Ogata, ``Point cloud
  pre-training with natural 3d structures,'' in \emph{CVPR}, 2022, pp.
  21\,283--21\,293.

\bibitem{zhang2023flattening}
Q.~Zhang, J.~Hou, Y.~Qian, Y.~Zeng, J.~Zhang, and Y.~He, ``Flattening-net: Deep
  regular 2d representation for 3d point cloud analysis,'' \emph{IEEE TPAMI},
  2023.

\bibitem{zhang2022reggeonet}
Q.~Zhang, J.~Hou, Y.~Qian, A.~B. Chan, J.~Zhang, and Y.~He, ``Reggeonet:
  Learning regular representations for large-scale 3d point clouds,''
  \emph{IJCV}, vol. 130, no.~12, pp. 3100--3122, 2022.

\bibitem{xue2022ulip}
L.~Xue, M.~Gao, C.~Xing, R.~Mart{\'\i}n-Mart{\'\i}n, J.~Wu, C.~Xiong, R.~Xu,
  J.~C. Niebles, and S.~Savarese, ``Ulip: Learning unified representation of
  language, image and point cloud for 3d understanding,''
  \emph{arXiv:2212.05171}, 2022.

\bibitem{janda2022self}
A.~Janda, B.~Wagstaff, E.~G. Ng, and J.~Kelly, ``Self-supervised pre-training
  of 3d point cloud networks with image data,'' \emph{arXiv:2211.11801}, 2022.

\bibitem{sun2022self}
C.~Sun, Z.~Zheng, X.~Wang, M.~Xu, and Y.~Yang, ``Self-supervised point cloud
  representation learning via separating mixed shapes,'' \emph{IEEE TMM}, 2022.

\bibitem{zhou2022pointcmc}
H.~Zhou, X.~Peng, J.~Mao, Z.~Wu, and M.~Zeng, ``Pointcmc: Cross-modal
  multi-scale correspondences learning for point cloud understanding,''
  \emph{arXiv:2211.12032}, 2022.

\bibitem{jing2021self}
L.~Jing, L.~Zhang, and Y.~Tian, ``Self-supervised feature learning by
  cross-modality and cross-view correspondences,'' in \emph{CVPR}, 2021, pp.
  1581--1591.

\bibitem{zhang2022learning}
R.~Zhang, L.~Wang, Y.~Qiao, P.~Gao, and H.~Li, ``Learning 3d representations
  from 2d pre-trained models via image-to-point masked autoencoders,''
  \emph{arXiv:2212.06785}, 2022.

\bibitem{zhang2019aet}
L.~Zhang, G.-J. Qi, L.~Wang, and J.~Luo, ``Aet vs. aed: Unsupervised
  representation learning by auto-encoding transformations rather than data,''
  in \emph{CVPR}, 2019, pp. 2547--2555.

\bibitem{gao2021self}
X.~Gao, W.~Hu, and G.-J. Qi, ``Self-supervised multi-view learning via
  auto-encoding 3d transformations,'' \emph{arXiv:2103.00787}, 2021.

\bibitem{tran2022self}
B.~Tran, B.-S. Hua, A.~T. Tran, and M.~Hoai, ``Self-supervised learning with
  multi-view rendering for 3d point cloud analysis,'' in \emph{ACCV}, 2022, pp.
  3086--3103.

\bibitem{zhang2022self}
Q.~Zhang and J.~Hou, ``Self-supervised pre-training for 3d point clouds via
  view-specific point-to-image translation,'' \emph{arXiv:2212.14197}, 2022.

\bibitem{wang2022p2p}
Z.~Wang, X.~Yu, Y.~Rao, J.~Zhou, and J.~Lu, ``P2p: Tuning pre-trained image
  models for point cloud analysis with point-to-pixel prompting,''
  \emph{arXiv:2208.02812}, 2022.

\bibitem{qian2022pix4point}
G.~Qian, X.~Zhang, A.~Hamdi, and B.~Ghanem, ``Pix4point: Image pretrained
  transformers for 3d point cloud understanding,'' \emph{arXiv:2208.12259},
  2022.

\bibitem{radford2021learning}
A.~Radford, J.~W. Kim, C.~Hallacy, A.~Ramesh, G.~Goh, S.~Agarwal, G.~Sastry,
  A.~Askell, P.~Mishkin, J.~Clark \emph{et~al.}, ``Learning transferable visual
  models from natural language supervision,'' in \emph{ICML}.\hskip 1em plus
  0.5em minus 0.4em\relax PMLR, 2021, pp. 8748--8763.

\bibitem{rombach2022high}
R.~Rombach, A.~Blattmann, D.~Lorenz, P.~Esser, and B.~Ommer, ``High-resolution
  image synthesis with latent diffusion models,'' in \emph{CVPR}, 2022, pp.
  10\,684--10\,695.

\bibitem{zhang2022pointclip}
R.~Zhang, Z.~Guo, W.~Zhang, K.~Li, X.~Miao, B.~Cui, Y.~Qiao, P.~Gao, and H.~Li,
  ``Pointclip: Point cloud understanding by clip,'' in \emph{CVPR}, 2022, pp.
  8552--8562.

\bibitem{zhu2022pointclip}
X.~Zhu, R.~Zhang, B.~He, Z.~Zeng, S.~Zhang, and P.~Gao, ``Pointclip v2:
  Adapting clip for powerful 3d open-world learning,'' \emph{arXiv:2211.11682},
  2022.

\bibitem{brown2020language}
T.~Brown, B.~Mann, N.~Ryder, M.~Subbiah, J.~D. Kaplan, P.~Dhariwal,
  A.~Neelakantan, P.~Shyam, G.~Sastry, A.~Askell \emph{et~al.}, ``Language
  models are few-shot learners,'' \emph{NeurIPS}, vol.~33, pp. 1877--1901,
  2020.

\bibitem{huang2022frozen}
X.~Huang, S.~Li, W.~Qu, T.~He, Y.~Zuo, and W.~Ouyang, ``Frozen clip model is
  efficient point cloud backbone,'' \emph{arXiv:2212.04098}, 2022.

\bibitem{nunes2022segcontrast}
L.~Nunes, R.~Marcuzzi, X.~Chen, J.~Behley, and C.~Stachniss, ``Segcontrast: 3d
  point cloud feature representation learning through self-supervised segment
  discrimination,'' \emph{IEEE RAL}, vol.~7, no.~2, pp. 2116--2123, 2022.

\bibitem{yin2022proposalcontrast}
J.~Yin, D.~Zhou, L.~Zhang, J.~Fang, C.-Z. Xu, J.~Shen, and W.~Wang,
  ``Proposalcontrast: Unsupervised pre-training for lidar-based 3d object
  detection,'' in \emph{ECCV}.\hskip 1em plus 0.5em minus 0.4em\relax Springer,
  2022, pp. 17--33.

\bibitem{min2022voxel}
C.~Min, D.~Zhao, L.~Xiao, Y.~Nie, and B.~Dai, ``Voxel-mae: Masked autoencoders
  for pre-training large-scale point clouds,'' \emph{arXiv:2206.09900}, 2022.

\bibitem{hess2022masked}
G.~Hess, J.~Jaxing, E.~Svensson, D.~Hagerman, C.~Petersson, and L.~Svensson,
  ``Masked autoencoders for self-supervised learning on automotive point
  clouds,'' \emph{arXiv:2207.00531}, 2022.

\bibitem{krispel2022maeli}
G.~Krispel, D.~Schinagl, C.~Fruhwirth-Reisinger, H.~Possegger, and H.~Bischof,
  ``Maeli--masked autoencoder for large-scale lidar point clouds,''
  \emph{arXiv:2212.07207}, 2022.

\bibitem{yang2022gd}
H.~Yang, T.~He, J.~Liu, H.~Chen, B.~Wu, B.~Lin, X.~He, and W.~Ouyang, ``Gd-mae:
  Generative decoder for mae pre-training on lidar point clouds,''
  \emph{arXiv:2212.03010}, 2022.

\bibitem{ren2022self}
Y.~Ren, P.~Cong, X.~Zhu, and Y.~Ma, ``Self-supervised point cloud completion on
  real traffic scenes via scene-concerned bottom-up mechanism,''
  \emph{arXiv:2203.10569}, 2022.

\bibitem{xie2022masked}
G.~Xie, Y.~Li, H.~Qu, and Z.~Sun, ``Masked autoencoder for pre-training on 3d
  point cloud object detection,'' \emph{Mathematics}, vol.~10, no.~19, p. 3549,
  2022.

\bibitem{boulch2022also}
A.~Boulch, C.~Sautier, B.~Michele, G.~Puy, and R.~Marlet, ``Also: Automotive
  lidar self-supervision by occupancy estimation,'' \emph{arXiv:2212.05867},
  2022.

\bibitem{mescheder2019occupancy}
L.~Mescheder, M.~Oechsle, M.~Niemeyer, S.~Nowozin, and A.~Geiger, ``Occupancy
  networks: Learning 3d reconstruction in function space,'' in \emph{CVPR},
  2019, pp. 4460--4470.

\bibitem{shi2023grid}
Y.~Shi, K.~Jiang, J.~Li, J.~Wen, Z.~Qian, M.~Yang, K.~Wang, and D.~Yang,
  ``Grid-centric traffic scenario perception for autonomous driving: A
  comprehensive review,'' \emph{arXiv:2303.01212}, 2023.

\bibitem{lin2022bev}
Z.~Lin and Y.~Wang, ``Bev-mae: Bird's eye view masked autoencoders for outdoor
  point cloud pre-training,'' \emph{arXiv:2212.05758}, 2022.

\bibitem{huang2023tri}
Y.~Huang, W.~Zheng, Y.~Zhang, J.~Zhou, and J.~Lu, ``Tri-perspective view for
  vision-based 3d semantic occupancy prediction,'' \emph{arXiv:2302.07817},
  2023.

\bibitem{shi2022self}
W.~Shi and R.~R. Rajkumar, ``Self-supervised pretraining for point cloud object
  detection in autonomous driving,'' in \emph{ITSC}.\hskip 1em plus 0.5em minus
  0.4em\relax IEEE, 2022, pp. 4341--4348.

\bibitem{liang2021exploring}
H.~Liang, C.~Jiang, D.~Feng, X.~Chen, H.~Xu, X.~Liang, W.~Zhang, Z.~Li, and
  L.~Van~Gool, ``Exploring geometry-aware contrast and clustering harmonization
  for self-supervised 3d object detection,'' in \emph{ICCV}, 2021, pp.
  3293--3302.

\bibitem{li2022simipu}
Z.~Li, Z.~Chen, A.~Li, L.~Fang, Q.~Jiang, X.~Liu, J.~Jiang, B.~Zhou, and
  H.~Zhao, ``Simipu: Simple 2d image and 3d point cloud unsupervised
  pre-training for spatial-aware visual representations,'' in \emph{AAAI},
  vol.~36, no.~2, 2022, pp. 1500--1508.

\bibitem{chen2022co}
R.~Chen, Y.~Mu, R.~Xu, W.~Shao, C.~Jiang, H.~Xu, Z.~Li, and P.~Luo, ``Co\^{} 3:
  Cooperative unsupervised 3d representation learning for autonomous driving,''
  \emph{arXiv:2206.04028}, 2022.

\bibitem{wang2022rethinking}
H.~Wang, X.~Guo, Z.-H. Deng, and Y.~Lu, ``Rethinking minimal sufficient
  representation in contrastive learning,'' in \emph{CVPR}, 2022, pp.
  16\,041--16\,050.

\bibitem{zheng2022boosting}
W.~Zheng, M.~Hong, L.~Jiang, and C.-W. Fu, ``Boosting 3d object detection by
  simulating multimodality on point clouds,'' in \emph{CVPR}, 2022, pp.
  13\,638--13\,647.

\bibitem{mersch2022self}
B.~Mersch, X.~Chen, J.~Behley, and C.~Stachniss, ``Self-supervised point cloud
  prediction using 3d spatio-temporal convolutional networks,'' in
  \emph{CoRL}.\hskip 1em plus 0.5em minus 0.4em\relax PMLR, 2022, pp.
  1444--1454.

\bibitem{weng2021inverting}
X.~Weng, J.~Wang, S.~Levine, K.~Kitani, and N.~Rhinehart, ``Inverting the pose
  forecasting pipeline with spf2: Sequential pointcloud forecasting for
  sequential pose forecasting,'' in \emph{CoRL}.\hskip 1em plus 0.5em minus
  0.4em\relax PMLR, 2021, pp. 11--20.

\bibitem{lu2021monet}
F.~Lu, G.~Chen, Z.~Li, L.~Zhang, Y.~Liu, S.~Qu, and A.~Knoll, ``Monet:
  Motion-based point cloud prediction network,'' \emph{IEEE TITS}, vol.~23,
  no.~8, pp. 13\,794--13\,804, 2021.

\bibitem{hoermann2018dynamic}
S.~Hoermann, M.~Bach, and K.~Dietmayer, ``Dynamic occupancy grid prediction for
  urban autonomous driving: A deep learning approach with fully automatic
  labeling,'' in \emph{ICRA}.\hskip 1em plus 0.5em minus 0.4em\relax IEEE,
  2018, pp. 2056--2063.

\bibitem{song20192d}
Y.~Song, Y.~Tian, G.~Wang, and M.~Li, ``2d lidar map prediction via estimating
  motion flow with gru,'' in \emph{ICRA}.\hskip 1em plus 0.5em minus
  0.4em\relax IEEE, 2019, pp. 6617--6623.

\bibitem{wu2020motionnet}
P.~Wu, S.~Chen, and D.~N. Metaxas, ``Motionnet: Joint perception and motion
  prediction for autonomous driving based on bird's eye view maps,'' in
  \emph{CVPR}, 2020, pp. 11\,385--11\,395.

\bibitem{toyungyernsub2021double}
M.~Toyungyernsub, M.~Itkina, R.~Senanayake, and M.~J. Kochenderfer,
  ``Double-prong convlstm for spatiotemporal occupancy prediction in dynamic
  environments,'' in \emph{ICRA}.\hskip 1em plus 0.5em minus 0.4em\relax IEEE,
  2021, pp. 13\,931--13\,937.

\bibitem{thabet2020self}
A.~Thabet, H.~Alwassel, and B.~Ghanem, ``Self-supervised learning of local
  features in 3d point clouds,'' in \emph{CVPR Workshops}, 2020, pp. 938--939.

\bibitem{erccelik20223d}
E.~Er{\c{c}}elik, E.~Yurtsever, M.~Liu, Z.~Yang, H.~Zhang, P.~Top{\c{c}}am,
  M.~Listl, Y.~K. {\c{C}}ayl{\i}, and A.~Knoll, ``3d object detection with a
  self-supervised lidar scene flow backbone,'' \emph{arXiv:2205.00705}, 2022.

\bibitem{liu2019flownet3d}
X.~Liu, C.~R. Qi, and L.~J. Guibas, ``Flownet3d: Learning scene flow in 3d
  point clouds,'' in \emph{CVPR}, 2019, pp. 529--537.

\bibitem{mittal2020just}
H.~Mittal, B.~Okorn, and D.~Held, ``Just go with the flow: Self-supervised
  scene flow estimation,'' in \emph{CVPR}, 2020, pp. 11\,177--11\,185.

\bibitem{li2021self}
R.~Li, G.~Lin, and L.~Xie, ``Self-point-flow: Self-supervised scene flow
  estimation from point clouds with optimal transport and random walk,'' in
  \emph{CVPR}, 2021, pp. 15\,577--15\,586.

\bibitem{altschuler2017near}
J.~Altschuler, J.~Niles-Weed, and P.~Rigollet, ``Near-linear time approximation
  algorithms for optimal transport via sinkhorn iteration,'' \emph{NeurIPS},
  vol.~30, 2017.

\bibitem{baur2021slim}
S.~A. Baur, D.~J. Emmerichs, F.~Moosmann, P.~Pinggera, B.~Ommer, and A.~Geiger,
  ``Slim: Self-supervised lidar scene flow and motion segmentation,'' in
  \emph{ICCV}, 2021, pp. 13\,126--13\,136.

\bibitem{kittenplon2021flowstep3d}
Y.~Kittenplon, Y.~C. Eldar, and D.~Raviv, ``Flowstep3d: Model unrolling for
  self-supervised scene flow estimation,'' in \emph{CVPR}, 2021, pp.
  4114--4123.

\bibitem{pontes2020scene}
J.~K. Pontes, J.~Hays, and S.~Lucey, ``Scene flow from point clouds with or
  without learning,'' in \emph{3DV}.\hskip 1em plus 0.5em minus 0.4em\relax
  IEEE, 2020, pp. 261--270.

\bibitem{tishchenko2020self}
I.~Tishchenko, S.~Lombardi, M.~R. Oswald, and M.~Pollefeys, ``Self-supervised
  learning of non-rigid residual flow and ego-motion,'' in \emph{3DV}.\hskip
  1em plus 0.5em minus 0.4em\relax IEEE, 2020, pp. 150--159.

\bibitem{wu2020pointpwc}
W.~Wu, Z.~Y. Wang, Z.~Li, W.~Liu, and L.~Fuxin, ``Pointpwc-net: Cost volume on
  point clouds for (self-) supervised scene flow estimation,'' in
  \emph{ECCV}.\hskip 1em plus 0.5em minus 0.4em\relax Springer, 2020, pp.
  88--107.

\bibitem{li2022rigidflow}
R.~Li, C.~Zhang, G.~Lin, Z.~Wang, and C.~Shen, ``Rigidflow: Self-supervised
  scene flow learning on point clouds by local rigidity prior,'' in
  \emph{CVPR}, 2022, pp. 16\,959--16\,968.

\bibitem{zhao2020maskflownet}
S.~Zhao, Y.~Sheng, Y.~Dong, E.~I. Chang, Y.~Xu \emph{et~al.}, ``Maskflownet:
  Asymmetric feature matching with learnable occlusion mask,'' in \emph{CVPR},
  2020, pp. 6278--6287.

\bibitem{saxena2019pwoc}
R.~Saxena, R.~Schuster, O.~Wasenmuller, and D.~Stricker, ``Pwoc-3d: Deep
  occlusion-aware end-to-end scene flow estimation,'' in \emph{IEEE IV}.\hskip
  1em plus 0.5em minus 0.4em\relax IEEE, 2019, pp. 324--331.

\bibitem{ouyang2021occlusion}
B.~Ouyang and D.~Raviv, ``Occlusion guided self-supervised scene flow
  estimation on 3d point clouds,'' in \emph{3DV}.\hskip 1em plus 0.5em minus
  0.4em\relax IEEE, 2021, pp. 782--791.

\bibitem{song2022ogc}
Z.~Song and B.~Yang, ``Ogc: Unsupervised 3d object segmentation from rigid
  dynamics of point clouds,'' \emph{arXiv:2210.04458}, 2022.

\bibitem{nichol2022point}
A.~Nichol, H.~Jun, P.~Dhariwal, P.~Mishkin, and M.~Chen, ``Point-e: A system
  for generating 3d point clouds from complex prompts,''
  \emph{arXiv:2212.08751}, 2022.

\bibitem{zhao2021relationship}
X.~Zhao, B.~Zhang, J.~Wu, R.~Hu, and T.~Komura, ``Relationship-based point
  cloud completion,'' \emph{IEEE TVCG}, vol.~28, no.~12, pp. 4940--4950, 2021.

\bibitem{kirillov2023segment}
A.~Kirillov, E.~Mintun, N.~Ravi, H.~Mao, C.~Rolland, L.~Gustafson, T.~Xiao,
  S.~Whitehead, A.~C. Berg, W.-Y. Lo \emph{et~al.}, ``Segment anything,''
  \emph{arXiv:2304.02643}, 2023.

\bibitem{wang2023seggpt}
X.~Wang, X.~Zhang, Y.~Cao, W.~Wang, C.~Shen, and T.~Huang, ``Seggpt: Segmenting
  everything in context,'' \emph{arXiv:2304.03284}, 2023.

\bibitem{lahoud2020rgb}
J.~Lahoud and B.~Ghanem, ``Rgb-based semantic segmentation using
  self-supervised depth pre-training,'' \emph{arXiv:2002.02200}, 2020.

\bibitem{wang2020train}
Y.~Wang, X.~Chen, Y.~You, L.~E. Li, B.~Hariharan, M.~Campbell, K.~Q.
  Weinberger, and W.-L. Chao, ``Train in germany, test in the usa: Making 3d
  object detectors generalize,'' in \emph{CVPR}, 2020, pp. 11\,713--11\,723.

\bibitem{yang2021st3d}
J.~Yang, S.~Shi, Z.~Wang, H.~Li, and X.~Qi, ``St3d: Self-training for
  unsupervised domain adaptation on 3d object detection,'' in \emph{CVPR},
  2021, pp. 10\,368--10\,378.

\bibitem{yang2022st3d++}
{Yang, Jihan and Shi, Shaoshuai and Wang, Zhe and Li, Hongsheng and Qi,
  Xiaojuan}, ``St3d++: denoised self-training for unsupervised domain
  adaptation on 3d object detection,'' \emph{IEEE TPAMI}, 2022.

\bibitem{niederlohner2022self}
D.~Niederl{\"o}hner, M.~Ulrich, S.~Braun, D.~K{\"o}hler, F.~Faion,
  C.~Gl{\"a}ser, A.~Treptow, and H.~Blume, ``Self-supervised velocity
  estimation for automotive radar object detection networks,'' in \emph{IEEE
  IV}.\hskip 1em plus 0.5em minus 0.4em\relax IEEE, 2022, pp. 352--359.

\bibitem{liu2022masked}
H.~Liu, M.~Cai, and Y.~J. Lee, ``Masked discrimination for self-supervised
  learning on point clouds,'' in \emph{ECCV}.\hskip 1em plus 0.5em minus
  0.4em\relax Springer, 2022, pp. 657--675.

\bibitem{huang2022ponder}
D.~Huang, S.~Peng, T.~He, X.~Zhou, and W.~Ouyang, ``Ponder: Point cloud
  pre-training via neural rendering,'' \emph{arXiv:2301.00157}, 2022.

\bibitem{dong2022autoencoders}
R.~Dong, Z.~Qi, L.~Zhang, J.~Zhang, J.~Sun, Z.~Ge, L.~Yi, and K.~Ma,
  ``Autoencoders as cross-modal teachers: Can pretrained 2d image transformers
  help 3d representation learning?'' \emph{arXiv:2212.08320}, 2022.

\bibitem{hamdi2021voint}
A.~Hamdi, S.~Giancola, and B.~Ghanem, ``Voint cloud: Multi-view point cloud
  representation for 3d understanding,'' \emph{arXiv:2111.15363}, 2021.

\bibitem{wang2022estimation}
K.~Wang and S.~Shen, ``Estimation and propagation: Scene flow prediction on
  occluded point clouds,'' \emph{IEEE RAL}, vol.~7, no.~4, pp.
  12\,201--12\,208, 2022.

\bibitem{ma2022seqot}
J.~Ma, X.~Chen, J.~Xu, and G.~Xiong, ``Seqot: A spatial-temporal transformer
  network for place recognition using sequential lidar data,'' \emph{IEEE TIE},
  2022.

\bibitem{zhang2020paranet}
Q.~Zhang, J.~Hou, Y.~Qian, J.~Zhang, and Y.~He, ``Paranet: Deep regular
  representation for 3d point clouds,'' \emph{arXiv:2012.03028}, 2020.

\end{thebibliography}

\appendices
The Supplementary Material includes other methods of object- and indoor scene-level SSL, other methods of outdoor scene-level SSL, a comparison of the advantages and limitations of these methods, and details about the performance comparisons on various downstream datasets in Table I-XI.

\section{Other Methods of Object- and Indoor Scene-level SSL}

In addition to the previously mentioned methods, another approach by Lahoud and Ghanem \cite{lahoud2020rgb} focus on RGB segmentation. This method capitalizes on the availability of depth sensors to generate automatically labeled data, which can be used for pre-training any semantic RGB segmentation method.
The pre-training process in this method involves leveraging height-normal (HN) labels, which are generated from depth sensors and represent different heights and normal patches in the data. The HN labels are particularly useful for extracting local semantic information. By using this automatically generated data, the pre-training approach can overcome the challenge of obtaining large-scale annotated data for semantic RGB segmentation tasks.

\section{Other Methods of Outdoor Scene-level SSL}
% The unordered and unstructured nature of point clouds makes it challenging to apply convolutional methods and makes point-wise approaches more desirable. The point-wise methods mainly consist of two steps, i.e., point-wise computations and aggregation. \cite{thabet2019mortonnet}, one of the first to adopt self-supervision on 3D point clouds, formulates the point-wise feature encoding as a point prediction task, using a multi-layer RNN to predict the next point in a point sequence and learning local features using Z-order Space Filling Curve. The method is first trained on S3DIS and then transferred to vKITTI dataset with an improvement over SOTA of 3.8$\%$, which indicates its merit to boost the performance of downstream tasks like 3D segmentation. 

3D detectors that are well-trained on a specific domain may experience a significant performance drop when transferred to a new domain due to changes like sensor types, geographical locations, object sizes, and even weather conditions. Thus, the unsupervised domain adaptation task aims to generalize models trained on labeled source domains to unlabeled target domains. Wang et al.~\cite{wang2020train} proposed SN that narrows the size-level domain gap by normalizing object sizes. However, this method requires statistical information and data distribution. In contrast, Yang et al.~\cite{yang2021st3d} developed a self-training pipeline called ST3D, which does not rely on target object statistics and incorporates three major adjustments: (1) A random object scaling (ROS) augmentation technique is used during labeled pre-training to reduce the negative impact of object size bias in the source domain. 
(2) To iteratively generate finer pseudo labels for the target domain, a quality-aware triplet memory bank (QTMB) mechanism is introduced, consisting of an IoU-based scoring criterion to assess localization quality, a triplet box partition scheme that reduces noisy pseudo labels from ambiguous boxes, and a memory bank that updates historical pseudo labels through memory ensemble and voting. 
(3) During the training phase, a curriculum data augmentation (CDA) strategy is adopted to generate diverse and challenging examples, making the model less prone to overfitting on easy positive examples with high confidence. After the labeled pre-train phase, ST3D iteratively switches between the pseudo-label generation phase and the training phase until convergence.

Building upon the work of ST3D, Yang et al.~\cite{yang2022st3d++} introduced the ST3D++ framework to further address the pseudo label noise, specifically localization noise and classification noise, in a  systematical manner. Two major redesigns were made in ST3D++: 
(1) The IoU-based scoring criterion in label generation was replaced with a hybrid quality-aware criterion that combines classification confidence and IoU scores in a weighted manner for improved assessment. 
(2) In the model training phase, the curriculum data augmentation strategy was complemented by source-assisted self-denoised training (SASD) to mitigate the negative impacts of noisy data on model optimization. SASD employs joint optimization on source and target domain data to reduce errors in gradient directions with clean and diverse source data. Additionally, to alleviate potential domain shifts caused by joint optimization, source and target data are separately normalized and transformed with shared scale and shift parameters.
Yang et al. validated ST3D++ across multiple categories in four adaptation settings.
% , SF-UD$A^{3D}$~\cite{saltori2020sf}, Dreaming~\cite{you2022exploiting} and MLC-Net~\cite{luo2021unsupervised} by a large margin.

Most of the above-mentioned works focus on LiDAR-based data and achieve promising results. However, LiDAR data lacks object velocity information, which is crucial for objection detection and tracking. In this context, automotive radar sensors serve as suitable alternatives due to their low cost, robustness, and ability to measure radial velocity. Niederlohner et al.~\cite{niederlohner2022self} introduced a two-step self-supervised method to learn object Cartesian velocities using only single-frame, oriented bounding boxes (OBB) labels. The OBB labels are only needed during the detection steps to pre-train the network, allowing it to generate OBB predictions on unlabeled data. In the velocity step, the distance between OBB predictions is used as a self-supervised velocity loss to guide the estimation of object velocity.

% \usepackage{graphicx}
\begin{table*}[htp]
\centering
\tabcolsep=0.05cm
\caption{Summary of self-supervised learning techniques for point cloud pre-training, focusing on object- and indoor scene-level point clouds.}
\resizebox{\textwidth}{!}{%
\begin{tabular}{>{\columncolor{C7!50}}l>{\columncolor{C6!50}}c>{\columncolor{C5!50}}c>{\columncolor{C4!50}}c>{\columncolor{C3!50}}c>{\columncolor{C2!50}}c>{\columncolor{C1!50}}c}
\toprule[1.5pt]
Methods   & Venue & Year & Category         & Pre-trained dataset & Contributions                                                                                                                         & Limitations                                                                       \\ \midrule[1pt]
RS~\cite{sauder2019self} & NeurIPS & 2019 & Reconstruction & ShapeNet55 & \begin{tabular}[c]{@{}c@{}}Eliminate computational cost and flawed \\ reconstruction losses or similarity metrics on point clouds. \end{tabular}& \begin{tabular}[c]{@{}c@{}}Many works that distort the 3D shape do not\\apply to RS due to simply rearrange shape parts.\end{tabular}\\\hline

OcCo~\cite{wang2021unsupervised} & ICCV & 2021 & Reconstruction & ModelNet40 & \begin{tabular}[c]{@{}c@{}} Reconstruct a point cloud whose\\ parts have
been randomly displaced. \end{tabular}& \begin{tabular}[c]{@{}c@{}} Designing the model that is aware of the occlusion procedure \\ may converge even quicker and require fewer parameters. \end{tabular} \\\hline

MaskSurf~\cite{zhang2022masked} & arXiv & 2022 & Reconstruction & ShapeNet55 & \begin{tabular}[c]{@{}c@{}} Firstly try to consider the local geometry information \\ explicitly into the masked auto-encoding. \end{tabular}& \begin{tabular}[c]{@{}c@{}} Can be extended to other surfel \\ orientations except for normals. \end{tabular} \\\hline

PointBERT~\cite{yu2022point} & CVPR      & 2022 & Reconstruction & ShapeNet55          & \begin{tabular}[c]{@{}c@{}} Firstly introduce BERT into point cloud \\ understanding by recovering masked object parts.\end{tabular} & \begin{tabular}[c]{@{}c@{}}Need dVAE to tokenize \\ the point cloud.\end{tabular}\\ \hline

McP-BERT~\cite{fu2022point} & arXiv & 2022 & Reconstruction & ShapeNet55 & \begin{tabular}[c]{@{}c@{}}Solve the problem of token-ambiguous \\and improper tokenizer for Point-BERT.\end{tabular} & \begin{tabular}[c]{@{}c@{}}The effectiveness of multi-choice tokens \\ should be further proved.\end{tabular}\\ \hline

POS-BERT~\cite{fu2022pos} & arXiv & 2022 & Reconstruction & ShapeNet55 & \begin{tabular}[c]{@{}c@{}}Use dynamically updated momentum encoder rather \\ than a frozen tokenizer for Point-BERT to exploit features.\end{tabular} & \begin{tabular}[c]{@{}c@{}}Rely on training a tokenizer for embedding \\ and recover the complete point cloud. \end{tabular}\\\hline

MAE3D~\cite{jiang2022masked} & arXiv & 2022 & Reconstruction & ShapeNet55 & \begin{tabular}[c]{@{}c@{}}MAE3D can be more lightweight for\\ pre-training on large-scale datasets.\end{tabular} & \begin{tabular}[c]{@{}c@{}}The FoldingNet-decoder \\ can be further enhanced.\end{tabular} \\\hline

Point-MAE~\cite{pang2022masked} & ECCV        & 2022 & Reconstruction & ShapeNet55          & \begin{tabular}[c]{@{}c@{}} Remove the dVAE and directly divide \\ the input 3D shape into point patches.\end{tabular} & \begin{tabular}[c]{@{}c@{}} Single level of reconstruction \\ make it underperform Point-M2AE. \end{tabular}\\ \hline

Point-M2AE~\cite{zhang2022point} & NeurIPS        & 2022 & Reconstruction & ShapeNet55  & \begin{tabular}[c]{@{}c@{}} Propose a multi-scale Mask-Autoencoder pre-training pipeline\\ for hierarchical SSL of 3D point clouds.\end{tabular} & \begin{tabular}[c]{@{}c@{}} Multi-level reconstruction results in\\ more computational cost. \end{tabular}\\ \hline

CP-Net~\cite{xu2022cp} & arXiv & 2022 & Reconstruction & \begin{tabular}[c]{@{}c@{}}ShapeNet55, ShapeNetPart \\ \& ModelNet40 \end{tabular} & \begin{tabular}[c]{@{}c@{}}Own the ability to distinguish the \\semantic parts of content components.\end{tabular} & \begin{tabular}[c]{@{}c@{}}The relationship between local structure \\and global shape is not considered.\end{tabular} \\\hline

Point-DAE~\cite{zhang2022point1} & arXiv & 2022 & Reconstruction & ShapeNet55 \&   SVD-Pose & \begin{tabular}[c]{@{}c@{}}Investigate more types of corruptions \\ beyond masking and estimate object poses automatically.\end{tabular}& \begin{tabular}[c]{@{}c@{}}The reconstruction results can be \\enhanced by density-aware loss.\end{tabular} \\\hline

SeRP~\cite{garg2022serp} & arXiv & 2022 & Reconstruction & ShapeNet55 & \begin{tabular}[c]{@{}c@{}}Address the limitations of Masked Auto-Encoders, which tend \\ to leak of location uneven density information.\end{tabular} & \begin{tabular}[c]{@{}c@{}}Encourage corruption at the edges than \\inside the volume for the future work.\end{tabular} \\\hline

UAE~\cite{zhang2022upsampling} & arXiv & 2022 & Reconstruction & ShapeNet55 & \begin{tabular}[c]{@{}c@{}}Use point cloud upsampling as a pre-training \\ task and designing a novel joint loss function.\end{tabular} & \begin{tabular}[c]{@{}c@{}}The performance can be  \\ further improved.\end{tabular} \\\hline

DefRec~\cite{achituve2021self} & WACV & 2021 & Reconstruction & PointDA-10 \& PointSegDA & \begin{tabular}[c]{@{}c@{}}Study the point cloud reconstruction pre-tasks \\ for domain adaptation on 3D point clouds.\end{tabular} & \begin{tabular}[c]{@{}c@{}}Variations in point clouds are not explicitly focused \\ on because of difference in configurations of LiDAR sensors.\end{tabular}\\\hline

ParAE~\cite{eckart2021self} & CVPR & 2021 & Reconstruction & ShapeNet55 &\begin{tabular}[c]{@{}c@{}} Remain agnostic to the underlying DNN architecture, and \\ leverage the geometric information of point clouds.\end{tabular} & \begin{tabular}[c]{@{}c@{}}ParAE is computationally intensive and relies \\excessively on reconstructing local details.\end{tabular}\\\hline

MaskPoint~\cite{liu2022masked} & ECCV        & 2022 & Reconstruction & ShapeNet55          & \begin{tabular}[c]{@{}c@{}} Introduce a Discriminator to \\ regard the point cloud as discrete occupancy values.\end{tabular} & \begin{tabular}[c]{@{}c@{}} The detection performance \\ can be promoted. \end{tabular}\\ \hline

Ponder~\cite{huang2022ponder} & arxiv        & 2022 & Reconstruction & ScanNet          & \begin{tabular}[c]{@{}c@{}}  Encodes rich geometry and appearance clues \\ by leveraging neural rendering.\end{tabular} & \begin{tabular}[c]{@{}c@{}} This protocol might be utilized \\ in outdoor scene understanding.  \end{tabular}\\ \hline

ACT~\cite{dong2022autoencoders} & arxiv        & 2022 & Reconstruction & ShapeNet55          & \begin{tabular}[c]{@{}c@{}}  Transformers pre-trained with 2D images or
texts \\ can benefit for 3D SSL.\end{tabular} & \begin{tabular}[c]{@{}c@{}}  The dVAE can be \\removed like Point-MAE. \end{tabular}\\

\hline
PointContrast~\cite{xie2020pointcontrast} & ECCV        & 2020 & Contrastive learning & ScanNet          & \begin{tabular}[c]{@{}c@{}} Firstly demonstrate the transferability \\of learned representation
in point clouds \\to high-level scene understanding \end{tabular} & \begin{tabular}[c]{@{}c@{}} Underutilization of spatial contextual information,  \\insufficient scalability and expensive constraints. \end{tabular}\\

\hline
CSC~\cite{hou2021exploring} & CVPR        & 2021 & Contrastive learning & ScanNet          & \begin{tabular}[c]{@{}c@{}} Integrate spatial contexts into the pre-training objective \\and firstly employ data-efficient learning\\
in a large-scale and real-world 3D scene understanding scenarios. \end{tabular} & \begin{tabular}[c]{@{}c@{}} Can be employed in \\outdoor scenes \end{tabular}\\

\hline
DepthContrast~\cite{zhang2021self} & ICCV        & 2021 & Contrastive learning & ScanNet-vid          & \begin{tabular}[c]{@{}c@{}} Single-view depth scans and joint training \\ of different input representations \\are powerful for learning features. \end{tabular} & \begin{tabular}[c]{@{}c@{}} Might encourage self-supervised \\ learning on range images \end{tabular}\\

\hline
\begin{tabular}[c]{@{}c@{}} 3D Contrastive \\ Learning~\cite{li2022closer} \end{tabular}
& ECCV        & 2022 & \begin{tabular}[c]{@{}c@{}} Contrastive learning\\ \& multi-modality\end{tabular} & ScanNet          & \begin{tabular}[c]{@{}c@{}} Firstly provide a unified framework \\for fair and systematic comparisons of \\various contrative-learning-based pre-training methods. \end{tabular} & \begin{tabular}[c]{@{}c@{}} Haven't fully investigated the optimization\\ and convergence of the joint pre-training. \end{tabular}\\

\hline
GSIR~\cite{chen2021unsupervised}    & ICCV        & 2021 & Contrastive learning & ShapeNetPart          & \begin{tabular}[c]{@{}c@{}} Explore the geometric sampling invariant representations \\that are
invariant under various sampling strategies and densties. \end{tabular} & \begin{tabular}[c]{@{}c@{}} Can be applied to existing \\ transformer-based methods. \end{tabular}\\

\hline
ContrastMPCT~\cite{wang2022self} & RA-L        & 2022 & Contrastive learning & ShapeNet          & \begin{tabular}[c]{@{}c@{}} Proposing a novel pre-training strategy via \\ contrastive learning and mask Transformer. \end{tabular} & \begin{tabular}[c]{@{}c@{}} The way of contrastive learning \\ can be further promoted\end{tabular}\\

\hline 
CoLMSA~\cite{yang2021unsupervised}     & ICCV        & 2021 & Contrastive learning & ScanObjectNN \& ModelNet40          & \begin{tabular}[c]{@{}c@{}} Firstly develop an end-to-end network \\for 3D object co-segmentation. \end{tabular} & \begin{tabular}[c]{@{}c@{}} Can be employed on \\ real-world segmentation. \end{tabular}\\

\hline
ConClu~\cite{mei2022unsupervised} & ICIP        & 2022 & Contrastive learning & ModelNet40          & \begin{tabular}[c]{@{}c@{}} Propose a SimSam-based point cloud SSL method \\to alleviate the reliance on the negative samples\\ of contrastive learning. \end{tabular} & \begin{tabular}[c]{@{}c@{}} Performance on ScanObjectNN \\ should be conducted. \end{tabular}\\

\hline
4DContrast~\cite{chen20224dcontrast} & ECCV        & 2022 & Contrastive learning & ModelNet40+ScanNet          & \begin{tabular}[c]{@{}c@{}} Firstly leverage 4D sequence information and constraints \\for 3D representation learning. \end{tabular} & \begin{tabular}[c]{@{}c@{}} Considerable memory requirement of 4D feature learning \\ with sparse convolutions, and more \\ complex dynamic objects need to be explored. \end{tabular}\\

% \hline
% DCGLR~\cite{fu2022distillation} & arXiv        & 2022 &Spatial     & ShapeNet          & \begin{tabular}[c]{@{}c@{}}  Knowledge refinement and comparative learning \\ are integrated to propose a ViT-based point cloud network.\end{tabular} & \begin{tabular}[c]{@{}c@{}}  - \end{tabular}\\

\hline
STRL~\cite{huang2021spatio} & ICCV        & 2021 &Spatial-temporal     & ShapeNet \& ScanNet          & \begin{tabular}[c]{@{}c@{}}  Learn from unlabeled 3D shapes \\ with the spatio-temporal contextual information. \end{tabular} & \begin{tabular}[c]{@{}c@{}} Can be extended to \\ more effective models \end{tabular}\\

% \hline
% HN-Cross-Stitch Network~\cite{lahoud2020rgb}  &CVPR      &2020   &Spatial      &-   & \begin{tabular}[c]{@{}c@{}}  The proposed method can be used to \\pre-train any semantic RGB segmentation.\end{tabular} & \begin{tabular}[c]{@{}c@{}}  - \end{tabular}\\

% \hline
% MD~\cite{sun2022self}   &IEEE TMM       &2022       &Spatial   &ModelNet40+S3DIS      &\begin{tabular}[c]{@{}c@{}} Design the SSL method of Mixing and Disentangling (MD)\\ for 3D point cloud representation learning, where the\\  reconstruction task is used as a pretext optimization objective.\end{tabular} & \begin{tabular}[c]{@{}c@{}}  - \end{tabular}\\

\hline
Cover Tree~\cite{sharma2020self}    & NeurIPS      &2020       &Spatial      &Modelnet40     &\begin{tabular}[c]{@{}c@{}} Using a coverage tree allows a subset of point clouds to be located \\ in balls of different radii at each layer of the coverage tree.\end{tabular} & \begin{tabular}[c]{@{}c@{}}  The ball cover can be replaced by \\ other accurate primitives. \end{tabular}\\

\hline
Canonical Capsules\cite{sun2021canonical}   &NeurIPS       &2021       &Spatial      &ShapeNet   &\begin{tabular}[c]{@{}c@{}} Learn capsule decomposition by \\ simple transform augmentation.\end{tabular} & \begin{tabular}[c]{@{}c@{}}  The effectiveness can be validated on more datasets. \end{tabular}\\

\hline
SL3D~\cite{cendra2022sl3d}  &arXiv       &2022        &Spatial  & ModelNet40   &\begin{tabular}[c]{@{}c@{}} Address two coupled objectives, namely\\ clustering and learning feature representation,to generate\\ pseudo-labeled data for unsupervised 3D recognition.\end{tabular} & \begin{tabular}[c]{@{}c@{}}  SL3D can be applied on outdoor-scene tasks. \end{tabular}\\ 

% \hline
% IAM~\cite{yan2022implicit}  &arXiv      &2022   &Spatial      &ModelNet40     &\begin{tabular}[c]{@{}c@{}} Implicit Autoencoder (IAE) with  \\continuous outputis introduced.\end{tabular} & \begin{tabular}[c]{@{}c@{}}  - \end{tabular}\\ 

\hline
Rotation prediction~\cite{poursaeed2020self}  &IEEE 3DV      &2020   &Spatial      &ShapeNet     &\begin{tabular}[c]{@{}c@{}} Supervised signals were created for the prerequisite\\ task of rotation prediction, and models were \\designed for learning agents and downstream tasks.\end{tabular} & \begin{tabular}[c]{@{}c@{}}   Focuses on point clouds and is not sufficient\\ for exploring other 3D representations \\to design effective agent tasks.\end{tabular}\\

% \hline
% PN and RSNet~\cite{thabet2020self}  &CVPR      &2020   &Spatial      &     &\begin{tabular}[c]{@{}c@{}} SSL tasks using multilayer RNNs and Morton's\\ order curve designs can learn meaningful point-level\\ features to encode local structure around points.\end{tabular} & \begin{tabular}[c]{@{}c@{}}  - \end{tabular}\\ 

% \hline
% ---~\cite{sun2021adversarially}  &NeurIPS      &2021   &Spatial      &---     &\begin{tabular}[c]{@{}c@{}} Improve the robustness of 3D point cloud recognition\\ and demonstrate that local feature learning is \\ideal for the adversarial robustness of point clouds.\end{tabular} & \begin{tabular}[c]{@{}c@{}}  - \end{tabular}\\

% \hline
% PointGLR~\cite{rao2022pointglr}  &IEEE TPAMI     &2022   &Spatial      &---     &\begin{tabular}[c]{@{}c@{}} A hierarchical reasoning scheme among local parts\\, structural proxies and the overall point cloud\end{tabular} & \begin{tabular}[c]{@{}c@{}}  - \end{tabular}\\ 

% \hline
% UP3DETR~\cite{sun2022unsupervised}  &CVPR      &2022   &---     &SUN-RGB-D     &\begin{tabular}[c]{@{}c@{}} A pretext task called random block detection is \\proposed to unsupervised pre-train 3DETRwith \\faster convergence on the object detection task.\end{tabular} & \begin{tabular}[c]{@{}c@{}}  - \end{tabular}\\

\hline
Shape Self-Correction~\cite{chen2021shape}  &CVPR      &2021  &Spatial      &ShapeNet     &\begin{tabular}[c]{@{}c@{}}  A shape self-correction method for 3D shape\\ analysis is developed, which significantly boosts\\ the
performance of supervised models.\end{tabular} & \begin{tabular}[c]{@{}c@{}}  Can be validated on real-world datasets. \end{tabular}\\ 

% \hline
% Scale-graph~\cite{hassani2019unsupervised}     &CVPR       &2019   &Spatial      &ShapeNet     &\begin{tabular}[c]{@{}c@{}}   The unsupervised multitask model is used to jointly \\learn point and shape features in the point cloud \\to train a multi-scale-graph-based encoder.\end{tabular} & \begin{tabular}[c]{@{}c@{}}  The impact of other features, such as normals and\\ geodesics, can be studied, as well as attempts to adapt\\ the model to perform the semantic segmentation task. \end{tabular}\\  

% \hline
% ParaNet~\cite{zhang2020paranet}     &arXiv      &2020       &Spatial      &ModelNet40
% &\begin{tabular}[c]{@{}c@{}} Propose ParaNet, a novel end-to-end deep learning\\ framework, for representing 3D point clouds\\ in a completely regular and nearly lossless manner.\end{tabular} & \begin{tabular}[c]{@{}c@{}}  - \end{tabular}\\ 

% \hline
% PC-FractalDB~\cite{yamada2022point}     &CVPR       &2022       &Spatial      &-      &\begin{tabular}[c]{@{}c@{}} PC-FractalDB pre-trained model is \\especially effective in training with limited data.\end{tabular} & \begin{tabular}[c]{@{}c@{}}  - \end{tabular}\\ 

\hline
VointNet~\cite{hamdi2021voint}  &arXiv      &2021       &Spatial     &ModelNet40       &\begin{tabular}[c]{@{}c@{}} Competitive performance of 3D semantic segmentation \\on shape mesh parts is achieved, and robustness\\ to rotation and occlusion is also improved.\end{tabular} & \begin{tabular}[c]{@{}c@{}}  How well-trained the 2D backbone is for\\  thedownstream 3D task and choose the proper \\viewpoint for segmentation are difficult.\end{tabular}\\  

\hline
CM-CV~\cite{jing2021self} & CVPR        & 2021 & Multi-modality & ModelNet40          & \begin{tabular}[c]{@{}c@{}} Combine cross-modality and cross-view correspondences \\for point cloud understanding. \end{tabular} & \begin{tabular}[c]{@{}c@{}} Insufficient exploration of local correspondence. \end{tabular}\\\hline

I2P-MAE~\cite{zhang2022learning} & arXiv & 2022 & Multi-modality & ShapeNet55 & \begin{tabular}[c]{@{}c@{}} 2D-guided masking and 2D-semantic reconstruction \\ are introduced to transfer the well\\ learned 2D knowledge into 3D domains.\end{tabular} &  \begin{tabular}[c]{@{}c@{}}The influence about the number \\ of 2D saliency maps can be studied.\end{tabular} \\
\hline

Pix4Point~\cite{qian2022pix4point} & arXiv & 2022 & Multi-modality & ImageNet-1K &  \begin{tabular}[c]{@{}c@{}}Improve the performance of standard \\Transformer models for point cloud understanding. \end{tabular} & \begin{tabular}[c]{@{}c@{}}The performance can be promoted by distillation as ~\cite{dong2022autoencoders}.\end{tabular} \\\hline

EPCL~\cite{huang2022frozen} & arXiv & 2022 & Multi-modality & - & \begin{tabular}[c]{@{}c@{}}Require neither 3D pre-training nor \\2D-3D data matching to train the model.\end{tabular} & \begin{tabular}[c]{@{}c@{}}Can be validated on real-world ScanObjectNN.\end{tabular} \\\hline

MVR~\cite{tran2022self} & ACCV        & 2022 & Multi-modality & ModelNet40          & \begin{tabular}[c]{@{}c@{}} Explore both global and local knowledge transfer \\from 2D to 3D. \end{tabular} & \begin{tabular}[c]{@{}c@{}} Obtaining local features by \\reconstructing from global features\\ which may affect local geometry. \end{tabular}\\

\hline
PointCMC~\cite{zhou2022pointcmc} & arXiv        & 2022 & Multi-modality & ShapeNet          & \begin{tabular}[c]{@{}c@{}} Firstly utilize the local-global correspondence \\in multi-modal learning between images and point clouds. \end{tabular} & \begin{tabular}[c]{@{}c@{}} Not taking full advantage of \\point cloud geometric knowledge. \end{tabular}\\

\hline
MV-TER~\cite{gao2021self} & arXiv        & 2021 & Multi-modality & ImageNet1K          & \begin{tabular}[c]{@{}c@{}} Firstly introduce the idea of "transformation equivariant" \\to 3D point cloud understanding. \end{tabular} & \begin{tabular}[c]{@{}c@{}} The number of views on the \\ performance should be studied. \end{tabular}\\

\hline
P2P~\cite{wang2022p2p} & NeurIPS        & 2022 & Multi-modality & ImageNet1K/ImageNet22K          & \begin{tabular}[c]{@{}c@{}} Making full use of knowledge from \\ any pre-trained 2D image models. \end{tabular} & \begin{tabular}[c]{@{}c@{}} Have difficulty in performing 3D tasks \\that concentrates on modality-dependent geometry. \end{tabular}\\

\hline
PointCLIP~\cite{zhang2022pointclip} & CVPR        & 2022 & Multi-modality & ImageNet          & \begin{tabular}[c]{@{}c@{}} Firstly transfer the knowledge of large-scale \\pre-trained image-text model into 3D domain. \end{tabular} & \begin{tabular}[c]{@{}c@{}} Sparse projection method \\and overly simple prompts. \end{tabular}\\

\hline
PointCLIP V2~\cite{zhu2022pointclip} & CVPR        & 2022 & Multi-modality & ImageNet          & \begin{tabular}[c]{@{}c@{}} More realistic projection \\ module and more descriptive prompts. \end{tabular} & \begin{tabular}[c]{@{}c@{}} The performance on \\ outdoor datasets can be studied \end{tabular}\\\hline

ULIP~\cite{xue2022ulip} & arXiv & 2022 & Multi-modality & - & \begin{tabular}[c]{@{}c@{}}ULIP can easily plug in any 3D backbones; ULIP \\can potentially enable more  cross-domain downstream tasks.\end{tabular}  & More modality can be added into this framework. \\\hline

MD~\cite{sun2022self} & IEEE TMM & 2022 & Multi-modality & \begin{tabular}[c]{@{}c@{}} S3DIS, ModelNet-40\\ \& ShapeNet-Part\end{tabular} & \begin{tabular}[c]{@{}c@{}}The mixing operation provides a larger online \\training sample pool, while the disentangling operation enables\\ the model to exploit the geometric prior.\end{tabular}& The experiments on ScanObjectNN can be conducted. \\

\bottomrule[1.5pt]
\end{tabular}%
}
\label{summary_object}
\end{table*}

% \usepackage{graphicx}
\begin{table*}[htp]
\centering

\setlength\tabcolsep {1.5pt}
\caption{Summary of self-supervised learning techniques for point cloud pre-training, focusing on outdoor scene-level point clouds.}
\resizebox{\textwidth}{!}{%
\begin{tabular}{>{\columncolor{C7!50}}l>{\columncolor{C6!50}}c>{\columncolor{C5!50}}c>{\columncolor{C4!50}}c>{\columncolor{C3!50}}c>{\columncolor{C2!50}}c>{\columncolor{C1!50}}c}
\toprule[1.5pt]
Methods   & Venue & Year & Category         & Pre-trained dataset & Contributions                                                                                                                         & Limitations                                                                       \\ \midrule[1pt]
ProposalContrast~\cite{yin2022proposalcontrast} & ECCV        & 2022 & Contrastive learning &  Unlabeled Waymo        & \begin{tabular}[c]{@{}c@{}} Firstly propose to learn 3D representations\\ via contrasting region proposals \end{tabular} & \begin{tabular}[c]{@{}c@{}} 
Need complex pre-processing to produce region proposals.\end{tabular} \\ 

\hline

GCC-3D~\cite{liang2021exploring} & ICCV & 2021 & Contrastive learning & Waymo/nuScenes & \begin{tabular}[c]{@{}c@{}}Integrate clustering harmonization as well as \\ geometry-aware contrast and without static partial views setup. \end{tabular} & \begin{tabular}[c]{@{}c@{}} Need complex pre-processing to produce pseudo instances. \end{tabular}\\ 
\hline

SegContrast~\cite{nunes2022segcontrast} & RAL & 2022 & Contrastive learning & \begin{tabular}[c]{@{}c@{}}SemanticKITTI/\\SemanticPOSS\end{tabular} & \begin{tabular}[c]{@{}c@{}}Learn a more descriptive feature representation and \\performs better using fewer labels compared to \\other contrastive methods. \end{tabular} & \begin{tabular}[c]{@{}c@{}} RANSAC for ground plane might be time-consuming \end{tabular}\\ 
\hline

CO\^{}3~\cite{chen2022co} & arXiv & 2022 & Contrastive learning & DAIR-V2X & \begin{tabular}[c]{@{}c@{}}Build views from both vehicle side and infrastructure side \\as contrastive samples using DAIR-V2X dataset. \end{tabular} & \begin{tabular}[c]{@{}c@{}}The relatively small size of \\ the released cooperation dataset. \end{tabular}\\ 
\hline

SimIPU~\cite{li2022simipu} & AAAI & 2022 & Contrastive learning & KITTI & \begin{tabular}[c]{@{}c@{}}Develop a Multi-Modal Contrastive Learning framework \\ to learn spatial-aware visual representations. \end{tabular} & \begin{tabular}[c]{@{}c@{}}Focus on the visual representations of spatial\\ perception but ignoring the semantic information. \end{tabular}\\ \hline

Voxel-MAE~\cite{hess2022masked}      &arXiv      &2022       &Reconstruction     &-      &\begin{tabular}[c]{@{}c@{}} Firstly apply MAE on outdoor scene-level SSL.\end{tabular} & \begin{tabular}[c]{@{}c@{}} Learn
both spatial and temporal representations\\ useful for multi-object tracking and motion prediction.\end{tabular}\\  

\hline
Voxel-MAE~\cite{min2022voxel}       &arXiv      &2022       &Reconstruction      &Waymo      &\begin{tabular}[c]{@{}c@{}} Propose range-aware random mask and binary \\ voxel classification for SSL on outdoor scene-level SSL.\end{tabular} & \begin{tabular}[c]{@{}c@{}} The single reconstruction loss might be improved.\end{tabular}\\
\hline

S2M2-SSD~\cite{zheng2022boosting} & CVPR & 2022 & Multi-modality & nuScenes/nuImages & \begin{tabular}[c]{@{}c@{}}Combine four different levels of knowledge distillation \\to guide the single-modality model to generate \\simulated multi-modality features. \end{tabular} & \begin{tabular}[c]{@{}c@{}} Take
only single-modality input at inference. \end{tabular}\\ 
\hline

Scene Flow Estimation~\cite{mittal2020just} & CVPR & 2020 & Flow-based & FlyingThing3D & \begin{tabular}[c]{@{}c@{}}Propose the cycle consistency loss to avoid the \\degenerated solution for scene flow.  \end{tabular} & \begin{tabular}[c]{@{}c@{}} Pre-train on the larger Waymo, nuScenes,\\ and ONCE dataset is not explored. \end{tabular}\\ 
\hline

Self-Point-Flow~\cite{li2021self} & CVPR & 2021 & Flow-based & FlyingThing3D/KITTI & \begin{tabular}[c]{@{}c@{}}Formulate the scene flow task as an optimal transportation \\problem under the constraint of one-to-one matching, \\taking into account measures like coordinates, color and \\surface normal to compute the matching cost. \end{tabular} & \begin{tabular}[c]{@{}c@{}} Can be promoted by\\ Piecewise pseudo label generation~\cite{li2022rigidflow}. \end{tabular}\\ 
\hline

RigidFlow~\cite{li2022rigidflow} & CVPR & 2022 & Flow-based & FlyingThing3D/KITTI & \begin{tabular}[c]{@{}c@{}}Present a piecewise rigid motion estimation method \\to generate pseudo labels. \end{tabular} & \begin{tabular}[c]{@{}c@{}} Pseudo label generation method \\ is on the basis of local rigidity assumption. \end{tabular}\\ 
\hline

3D-OGFlow~\cite{ouyang2021occlusion} & arXiv & 2021 & Flow-based & FlyingThing3D & \begin{tabular}[c]{@{}c@{}}Adopt a four-level feature pyramid architecture \\merging two networks across all layers to conduct \\flow estimation and occlusion detection at the same time. \end{tabular} & \begin{tabular}[c]{@{}c@{}} Flow predictors attempt to regress the motion\\ flows of occluded points from those invalid costs~\cite{wang2022estimation}. \end{tabular}\\ 
\hline

Self-supervised Scene Flow~\cite{erccelik20223d} & arXiv & 2022 & Flow-based & KITTI & \begin{tabular}[c]{@{}c@{}}Leverage learned scene flow and motion representation \\to guide the 3D object detection tasks. \end{tabular} & \begin{tabular}[c]{@{}c@{}} Can be applied to more datasets. \end{tabular}\\ 
\hline

OGC~\cite{ouyang2021occlusion} & arXiv & 2022 & Flow-based & KITTI-SF & \begin{tabular}[c]{@{}c@{}}Utilize learned dynamic motion patterns as supervision \\signals to train the network to simultaneously segment \\multiple objects in a single forward pass. \end{tabular} & \begin{tabular}[c]{@{}c@{}}Can be general for other domains such as AR/VR. \end{tabular}\\ 
\hline

Point Cloud Prediction~\cite{mersch2022self} & CoRL & 2022 & Predictive-based & KITTI Odometry & \begin{tabular}[c]{@{}c@{}} Jointly process spatial and temporal information and predict \\ large-scale point clouds without voxelization. \end{tabular} & \begin{tabular}[c]{@{}c@{}} Can be extended to Transformer backbone~\cite{ma2022seqot}. \end{tabular}\\ 

\bottomrule[1.5pt]
\end{tabular}%
}
\label{summary_outdoor}
\end{table*}

% !TEX root = ../bare_jrnl_new_sample4.tex
% \usepackage{graphicx}
\begin{table*}[b]
\centering
\caption{Comparisons of \textbf{few-shot classification} performance on \textbf{ModelNet40} datasets}
\resizebox{\textwidth}{!}{%
\begin{tabular}{lccccccccc}
\toprule[1.5pt]
Methods    & Publication & Years & Types of Methods   & Pre-trained Datasets    & Backbone   & 5way 10shot & 5way 20shot & 10way 10shot & 10way 20shot \\ \midrule[1pt]
\rowcolor{C6!50} OcCo~\cite{wang2021unsupervised} & ICCV & 2021 & Reconstruction-based & ModelNet40 & PointNet & 89.7 ± 1.9 & 92.4 ± 1.6 & 83.9 ± 1.8 & 89.7 ± 1.5 \\
\rowcolor{C6!50} OcCo~\cite{wang2021unsupervised} & ICCV & 2021 & Reconstruction-based & ModelNet40 & DGCNN & 90.6 ± 2.8 & 92.5 ± 1.9 & 82.9 ± 1.3 & 86.5 ± 2.2 \\
\rowcolor{C6!50} MaskSurf~\cite{zhang2022masked} & arXiv & 2022 & Reconstruction-based & ShapeNet55 & Transformer (Transferring features protocol) & 96.5 ± 2.5 & 98.0 ± 1.4 & 93.0 ± 4.1 & 95.3 ± 3.0 \\
\rowcolor{C6!50} MaskSurf~\cite{zhang2022masked} & arXiv & 2022 & Reconstruction-based & ShapeNet55 & Transformer (Linear classification protocol) & 87.1 ± 4.6 & 92.3 ± 4.9 & 89.3 ± 4.2 & 94.9 ± 3.2 \\
\rowcolor{C6!50} MaskSurf~\cite{zhang2022masked} & arXiv & 2022 & Reconstruction-based & ShapeNet55 & Transformer (Non-linear classification protocol) & 95.4 ± 2.9 & 97.6 ± 1.4 & 90.9 ± 4.6 & 94.7 ± 3.3 \\
\rowcolor{C6!50} Point-BERT~\cite{yu2022point} & CVPR        & 2022  & Reconstruction-based   &ShapeNet55      & Transformer      & 94.6 ± 3.1  & 96.3 ± 2.7  & 91.0 ± 5.4   & 92.7 ± 5.1   \\ 
\rowcolor{C6!50} McP-BERT~\cite{fu2022point} & arXiv & 2022 & Reconstruction-based & ShapeNet55 & Transformer & 97.1 ± 1.8 & 98.3 ± 1.2 & 92.4 ± 4.3 & 94.9 ± 3.7 \\
\rowcolor{C6!50} POS-BERT~\cite{fu2022pos} & arXiv & 2022 & Reconstruction-based & ShapeNet55 & Transformer & 96.4 ± 1.9 & 97.0 ± 2.2 & 92.6 ± 4.0 & 94.9 ± 2.9 \\
\rowcolor{C6!50} MAE3D~\cite{jiang2022masked} & arXiv & 2022 & Reconstruction-based & ShapeNet55 & DGCNN & 95.2 ± 3.1 & 97.9 ± 1.6 & 91.1 ± 4.6 & 94.2 ± 3.8 \\
\rowcolor{C6!50} Point-MAE~\cite{pang2022masked} & ECCV & 2022 & Reconstruction-based & ShapeNet55 & Transformer & 96.3 ± 2.5 & 97.8 ± 1.8 & 92.6 ± 4.1 & 95.0 ± 3.0 \\
\rowcolor{C6!50} Point-M2AE~\cite{zhang2022point} & NeurIPS        & 2022  & Reconstruction-based  &ShapeNet55      & Transformer      & 96.8 ± 1.8  & 98.3 ± 4.5  & 92.3 ± 4.5   & 95.0 ± 3.0   \\ 
\rowcolor{C6!50} Point-DAE~\cite{zhang2022point1} & arXiv & 2022 & Reconstruction-based & ShapeNet55 & PointNet (Transferring Features Protocol) & 93.0 ± 3.7 & 94.9 ± 3.3 & 86.7 ± 5.8 & 92.1 ± 4.6 \\
\rowcolor{C6!50} Point-DAE~\cite{zhang2022point1} & arXiv & 2022 & Reconstruction-based & SVD-Pose & PointNet (Transferring Features Protocol) & 92.1 ± 4.6 & 94.9 ± 3.6 & 86.3 ± 4.9 & 91.3 ± 4.4 \\
\rowcolor{C6!50} Point-DAE~\cite{zhang2022point1} & arXiv & 2022 & Reconstruction-based & ShapeNet55 & DGCNN  (Transferring   Features Protocol) & 96.7 ± 2.5 & 97.7 ± 1.6 & 93.0 ± 4.8 & 95.6 ± 2.6 \\
\rowcolor{C6!50} Point-DAE~\cite{zhang2022point1} & arXiv & 2022 & Reconstruction-based & SVD-Pose & DGCNN (Transferring Features Protocol) & 95.3 ± 2.9 & 97.5 ± 1.8 & 91.7 ± 3.9 & 94.8 ± 3.2\\
\rowcolor{C6!50} PointGLR~\cite{rao2022pointglr}    &IEEE TPAMI       &2022       &Reconstruction-based       &-      &PointNet++   &88.3 ± 5.9        & 91.9 ± 4.4        &77.8 ± 2.9     & 81.5 ± 1.9\\

\rowcolor{C5!50} ContrastMPCT~\cite{wang2022self} & RA-L        & 2022 & Contrastive-learning-based   & ShapeNet55 & Transformer         &96.5 ± 1.7    &98.5 ± 1.7     &93.0 ± 2.4     &95.2 ± 2.0 \\
\rowcolor{C4!50} Cover-tree~\cite{sharma2020self}  &NeurIPS      &2020     &Spatial-based        &ShapeNet     &PointNet    &63.2 ± 10.72   &68.90 ± 9.41     &49.15 ± 6.09  &50.10 ± 5.0 \\
\rowcolor{C4!50} Cover-tree~\cite{sharma2020self}  &NeurIPS      &2020     &Spatial-based        &ShapeNet     &DGCNN    &60.0 ± 8.87   &65.70 ± 8.37     &48.50 ± 5.63  &53.0 ± 4.08 \\

\rowcolor{C3!50} I2P-MAE~\cite{zhang2022learning} & arXiv & 2022 & Multi-modality & ShapeNet55 & transformer & 97.0 ± 1.8 & 98.3 ± 1.3 & 92.6 ± 5.0 & 95.5 ± 3.0 \\
\rowcolor{C3!50} EPCL~\cite{huang2022frozen} & arXiv & 2022 & Multi-modality & - & CLIP image encoder & 95.1 ± 2.7 & 97.3 ± 1.6 & 91.1 ± 4.2 & 93.5 ± 3.8\\
\rowcolor{C3!50} PointCMC~\cite{zhou2022pointcmc} & arXiv        & 2022 & Multi-modality   &DGCNN    &ShapeNet         &92.2 ± 5.0     &95.5 ± 3.3     &87.5 ± 5.1    &91.4 ± 3.0 \\
\rowcolor{C3!50} PointCMC~\cite{zhou2022pointcmc} & arXiv        & 2022 & Multi-modality   &RSCNN    &ShapeNet         &93.5 ± 5.1     &95.8 ± 3.4     &90.2 ± 4.3    &92.6 ± 3.3 \\

\bottomrule[1.5pt]
\end{tabular}%
}
\end{table*}

% !TEX root = ../bare_jrnl_new_sample4.tex
% \usepackage{graphicx}
\begin{table*}[b]
\centering
\caption{Comparisons of \textbf{shape classification} performance on \textbf{ModelNet40} datasets}
\resizebox{\textwidth}{!}{%
\begin{tabular}{lccccccccc}
\toprule[1.5pt]
Methods    & Publication & Years & Type of Methods      & Backbone & Pre-train Dataset & Number of Points & Image Views & Accuracy & Accuracy (SVM) \\ \midrule[1pt]
\rowcolor{C6!50} RS~\cite{sauder2019self} & NeurIPS & 2019 & Reconstruction-based & PointNet & ShapeNet55 & 2048 & - & - & 87.3 \\
\rowcolor{C6!50} RS~\cite{sauder2019self} & NeurIPS & 2019 & Reconstruction-based & DGCNN & ShapeNet55 & 2048 & - & - & 90.6 \\
\rowcolor{C6!50} RS~\cite{sauder2019self} & NeurIPS & 2019 & Reconstruction-based & DGCNN & ShapeNet55 & 1024 & - & 92.4 & - \\
\rowcolor{C6!50} OcCo~\cite{wang2021unsupervised} & ICCV & 2021 & Reconstruction-based & PointNet & ModelNet40 & 1024 & - & 90.1 ± 0.1 & - \\
\rowcolor{C6!50} OcCo~\cite{wang2021unsupervised} & ICCV & 2021 & Reconstruction-based & PCN & ModelNet40 & 1024 & - & 90.3 ± 0.2 & - \\
\rowcolor{C6!50} OcCo~\cite{wang2021unsupervised} & ICCV & 2021 & Reconstruction-based & DGCNN & ModelNet40 & 1024 & - & 93.0 ± 0.2 & - \\
\rowcolor{C6!50} MaskSurf~\cite{zhang2022masked} & arXiv & 2022 & Reconstruction-based & Transformer & ShapeNet55 & 1024 & - & 93.2 ± 0.2 & - \\
\rowcolor{C6!50} MaskSurf~\cite{zhang2022masked} & arXiv & 2022 & Reconstruction-based & Transformer (Transferring features protocol) & ShapeNet55 & 1024 & - & 93.4 & - \\
\rowcolor{C6!50} MaskSurf~\cite{zhang2022masked} & arXiv & 2022 & Reconstruction-based & Transformer (Linear classification protocol) & ShapeNet55 & 1024 & - & - & 92.3 ± 0.0 \\
\rowcolor{C6!50} MaskSurf~\cite{zhang2022masked} & arXiv & 2022 & Reconstruction-based & Transformer (Non-linear classification protocol) & ShapeNet55 & 1024 & - & 93.4 ± 0.0 & - \\
\rowcolor{C6!50} Point-BERT~\cite{yu2022point} & CVPR & 2022 & Reconstruction-based & Transformer & ShapeNet55 & 1024 & - & 93.2 & - \\
\rowcolor{C6!50} Point-BERT~\cite{yu2022point} & CVPR & 2022 & Reconstruction-based & Transformer & ShapeNet55 & 4096 & - & 93.4 & - \\
\rowcolor{C6!50} Point-BERT~\cite{yu2022point} & CVPR & 2022 & Reconstruction-based & Transformer & ShapeNet55 & 8192 & - & 93.8 & - \\
\rowcolor{C6!50} McP-BERT~\cite{fu2022point} & arXiv & 2022 & Reconstruction-based & Transformer & ShapeNet55 & 8192 & - & 94.1 & - \\
\rowcolor{C6!50} POS-BERT~\cite{fu2022pos} & arXiv & 2022 & Reconstruction-based & Transformer & ShapeNet55 & 1024 & - & 93.6 & 92.1 \\
\rowcolor{C6!50} POS-BERT~\cite{fu2022pos} & arXiv & 2022 & Reconstruction-based  & Transformer (voting strategy) & ShapeNet55 & 1024 & - & 93.8  & - \\
\rowcolor{C6!50} ContrastMPCT~\cite{wang2022self} & IEEE RAL & 2022 & Reconstruction-based & DGCNN & ShapeNet55 & 1024 & - & 93.7 & - \\
\rowcolor{C6!50} ContrastMPCT~\cite{wang2022self} & IEEE RAL & 2022 & Reconstruction-based & PCT & ShapeNet55 & 1024 & - & 93.6 & - \\
\rowcolor{C6!50} MAE3D~\cite{jiang2022masked} & arXiv & 2022 & Reconstruction-based & PointNet & ShapeNet55 & 1024 & - & 90.6 & - \\
\rowcolor{C6!50} MAE3D~\cite{jiang2022masked} & arXiv & 2022 & Reconstruction-based & DGCNN & ShapeNet55 & 1024 & - & 93.4 & - \\
\rowcolor{C6!50} MAE3D~\cite{jiang2022masked} & arXiv & 2022 & Reconstruction-based & DGCNN & ShapeNet55 & 2048 & - & 92.5 & - \\
\rowcolor{C6!50} MAE3D~\cite{jiang2022masked} & arXiv & 2022 & Reconstruction-based & DGCNN & ModelNet40 & 1024 & - & 92.1 & - \\
\rowcolor{C6!50} Point-MAE~\cite{pang2022masked} & ECCV & 2022 & Reconstruction-based & Transformer & ShapeNet55 & 1024 & - & 93.8 & - \\
\rowcolor{C6!50} Point-MAE~\cite{pang2022masked} & ECCV & 2022 & Reconstruction-based & Transformer & ShapeNet55 & 8192 & - & 94.0 & - \\
\rowcolor{C6!50} Point-M2AE~\cite{zhang2022point} & NeurIPS        & 2022  & Reconstruction-based &  Transformer        &   ShapeNet55                & 1024             & -           & 94.0   & 92.9     \\
\rowcolor{C6!50} CP-Net~\cite{xu2022cp} & arXiv & 2022 & Reconstruction-based & PointNet++ & ModelNet40 & 1024 & - & 92.5 & - \\
\rowcolor{C6!50} CP-Net~\cite{xu2022cp}  & arXiv & 2022 & Reconstruction-based & PointNet++ & ShapeNet55 & 1024 & - & 91.9 & - \\
\rowcolor{C6!50} Point-DAE~\cite{zhang2022point1} & arXiv & 2022 & Reconstruction-based & PointNet (Linear Classification Protocol) & ShapeNet55 & 1024 & - & - & 89.3 ± 0.1 \\
\rowcolor{C6!50} Point-DAE~\cite{zhang2022point1} & arXiv & 2022 & Reconstruction-based & DGCNN  (Linear   Classification Protocol) & ShapeNet55 & 1024 & - & - & 91.9 ± 0.2 \\
\rowcolor{C6!50} Point-DAE~\cite{zhang2022point1} & arXiv & 2022 & Reconstruction-based & PointNet (Linear Classification Protocol) & SVD-Pose & 1024 & - & - & 89.0 ± 0.1 \\
\rowcolor{C6!50} Point-DAE~\cite{zhang2022point1} & arXiv & 2022 & Reconstruction-based & DGCNN  (Linear   Classification Protocol) & SVD-Pose & 1024 & - & - & 90.7 ± 0.1 \\
\rowcolor{C6!50} Point-DAE~\cite{zhang2022point1} & arXiv & 2022 & Reconstruction-based & PointNet (Transferring Features Protocol) & ShapeNet55 & 1024 & - & 90.6 ± 0.1 & - \\
\rowcolor{C6!50} Point-DAE~\cite{zhang2022point1} & arXiv & 2022 & Reconstruction-based & DGCNN  (Transferring   Features Protocol) & ShapeNet55 & 1024 & - & 93.3 ± 0.1 & - \\
\rowcolor{C6!50} Point-DAE~\cite{zhang2022point1} & arXiv & 2022 & Reconstruction-based & PointNet (Transferring Features Protocol) & SVD-Pose & 1024 & - & 90.6 ± 0.1 & - \\
\rowcolor{C6!50} Point-DAE~\cite{zhang2022point1} & arXiv & 2022 & Reconstruction-based & DGCNN (Transferring Features Protocol) & SVD-Pose & 1024 & - & 93.0 ± 0.1 & - \\
\rowcolor{C6!50} SeRP~\cite{garg2022serp} & arXiv & 2022 & Reconstruction-based & SeRP-Transformer (SeRP-Net) & ShapeNet55 & 1024 & - & 89.1 & - \\
\rowcolor{C6!50} SeRP~\cite{garg2022serp} & arXiv & 2022 & Reconstruction-based & SeRP-Transformer (VASP) & ShapeNet55 & 1024 & - & 87.9 & - \\
\rowcolor{C6!50} SeRP~\cite{garg2022serp} & arXiv & 2022 & Reconstruction-based & SeRP-PointNet ($\delta$ learn) & ShapeNet55 & 1024 & - & 84.1 & - \\
\rowcolor{C6!50} SeRP~\cite{garg2022serp}& arXiv & 2022 & Reconstruction-based & SeRP-PointNet ($cd\ell  2$ learn) & ShapeNet55 & 1024 & - & 84.1 & - \\
\rowcolor{C6!50} UAE~\cite{zhang2022upsampling} & arXiv & 2022 & Reconstruction-based & DGCNN (Unsupervised Transfer Learning) & ShapeNet55 & 1024 & - & 92.9 & - \\
\rowcolor{C6!50} UAE~\cite{zhang2022upsampling} & arXiv & 2022 & Reconstruction-based & DGCNN (Supervised Fine-tuning) & ShapeNet55 & 1024 & - & 93.2 & - \\
\rowcolor{C6!50} ParAE~\cite{eckart2021self} & CVPR & 2021 & Reconstruction-based & PointNet & ShapeNet55 & 1024 & - & - & 90.3 \\
\rowcolor{C6!50} ParAE~\cite{eckart2021self} & CVPR & 2021 & Reconstruction-based & DGCNN & ShapeNet55 & 1024 & - & - & 91.6 \\
\rowcolor{C6!50} ParAE~\cite{eckart2021self} & CVPR & 2021 & Reconstruction-based & PointNet (Fully Supervised) & ShapeNet55 & 1024 & - & 90.5 & - \\
\rowcolor{C6!50} ParAE~\cite{eckart2021self} & CVPR & 2021 & Reconstruction-based & DGCNN (Fully Supervised) & ShapeNet55 & 1024 & - & 92.9 & -\\
\rowcolor{C6!50} PointGLR~\cite{rao2022pointglr} & IEEE TPAMI & 2022 & Reconstruction-based & PointNet++ & ModelNet40 & 1024 & - & 92.3 (Small)/93.7 (Large)  & -\\
\rowcolor{C6!50} PointGLR~\cite{rao2022pointglr} & IEEE TPAMI & 2022 & Reconstruction-based & PointNet++ & ModelNet40 & 1024 & - & 92.2 (Small)/92.9 (Large)  & -\\
\rowcolor{C6!50} IAE~\cite{yan2022implicit}     &arXiv      &2022       &Reconstruction-based     &DGCNN      &ShapeNet      &1024      &-     &94.2(+1.2)      &-\\
\rowcolor{C6!50} IAE~\cite{yan2022implicit}     &arXiv      &2022       &Reconstruction-based     &DGCNN      &ShapeNet      &2048      &-      &-  &92.1(+0.9) \\

\rowcolor{C5!50} DepthContrast~\cite{zhang2021self} & ICCV        & 2021 & Contrastive-learning-based & PointNet++ & ScanNet-vid         & 1024    &-     &-     &85.4 \\
\rowcolor{C5!50} GSIR~\cite{chen2021unsupervised}    & ICCV       & 2021 & Contrastive-learning-based   &DGCNN    &ModelNet40         & 1024    &-     &-     &90.36 \\
\rowcolor{C5!50} ContrastMPCT~\cite{wang2022self} & RA-L        & 2022 & Contrastive-learning-based   &Transformer    &ShapeNet         & 1024    &-     &93.30     &- \\
\rowcolor{C5!50} ConClu~\cite{mei2022unsupervised} & ICIP        & 2022 & Contrastive-learning-based   &DGCNN    &ModelNet40         & 2048    &-     &-    &91.6 \\

% Distillation with Contrast~\cite{fu2022distillation} & Arxiv        & 2022 & Contrastive-learning-based   &Transformer    &ShapeNet55         & 2048    &-     &93.4    &91.4 \\

\rowcolor{C5!50} DCGLR~\cite{fu2022distillation} & arXiv      &2022	&Contrastive-learning-based  & 3D-ViT     & ShapeNet	     &2048	     &-        &92.2         	& 91.3        \\
\rowcolor{C5!50} DCGLR~\cite{fu2022distillation} & arXiv      &2022	&Contrastive-learning-based  & PCT    & ShapeNet	     &2048	     &-        &93.4         	& 91.4        \\

\rowcolor{C4!50} STRL~\cite{huang2021spatio} & ICCV       &2021       &Spatial-Temporal-based   & PointNet    &   ShapeNet          &2048           &-       & -     & 88.3    \\
\rowcolor{C4!50} STRL~\cite{huang2021spatio} & ICCV       &2021       &Spatial-Temporal-based   & DGCNN    &   ShapeNet            &2048           &-       & -     & 90.9    \\

% MD~\cite{sun2022self}   &IEEE TMM       &2022       &Spatial-based      & GNN       &ModelNet40           &1024       &-       &93.4          &-\\
% MD~\cite{sun2022self}   &IEEE TMM       &2022       &Spatial-based      & GNN       &S3DIS          &1024       &-        &92.8         &-   \\
\rowcolor{C4!50} SL3D~\cite{cendra2022sl3d}    &NeuIPS        &2022         &Spatial-based      &Point Transformer          & ModelNet40      &2048       &-      &77.2      &-  \\
\rowcolor{C4!50} Rotation prediction~\cite{poursaeed2020self}     &IEEE 3DV    &2020       &Spatial-based     &PointNet      &ShapeNet     & 36 angles     &-         &-  &88.6  \\
\rowcolor{C4!50} Rotation prediction~\cite{poursaeed2020self}     &IEEE 3DV    &2020       &Spatial-based     &DGCNN      &ShapeNet     & 18 angles    &-         &-    &90.75 \\

\rowcolor{C4!50} Shape Self-Correction~\cite{chen2021shape}       &CVPR      &2021       &Spatial-based      &PointNet       &ShapeNet       &1024       &-        &-        &90.9 \\
\rowcolor{C4!50} Shape Self-Correction~\cite{chen2021shape}       &CVPR      &2021       &Spatial-based      &RSCNN       &ShapeNet       &1024       &-        &-       &94.3\\

\rowcolor{C4!50} NMI~\cite{hassani2019unsupervised}      &CVPR     &2019       &Spatial-based      &-   &ShapeNet      &2048       &-      &-      &89.1\\
\rowcolor{C4!50} ParaNet-SC (JO)~\cite{zhang2020paranet}     &arXiv      &2020       &Spatial-based      &DGCNN      &ModelNet40      &1024       &-        &93.1        &-\\
\rowcolor{C4!50} ParaNet-SC (Off-IO)~\cite{zhang2020paranet}     &arXiv      &2020       &Spatial-based      &DGCNN      &ModelNet40      &1024       &-       &92.7&-      \\
\rowcolor{C4!50} ParaNet-SC (On-IO)~\cite{zhang2020paranet}     &arXiv      &2020       &Spatial-based      &DGCNN      &ModelNet40      &1024       &-       &92.9      &-\\
\rowcolor{C3!50} VointNet~\cite{hamdi2021voint}      &arXiv      &2021       &Multi-modality     &-    &ModelNet40    & ViT-B      &12      &92.8      &-  \\
\rowcolor{C3!50} I2P-MAE~\cite{zhang2022learning}& arXiv & 2022 & Multi-modality & transformer (No Voting) & ShapeNet55 & 1024 & - & 93.7 & 93.4 \\
\rowcolor{C3!50} I2P-MAE~\cite{zhang2022learning} & arXiv & 2022 & Multi-modality & transformer (Voting) & ShapeNet55 & 1024 & - & 94.1 & - \\
\rowcolor{C3!50} EPCL~\cite{huang2022frozen} & arXiv & 2022 & Multi-modality & CLIP image encoder & - & - & - & 92.9 & -\\
\rowcolor{C3!50} Multi-View Rendering~\cite{tran2022self}      &ACCV      &2022       &Multi-modality     &DGCNN+ResNet50    &ModelNet40    &1024      &-      &$93.2\pm0.1$      &91.7  \\
\rowcolor{C3!50} PointCMC~\cite{zhou2022pointcmc} & arXiv        & 2022 & Multi-modality   &DGCNN    &ShapeNet         & 1024    &-     &-    &91.7 \\
\rowcolor{C3!50} PointCMC~\cite{zhou2022pointcmc} & arXiv        & 2022 & Multi-modality   &RSCNN    &ShapeNet         & 1024    &-     &-    &91.5 \\
\rowcolor{C3!50} MV-TER~\cite{gao2021self} & arXiv        & 2021 & Multi-modality   &MVCNN    &ImageNet1K         &-    &12     &95.5    &- \\
\rowcolor{C3!50} MV-TER~\cite{gao2021self} & arXiv        & 2021 & Multi-modality   &GVCNN    &ImageNet1K         &-    &12     &97.0    &- \\
\rowcolor{C3!50} P2P~\cite{wang2022p2p} & NeurIPS        & 2022 & Multi-modality   &ResNet101    &ImageNet1K         &4096    &     &93.1    &- \\
\rowcolor{C3!50} P2P~\cite{wang2022p2p} & NeurIPS        & 2022 & Multi-modality   &HorNet-L    &ImageNet22K         &4096    &     &94.0    &- \\
\rowcolor{C3!50} PointCLIP~\cite{zhang2022pointclip} & CVPR        & 2022 & Multi-modality   &ResNet50 \& Transformer    &ImageNet        &1024    &6     &23.78 (Zero-shot)    &- \\
\rowcolor{C3!50} PointCLIP~\cite{zhang2022pointclip} & CVPR        & 2022 & Multi-modality   &ResNet101 \& Transformer    &ImageNet         &1024    &10     &87.20 (16-shot)    &- \\
\rowcolor{C3!50} PointCLIP V2~\cite{zhu2022pointclip} & CVPR        & 2022 & Multi-modality   &ViT-B$\backslash$16  \& Transformer    &ImageNet         &1024    &10     & 64.22 (Zero-shot)    &- \\
\rowcolor{C3!50} PointCLIP V2~\cite{zhu2022pointclip} & CVPR        & 2022 & Multi-modality   &ViT-B$\backslash$16  \& Transformer    &ImageNet         &1024    &10     & 89.55 (16-shot)    &- \\
\rowcolor{C3!50} ULIP~\cite{xue2022ulip} & arXiv & 2022 & Multi-modality & PointNet++(ssg) & - & - & - & 93.4 & - \\
\rowcolor{C3!50} ULIP~\cite{xue2022ulip} & arXiv & 2022 & Multi-modality & PointBERT & - & - & - & 94.1 & - \\
\rowcolor{C3!50} ULIP~\cite{xue2022ulip} & arXiv & 2022 & Multi-modality & PointMLP & - & - & - & 94.3 & - \\
\rowcolor{C3!50} ULIP~\cite{xue2022ulip} & arXiv & 2022 & Multi-modality & PointMLP (voting technique) & - & - & - & 94.7 & - \\
\rowcolor{C3!50} MD~\cite{sun2022self} & IEEE TMM & 2022 & Multi-modality & - & S3DIS & 4096 & - & 92.8 & - \\
\rowcolor{C3!50} MD~\cite{sun2022self} & IEEE TMM & 2022 & Multi-modality & - & ShapeNet-Part & 1024 & - & 92.8 & - \\
\rowcolor{C3!50} MD~\cite{sun2022self} & IEEE TMM & 2022 & Multi-modality & - & ModelNet-40 & 1024 & - & 93.4 & - \\
\rowcolor{C3!50} MD~\cite{sun2022self} & IEEE TMM & 2022 & Multi-modality & PointNet++ & ShapeNet-Part & 1024 & - & 92.2 & - \\
\rowcolor{C3!50} MD~\cite{sun2022self} & IEEE TMM & 2022 & Multi-modality & PointNet++ & ModelNet-40 & 1024 & - & 92.6 & - \\
\rowcolor{C3!50} MD~\cite{sun2022self} & IEEE TMM & 2022 & Multi-modality & OGNet & ShapeNet-Part & 1024 & - & 93.4 & - \\
\rowcolor{C3!50} MD~\cite{sun2022self} & IEEE TMM & 2022 & Multi-modality & OGNet & ModelNet-40 & 1024 & - & 93.3 & - \\
\rowcolor{C3!50} MD~\cite{sun2022self} & IEEE TMM & 2022 & Multi-modality & PAConv & ShapeNet-Part & 1024 & - & 92.7 & - \\
\rowcolor{C3!50} MD~\cite{sun2022self} & IEEE TMM & 2022 & Multi-modality & PAConv & ModelNet-40 & 1024 & - & 92.8 & - \\
\rowcolor{C3!50} MD~\cite{sun2022self} & IEEE TMM & 2022 & Multi-modality & Point Transformer & ShapeNet-Part & 1024 & - & 92.0 & - \\
\rowcolor{C3!50} MD~\cite{sun2022self} & IEEE TMM & 2022 & Multi-modality & Point Transformer & ModelNet-40 & 1024 & - & 92.1 & - \\
\rowcolor{C3!50} MD~\cite{sun2022self} & IEEE TMM & 2022 & Multi-modality & Point Cloud Transformer & ShapeNet-Part & 1024 & - & 93.1 & - \\
\rowcolor{C3!50} MD~\cite{sun2022self} & IEEE TMM & 2022 & Multi-modality & Point Cloud Transformer & ModelNet-40 & 1024 & - & 93.2 & - \\

\bottomrule[1.5pt]
\end{tabular}%
}
\end{table*}

% !TEX root = ../bare_jrnl_new_sample4.tex
% \usepackage{graphicx}
\begin{table*}[htp]
\centering
\caption{Comparisons of \textbf{shape classification} performance on \textbf{ScanObjectNN} datasets}
\resizebox{\textwidth}{!}{%
\begin{tabular}{lccccccccc}
\toprule[1.5pt]
Methods    & Publication & Years & Types of Methods     & Backbone    & Pre-train Datasets & Image Views & Accuracy on OBJ-BG & Accuracy on OBJ-ONLY & Accuracy on PB-T50-RS \\ \midrule[1pt]
\rowcolor{C6!50} MaskSurf~\cite{zhang2022masked} & arXiv & 2022 & Reconstruction-based & Transformer & ShapeNet55 & - & 90.76 ± 0.53 & 88.74 ± 0.23 & 85.35 ± 0.24 \\
\rowcolor{C6!50} MaskSurf~\cite{zhang2022masked} & arXiv & 2022 & Reconstruction-based & Transformer (Transferring features protocol) & ShapeNet55 & - & 91.22 & 89.17 & 85.81 \\
\rowcolor{C6!50} MaskSurf~\cite{zhang2022masked} & arXiv & 2022 & Reconstruction-based & Transformer (Linear classification protocol) & ShapeNet55 & - & 82.07 ± 0.00 & 83.48 ± 0.00 & 72.59 ± 0.00 \\
\rowcolor{C6!50} MaskSurf~\cite{zhang2022masked} & arXiv & 2022 & Reconstruction-based & Transformer (Non-linear classification protocol) & ShapeNet55 & - & 84.45 ± 0.21 & 86.45 ± 0.08 & 76.48 ± 0.09 \\
\rowcolor{C6!50} Point-BERT~\cite{yu2022point} & CVPR        & 2022  & Reconstruction-based & Transformer & ShapeNet55         & -           & 87.43              & 88.12                & 83.07 \\  
\rowcolor{C6!50} McP-BERT~\cite{fu2022point} & arXiv & 2022 & Reconstruction-based & Transformer & ShapeNet55 & - & 88.98 & 90.02 & 84.28 \\
\rowcolor{C6!50} POS-BERT~\cite{fu2022pos} & arXiv & 2022 & Reconstruction-based & Transformer & ShapeNet55 & - & 90.88 & 90.88 & 83.21 \\
\rowcolor{C6!50} MAE3D~\cite{jiang2022masked} & arXiv & 2022 & Reconstruction-based & DGCNN & ShapeNet55 & - & - & - & 86.20 \\
\rowcolor{C6!50} Point-MAE~\cite{pang2022masked} & ECCV       & 2022  & Reconstruction-based & Transformer & ShapeNet55         & -           & 90.02              & 88.29                & 85.18\\  
\rowcolor{C6!50} Point-M2AE~\cite{zhang2022point} & NeurIPS       & 2022  & Reconstruction-based & Transformer & ShapeNet55         & -           & 91.22              & 88.81                & 86.43 \\
\rowcolor{C6!50} CP-Net~\cite{xu2022cp} & arXiv & 2022 & Reconstruction-based & PointNet++ & ShapeNet55 & - & - & 87.90 & - \\
\rowcolor{C6!50} Point-DAE~\cite{zhang2022point1} & arXiv & 2022 & Reconstruction-based & PointNet (Linear Classification Protocol) & ShapeNet55 & - & 78.10 ± 0.30 & - & - \\
\rowcolor{C6!50} Point-DAE~\cite{zhang2022point1} & arXiv & 2022 & Reconstruction-based & DGCNN  (Linear   Classification Protocol) & ShapeNet55 & - & 87.90 ± 0.20 & - & - \\
\rowcolor{C6!50} Point-DAE~\cite{zhang2022point1} & arXiv & 2022 & Reconstruction-based & PointNet (Linear Classification Protocol) & SVD-Pose & - & 77.80 ± 0.10 & - & - \\
\rowcolor{C6!50} Point-DAE~\cite{zhang2022point1} & arXiv & 2022 & Reconstruction-based & DGCNN  (Linear   Classification Protocol) & SVD-Pose & - & 82.40 ± 0.90 & - & - \\
\rowcolor{C6!50} Point-DAE~\cite{zhang2022point1} & arXiv & 2022 & Reconstruction-based & PointNet (Transferring Features Protocol) & ShapeNet55 & - & 80.20 ± 0.20 & - & - \\
\rowcolor{C6!50} Point-DAE~\cite{zhang2022point1} & arXiv & 2022 & Reconstruction-based & DGCNN (Transferring Features Protocol) & ShapeNet55 & - & 92.10 ± 0.20 & - & - \\
\rowcolor{C6!50} Point-DAE~\cite{zhang2022point1} & arXiv & 2022 & Reconstruction-based & PointNet (Transferring Features Protocol) & SVD-Pose & - & 80.20 ± 0.20 & - & - \\
\rowcolor{C6!50} Point-DAE~\cite{zhang2022point1} & arXiv & 2022 & Reconstruction-based & DGCNN (Transferring Features Protocol) & SVD-Pose & - & 89.90 ± 0.30 & - & -\\
\rowcolor{C6!50} PointGLR~\cite{rao2022pointglr} & IEEE TPAMI & 2022 & Reconstruction-based & PointNet++ & ScanObjectNN & - & 87.2 & - & -\\
\rowcolor{C6!50} PointGLR~\cite{rao2022pointglr} & IEEE TPAMI & 2022 & Reconstruction-based & RSCNN & ScanObjectNN & - & 86.9 & - & -\\

\rowcolor{C5!50} ContrastMPCT~\cite{wang2022self} & RA-L        & 2022 & Contrastive-learning-based   &Transformer    &ShapeNet55         & -    &90.42     &90.15     &85.50 \\

\rowcolor{C3!50} VointNet~\cite{hamdi2021voint}  &arXiv      &2021       &Multi-modality     &ViT-B      &ScanObjectNN      &12      &93.7       &94.0       &-\\
\rowcolor{C3!50} I2P-MAE~\cite{zhang2022learning} & arXiv & 2022 & Multi-modality & transformer & ShapeNet55 & - & 94.15 & 91.57 & 90.11 \\
\rowcolor{C3!50} Pix4Point~\cite{qian2022pix4point} & arXiv & 2022 & Multi-modality & PViT & ImageNet-1K & - & - & - & 85.70 \\
\rowcolor{C3!50} Pix4Point~\cite{qian2022pix4point} & arXiv & 2022 & Multi-modality & PViT+Pix4Point & ImageNet-1K & - & - & - & 87.90\\
\rowcolor{C3!50} MVR~\cite{tran2022self}      &ACCV      &2022       &Multi-modality     &DGCNN \& ResNet50    &ModelNet40    &-      &84.50 ± 0.60      &84.30 ± 0.60    &-     \\
\rowcolor{C3!50} P2P~\cite{wang2022p2p} & NeurIPS        & 2022 & Multi-modality   &ResNet101    &ImageNet1K         &-    &-     &-    &87.4 \\
\rowcolor{C3!50} P2P~\cite{wang2022p2p} & NeurIPS        & 2022 & Multi-modality   &HorNet-L    &ImageNet22K         &-    &-     &-    &89.3 \\
\rowcolor{C3!50} PointCLIP~\cite{zhang2022pointclip} & CVPR        & 2022 & Multi-modality   &ResNet50 \& Transformer    &ImageNet         &6    &19.28 (Zero-shot)     & 21.34 (Zero-shot)    &15.38 (Zero-shot) \\
\rowcolor{C3!50} PointCLIP V2~\cite{zhu2022pointclip} & CVPR        & 2022 & Multi-modality   &ViT-B$\backslash$16 \& Transformer    &ImageNet         &10    &41.22 (Zero-shot)     & 50.09 (Zero-shot)    &35.36 (Zero-shot) \\

\bottomrule[1.5pt]
\end{tabular}%
}
\end{table*}
% !TEX root = ../bare_jrnl_new_sample4.tex
% \usepackage{graphicx}
\begin{table*}[htp]
\centering
\caption{Comparisons of \textbf{part segmentation} performance on \textbf{ShapeNetPart} datasets}
\resizebox{\textwidth}{!}{%
\begin{tabular}{lcccccccc}
\toprule[1.5pt]
Methods    & Publication & Years & Types of Methods     & Backbone    & Pre-train Datasets & Image Views & Class mIoU & Instance mIoU \\  \midrule[1pt]
\rowcolor{C6!50} RS~\cite{sauder2019self} & NeurIPS & 2019 & Reconstruction-based & DGCNN & ShapeNet55 & - & - & 85.3\\
\rowcolor{C6!50} OcCo~\cite{wang2021unsupervised} & ICCV & 2021 & Reconstruction-based & PointNet & ModelNet40 & - & 83.4 ± 1.9 & - \\
\rowcolor{C6!50} OcCo~\cite{wang2021unsupervised} & ICCV & 2021 & Reconstruction-based & PCN & ModelNet40 & - & 82.3 ± 2.4 & - \\
\rowcolor{C6!50} OcCo~\cite{wang2021unsupervised} & ICCV & 2021 & Reconstruction-based & DGCNN & ModelNet40 & - & 85.0 ± 1.0 & - \\
\rowcolor{C6!50} MaskSurf~\cite{zhang2022masked} & arXiv & 2022 & Reconstruction-based & Transformer (Transferring features protocol) & ShapeNet55 & 9 & 84.4 & 86.1 \\
\rowcolor{C6!50} MaskSurf~\cite{zhang2022masked} & arXiv & 2022 & Reconstruction-based & Transformer (Non-linear classification protocol) & ShapeNet55 & 9 & 83.3 & 85.3 \\
\rowcolor{C6!50} Point-BERT~\cite{yu2022point} & CVPR        & 2022  & Reconstruction-based & Transformer & ShapeNet55        & -     & 84.1      & 85.6 \\ 
\rowcolor{C6!50} McP-BERT~\cite{fu2022point} & arXiv & 2022 & Reconstruction-based & Transformer & ShapeNet55 & - & 84.4 & 86.1 \\
\rowcolor{C6!50} POS-BERT~\cite{fu2022pos} & arXiv & 2022 & Reconstruction-based & Transformer & ShapeNet55 & - & 84.2 & 86.0 \\
\rowcolor{C6!50} Point-MAE~\cite{pang2022masked} & ECCV        & 2022  & Reconstruction-based & Transformer & ShapeNet55         & -         & -       & 86.1 \\
\rowcolor{C6!50} Point-M2AE~\cite{zhang2022point} & NeurIPS        & 2022  & Reconstruction-based & Transformer & ShapeNet55         & -       & 84.86     &86.51 \\
\rowcolor{C6!50} CP-Net~\cite{xu2022cp} & arXiv & 2022 & Reconstruction-based & PointNet++ & ShapeNetPart & - & 76.4 (5$\%$ of training annotations) & 81.5  \\
\rowcolor{C6!50} CP-Net~\cite{xu2022cp} & arXiv & 2022 & Reconstruction-based & PointNet++ & ShapeNetPart & - & 76.4 (10$\%$ of training annotations) & 81.6  \\
\rowcolor{C6!50} CP-Net~\cite{xu2022cp} & arXiv & 2022 & Reconstruction-based & PointNet++ & ShapeNetPart & - & 78.8 (50$\%$ of training annotations) & 82.5  \\
\rowcolor{C6!50} CP-Net~\cite{xu2022cp} & arXiv & 2022 & Reconstruction-based & PointNet++ & ShapeNet55 & - & - & 79.3 (1$\%$ of training data) \\
\rowcolor{C6!50} CP-Net~\cite{xu2022cp} & arXiv & 2022 & Reconstruction-based & PointNet++ & ShapeNet55 & - & - & 81.2 (5$\%$ of training data) \\
\rowcolor{C6!50} Point-DAE~\cite{zhang2022point1} & arXiv & 2022 & Reconstruction-based & PointNet & ShapeNet55 & - & 84.7 ± 0.1 & - \\
\rowcolor{C6!50} Point-DAE~\cite{zhang2022point1} & arXiv & 2022 & Reconstruction-based & DGCNN & ShapeNet55 & - & 85.9 ± 0.1 & - \\
\rowcolor{C6!50} Point-DAE~\cite{zhang2022point1} & arXiv & 2022 & Reconstruction-based & PointNet & SVD-Pose & - & 84.4 ± 0.1 & - \\
\rowcolor{C6!50} Point-DAE~\cite{zhang2022point1} & arXiv & 2022 & Reconstruction-based & DGCNN & SVD-Pose & - & 85.7 ± 0.1 & - \\
\rowcolor{C6!50} UAE~\cite{zhang2022upsampling}& arXiv & 2022 & Reconstruction-based & DGCNN (Unsupervised Transfer Learning) & ShapeNet55 & 14 & - & 85.0 \\
\rowcolor{C6!50} UAE~\cite{zhang2022upsampling} & arXiv & 2022 & Reconstruction-based & DGCNN (Supervised Fine-tuning) & ShapeNet55 & 14 & - & 85.6\\
\rowcolor{C6!50} PointGLR~\cite{rao2022pointglr}     &IEEE TPAMI      &2020       &Reconstruction-based      &PointNet++        &-       &-    &89.6       &71.9\\

\rowcolor{C5!50} PointContrast~\cite{xie2020pointcontrast} & ECCV        & 2020  & Contrastive-learning-based & SR-UNet & ScanNet         & -       & -     &74.0 (1$\%$ of train data) \\
\rowcolor{C5!50} PointContrast~\cite{xie2020pointcontrast} & ECCV        & 2020  & Contrastive-learning-based & SR-UNet & ScanNet         & -       & -     &79.9 (5$\%$ of train data) \\
\rowcolor{C5!50} PointContrast~\cite{xie2020pointcontrast} & ECCV        & 2020  & Contrastive-learning-based & SR-UNet & ScanNet         & -       & -     &85.1 \\
\rowcolor{C5!50} GSIR~\cite{chen2021unsupervised}    & ICCV       & 2021 & Contrastive-learning-based   &DGCNN    &ShapeNetPart      &-     &-     &71.6 (1$\%$ of train data) \\
\rowcolor{C5!50} GSIR~\cite{chen2021unsupervised}    & ICCV       & 2021 & Contrastive-learning-based   &DGCNN    &ShapeNetPart      &-     &-     &78.2 (5$\%$ of train data) \\
\rowcolor{C5!50} ContrastMPCT~\cite{wang2022self} & RA-L        & 2022 & Contrastive-learning-based   &Transformer    &ShapeNet         &-       &-     &86.2 \\
\rowcolor{C5!50} ConClu~\cite{mei2022unsupervised} & ICIP        & 2022 & Contrastive-learning-based   &DGCNN    &ModelNet40      &-     &85.4    &- \\

% MD~\cite{sun2022self}   &IEEE TMM       &2022       &Spatial-based      & GNN       & S3DIS     &-      &-        &85.37 \\
% MD~\cite{sun2022self}   &IEEE TMM       &2022       &Spatial-based      & GNN       & ShapeNetPart     &-      &-        &85.37 \\

\rowcolor{C4!50} Shape Self-Correction~\cite{chen2021shape}       &CVPR      &2021       &Spatial-based      &PointNet       &ShapeNet       &-      &-      &69.7 (1$\%$ of train data)\\
\rowcolor{C4!50} Shape Self-Correction~\cite{chen2021shape}       &CVPR      &2021       &Spatial-based      &PointNet       &ShapeNet       &-      &-      &74.0 (5$\%$ of train data)\\
\rowcolor{C4!50} Shape Self-Correction~\cite{chen2021shape}       &CVPR      &2021       &Spatial-based      &RSCNN       &ShapeNet       &-      &-      & 74.1 (1$\%$ of train data)\\
\rowcolor{C4!50} Shape Self-Correction~\cite{chen2021shape}       &CVPR      &2021       &Spatial-based      &RSCNN       &ShapeNet       &-      &-      & 80.1 (5$\%$ of train data)\\

\rowcolor{C4!50} NMI~\cite{hassani2019unsupervised}      &CVPR       &2019        &Spatial-based        &-   &ShapeNet       &-         &-        & 77.7\\

\rowcolor{C3!50} VointNet~\cite{hamdi2021voint}      &arXiv      &2021    &Multi-modality     &ViT-B      &ShapeNetPart         &12      &-      &81.2\\
\rowcolor{C3!50} I2P-MAE~\cite{zhang2022learning} & arXiv & 2022 & Multi-modality & transformer & ShapeNet55 & - & 85.2 & 86.8 \\
\rowcolor{C3!50} Pix4Point~\cite{qian2022pix4point} & arXiv & 2022 & Multi-modality & PViT & ImageNet-1K & - & 83.7 & 85.7 \\
\rowcolor{C3!50} Pix4Point~\cite{qian2022pix4point} & arXiv & 2022 & Multi-modality & PViT+Pix4Point & ImageNet-1K & - & 85.6 & 86.8\\
\rowcolor{C3!50} CM-CV~\cite{jing2021self} & CVPR        & 2021  & Multi-modality & DGCNN \& ResNet18 & ModelNet40         &-       & 79.1     &83.7 \\
\rowcolor{C3!50} MVR~\cite{tran2022self}      &ACCV      &2022       &Multi-modality     &DGCNN \& ResNet50    &ModelNet40    &-      &84.7 ± 0.1    &- \\
\rowcolor{C3!50} PointCMC~\cite{zhou2022pointcmc} & arXiv        & 2022 & Multi-modality   &RSCNN    &ShapeNet         &-     &-     &85.7  \\
\rowcolor{C3!50} P2P~\cite{wang2022p2p} & NeurIPS        & 2022 & Multi-modality   &CN-B-SFPN    &-       &-    &82.5     &85.7  \\
\rowcolor{C3!50} P2P~\cite{wang2022p2p} & NeurIPS        & 2022 & Multi-modality   &CN-L-UPer    &-       &-    &84.1     &86.5  \\
\rowcolor{C3!50} PointCLIP V2~\cite{zhu2022pointclip} & CVPR        & 2022 & Multi-modality   &ViT-B$\backslash$16 \& Transformer    &ImageNet         &10     &-     &48.4 (Zero-shot) \\
\rowcolor{C3!50} MD~\cite{sun2022self} & IEEE TMM & 2022 & Multi-modality & - & ShapeNet-Part & - & - & 85.5 \\
\rowcolor{C3!50} MD~\cite{sun2022self} & IEEE TMM & 2022 & Multi-modality & - & S3DIS & - & - & 85.4 \\
\bottomrule[1.5pt]
\end{tabular}%
}
\end{table*}
% !TEX root = ../bare_jrnl_new_sample4.tex
% \usepackage{graphicx}
\begin{table*}[htp]
\centering
\caption{Comparisons of \textbf{semantic segmentation} performance on \textbf{S3DIS} datasets}
\resizebox{\textwidth}{!}{%
\begin{tabular}{lcccccccc}
\toprule[1.5pt]
Methods & Publication & Years & Types of Methods     & Backbone            & Pre-train Datasets & Image Views & mACC & mIoU \\ \midrule[1pt]
\rowcolor{C6!50} RS~\cite{sauder2019self} & NeurIPS & 2019 & Reconstruction-based & DGCNN & - & - & 83.5 (Area 1) & 44.7 (Area 1) \\
\rowcolor{C6!50} RS~\cite{sauder2019self} & NeurIPS & 2019 & Reconstruction-based & DGCNN & - & - & 81.2 (Area 2) & 34.9 (Area 2) \\
\rowcolor{C6!50} RS~\cite{sauder2019self} & NeurIPS & 2019 & Reconstruction-based & DGCNN & - & - & 84.0 (Area 3) & 42.4 (Area 3) \\
\rowcolor{C6!50} RS~\cite{sauder2019self} & NeurIPS & 2019 & Reconstruction-based & DGCNN & - & - & 82.9 (Area 4) & 39.9 (Area 4) \\
\rowcolor{C6!50} RS~\cite{sauder2019self} & NeurIPS & 2019 & Reconstruction-based & DGCNN & - & - & 83.3 (Area 6) & 43.9 (Area 6) \\
\rowcolor{C6!50} OcCo~\cite{wang2021unsupervised} & ICCV & 2021 & Reconstruction-based & PointNet & ModelNet40 & - & - & 54.9 ± 1.0 \\
\rowcolor{C6!50} OcCo~\cite{wang2021unsupervised} & ICCV & 2021 & Reconstruction-based & PCN & ModelNet40 & - & - & 53.4 ± 2.1 \\
\rowcolor{C6!50} OcCo~\cite{wang2021unsupervised} & ICCV & 2021 & Reconstruction-based & DGCNN & ModelNet40 & - & - & 58.0 ± 1.7 \\
\rowcolor{C6!50} MaskSurf~\cite{zhang2022masked} & arXiv & 2022 & Reconstruction-based & Transformer (Transferring features protocol) & ShapeNet55 & 4 & 69.9 & 61.6 \\
\rowcolor{C6!50} MaskSurf~\cite{zhang2022masked} & arXiv & 2022 & Reconstruction-based & Transformer (Non-linear classification protocol) & ShapeNet55 & 4 & 66.6 & 56.6\\
\rowcolor{C6!50} ParAE~\cite{eckart2021self} & CVPR & 2021 & Reconstruction-based & DGCNN & - & - & 91.8 (Area 1) & 53.5 (Area 1) \\
\rowcolor{C6!50} ParAE~\cite{eckart2021self} & CVPR & 2021 & Reconstruction-based & DGCNN & - & - & 82.3 (Area 2) & 38.5 (Area 2) \\
\rowcolor{C6!50} ParAE~\cite{eckart2021self} & CVPR & 2021 & Reconstruction-based & DGCNN & - & - & 89.5 (Area 3) & 48.4 (Area 3) \\
\rowcolor{C6!50} ParAE~\cite{eckart2021self} & CVPR & 2021 & Reconstruction-based & DGCNN & - & - & 88.2 (Area 4) & 45.0 (Area 4) \\
\rowcolor{C6!50} ParAE~\cite{eckart2021self} & CVPR & 2021 & Reconstruction-based & DGCNN & - & - & 86.4 (Area 6) & 49.2 (Area 6)\\
\rowcolor{C6!50} IAE~\cite{yan2022implicit}     & arXiv      &2022       & Reconstruction-based     &PointNeXt      &ShapeNet    &-      &-      &75.3 (+0.4) \\

\rowcolor{C5!50} PointContrast~\cite{xie2020pointcontrast} & ECCV        & 2020  & Contrastive-learning-based & SR-UNet & ScanNet         & -       & 77.0     & 70.9 \\
\rowcolor{C5!50} CSC~\cite{hou2021exploring} & CVPR        & 2021 & Contrastive-learning-based & SR-UNet & ScanNet         & -       & -     & 72.2 \\
\rowcolor{C5!50} DVCo~\cite{hou2021exploring} & ECCV        & 2022 & \begin{tabular}[c]{@{}c@{}} Contrastive-learning-based\\ \& Multi-modality\end{tabular} & PointNet++ \& SR-UNet & ScanNet         & -       & -     & 67.2 \\
\rowcolor{C5!50} DCGLR~\cite{fu2022distillation} &arXiv     &2022	&Contrastive-learning-based  & 3D-ViT     & ShapeNet	        &-        &85.33 (Area 1)        &60.61 (Area 1) \\
\rowcolor{C5!50} DCGLR~\cite{fu2022distillation} & arXiv      &2022	&Contrastive-learning-based  & 3D-ViT     & ShapeNet	        &-        &73.11 (Area 2)        &40.01 (Area 2) \\
\rowcolor{C5!50} DCGLR~\cite{fu2022distillation} & arXiv      &2022	&Contrastive-learning-based  & 3D-ViT     & ShapeNet	        &-        &79.06 (Area 3)        &51.98 (Area 3) \\
\rowcolor{C5!50} DCGLR~\cite{fu2022distillation} & arXiv      &2022	&Contrastive-learning-based  & 3D-ViT     & ShapeNet	        &-        &79.06 (Area 3)        &51.98 (Area 3) \\
\rowcolor{C5!50} DCGLR~\cite{fu2022distillation} & arXiv      &2022	&Contrastive-learning-based  & 3D-ViT     & ShapeNet	        &-        &74.05 (Area 4)        &51.98 (Area 4) \\
\rowcolor{C5!50} DCGLR~\cite{fu2022distillation} & arXiv      &2022	&Contrastive-learning-based  & 3D-ViT     & ShapeNet	        &-        &78.24 (Area 5)        &50.22 (Area 5) \\

\rowcolor{C4!50} STRL~\cite{huang2021spatio} &ICCV        &2021    &Spatial-temporal-based   & DGCNN    & ScanNet      &-      &85.28 (Area 1)     &59.15 (Area 1) \\
\rowcolor{C4!50} STRL~\cite{huang2021spatio} &ICCV        &2021    &Spatial-temporal-based   & DGCNN    & ScanNet      &-      &72.37 (Area 2)     &39.21 (Area 2) \\
\rowcolor{C4!50} STRL~\cite{huang2021spatio} &ICCV        &2021    &Spatial-temporal-based   & DGCNN    & ScanNet      &-      &79.12 (Area 3)     &51.88 (Area 3) \\
\rowcolor{C4!50} STRL~\cite{huang2021spatio} &ICCV        &2021    &Spatial-temporal-based   & DGCNN    & ScanNet      &-      &73.81 (Area 4)     &39.28 (Area 4) \\
\rowcolor{C4!50} STRL~\cite{huang2021spatio} &ICCV        &2021    &Spatial-temporal-based   & DGCNN    & ScanNet      &-      &77.28 (Area 5)     &49.53 (Area 5) \\
\rowcolor{C4!50} PN~\cite{thabet2020self}   &CVPR   &2020   &Spatial-based      & PointNet      &-      &-      &63.1     &44.4 \\
\rowcolor{C4!50} RSNet~\cite{thabet2020self}   &CVPR   &2020   &Spatial-based      & PointNet      &-      &-      &61.2     &55.0 \\

\rowcolor{C3!50} Pix4Point~\cite{qian2022pix4point} & arXiv & 2022 & Multi-modality & PViT & ImageNet-1K & 15 & 69.9 & 64.4 \\
\rowcolor{C3!50} Pix4Point~\cite{qian2022pix4point} & arXiv & 2022 & Multi-modality & PViT + Pix4Point & ImageNet-1K & 15 & 75.2 & 69.6 \\
\rowcolor{C3!50} EPCL~\cite{huang2022frozen} & arXiv & 2022 & Multi-modality & CLIP image encoder & - & - & 84.1 & 72.6\\
\rowcolor{C3!50} MVR~\cite{tran2022self}      &ACCV      &2022       &Multi-modality     &DGCNN+ResNet50    &ModelNet40    &-      &87.0     &49.9  \\

\rowcolor{C3!50} MVR~\cite{tran2022self}      &ACCV      &2022       &Multi-modality     &SR-UNet+ResNet50    &ModelNet40    &-      &73.2      &66.0  \\

\rowcolor{C3!50} MVR~\cite{tran2022self}      &ACCV      &2022       &Multi-modality     &SR-UNet+ResNet50    &ScanNet    &-      &73.0      &66.5  \\

\rowcolor{C3!50} EPCL~\cite{huang2022frozen}    & -        & 2022  & Multi-modality & Transformer \& CLIP & -    & -    & 84.1     & 72.6 \\ 

\bottomrule[1.5pt]
\end{tabular}%
}
\end{table*}
% !TEX root = ../bare_jrnl_new_sample4.tex
% \usepackage{graphicx}
\begin{table*}[htp]\small
\centering
\caption{Comparisons of \textbf{indoor object detection} performance on \textbf{SUN RGB-D} and \textbf{ScanNet-V2} datasets}
\resizebox{\textwidth}{!}{%
\begin{tabular}{lcccccccccc}
\toprule[1.5pt]
Methods & Publication & Years & Types of Methods     & Backbone            & Pre-train Datasets & Image Views & \multicolumn{2}{c}{SUN RGB-D} & \multicolumn{2}{c}{ScanNet-V2} \\ 
        &             &       &                      &                     &                    &             & AP@25         & AP@50         & AP@25          & AP@50         \\ \midrule[1pt]
\rowcolor{C6!50} PointGLR~\cite{rao2022pointglr}     &IEEE TPAMI       &2022      &Reconstruction-based    &VoteNet      & ScanNet-V2       &-      &-      &-      &60.7     &35.6 \\
\rowcolor{C6!50} PointGLR~\cite{rao2022pointglr}     &IEEE TPAMI       &2022      &Reconstruction-based    &H3DNet      & ScanNet-V2        &-      &-      &-      &68.4    &51.2\\
\rowcolor{C6!50} IAE~\cite{yan2022implicit}     &arXiv      &2022       &Reconstruction-based     &VoteNet \& FCAF3D    &ScanNet     &-       &50.0(+1.1)     &65.0(+0.8)     &58.6(+1.3)     &72.5(+1.0) \\ 
\rowcolor{C6!50} UP3DETR~\cite{sun2022unsupervised}  &PRCV       &2022       &Reconstruction-based     &Transformer        &SUN RGB-D      &-      &56.6(+0.4)     &32.5(+2.8)     &63.1(+0.4)     &43.7(+6.2) \\

\rowcolor{C5!50} PointContrast~\cite{xie2020pointcontrast} & ECCV        & 2020  & Contrastive-learning-based & SR-UNet & ScanNet         & -       &57.5     &34.8      &59.2      &38.0\\
\rowcolor{C5!50} CSC~\cite{hou2021exploring} & CVPR        & 2021 & Contrastive-learning-based & SR-UNet & ScanNet         & -       &-     &36.4      &-      &39.3\\
\rowcolor{C5!50} DepthContrast~\cite{zhang2021self} & ICCV        & 2021 & Contrastive-learning-based & PointNet++ 3× & Redwood-vid \& ScanNet-vid         & -      &63.5      &43.4     &69.0     &50.0 \\
\rowcolor{C5!50} DPCo~\cite{li2022closer}    & ECCV       & 2022 & \begin{tabular}[c]{@{}c@{}} Contrastive-learning-based\\ \& Multi-modality\end{tabular}   & PointNet++ \& U-shaped 2D CNN    & ScanNet        & -      &59.8      &35.6     &64.2     &41.5 \\
\rowcolor{C5!50} 4DContrast~\cite{chen20224dcontrast}      &ECCV      &2022       &Contrastive-learning-based     &U-Net    &ModelNet40 \& ScanNet    &-      &-      &38.2   &-   &-  \\

\rowcolor{C4!50} STRL~\cite{huang2021spatio} &ICCV        &2021       &Spatial-based   & VoteNet    &ScanNet       &-      &58.2       &-      &-      &-\\
\rowcolor{C4!50} STRL~\cite{huang2021spatio} &ICCV        &2021       &Spatial-based   &  VoteNet    &ShapeNet      &-      &59.2       &-      &-      &-\\

\rowcolor{C4!50} SL3D~\cite{cendra2022sl3d} &NeurIPS        &2022       &Spatial-based      &PointNet++      &ScanNet      &-      &-      &-      & 18.6/4.6 (50 pseudo classes)(ScanNet)        &- \\
\rowcolor{C4!50} SL3D~\cite{cendra2022sl3d} &NeurIPS        &2022       &Spatial-based      &Transformer     &ScanNet      &-      &-      &-      &  17.8/7.6 (100 pseudo classes)(ScanNet)        &- \\
\rowcolor{C4!50} SL3D~\cite{cendra2022sl3d} &NeurIPS        &2022       &Spatial-based      &Transformer     &ScanNet      &-      &-      &-      &  20.3/7.9 (200 pseudo classes)(ScanNet)      &- \\
\rowcolor{C4!50} SL3D~\cite{cendra2022sl3d} &NeurIPS        &2022       &Spatial-based      &Transformer     &ScanNet      &-      &-      &-      &  19.1/9.3 (400 pseudo classes)(ScanNet)      &- \\

\rowcolor{C4!50} PC-FractalDB~\cite{yamada2022point}   &CVPR   &2022   &Spatial-based     &PointNet++     & ScanNet-V2   &      &59.4       &33.9       &61.9       &38.3\\
\rowcolor{C4!50} PC-FractalDB~\cite{yamada2022point}   &CVPR    &2022   &Spatial-based     &PointNet++ ×2     &ScanNet-V2      &-      &60.2       &35.2       &63.4       &39.9\\
\rowcolor{C4!50} PC-FractalDB~\cite{yamada2022point}   &CVPR    &2022   &Spatial-based      &SR-UNet     &ScanNet-V2      &-      &57.1       &35.9       &59.4       &37.0\\

\rowcolor{C3!50} MVR~\cite{tran2022self}      &ACCV      &2022       &Multi-modality     &SR-UNet \& ResNet50    &ModelNet40    &-      &58.1      &34.9   &58.4   &38.2  \\
\rowcolor{C3!50} MVR~\cite{tran2022self}      &ACCV      &2022       &Multi-modality     &SR-UNet \& ResNet50    &ScanNet    &-      &57.8      &35.1   &60.3   &39.2  \\
\rowcolor{C3!50} PointCLIP V2~\cite{zhu2022pointclip} & CVPR        & 2022 & Multi-modality   &3DETR-m \& Transformer    &-         &10    &-     &-    &18.97 (Zero-shot)     &11.53 (Zero-shot) \\

\rowcolor{C3!50} EPCL~\cite{huang2022frozen}    & arXiv           & 2022  & Multi-modality & Transformer \& CLIP & -                  & -           & -             & -             & -              & 43.0          \\

\bottomrule[1.5pt]
\end{tabular}%
}
\end{table*}
% !TEX root = ../bare_jrnl_new_sample4.tex
% \usepackage{graphicx}
\begin{table*}[htp]
\centering
\caption{Comparisons of \textbf{instance segmentation} performance on \textbf{S3DIS} and \textbf{ScanNet} datasets}
\resizebox{\textwidth}{!}{%
\begin{tabular}{lcccccccc}
\toprule[1.5pt]
Methods       & Publication & Years & Types of Methods           & Backbone & Pre-train Datasets & Image Views &S3DIS (mIoU)  & ScanNet (mIoU) \\    
\midrule[1pt]

\rowcolor{C6!50} PointGLR~\cite{xie2020pointcontrast} & IEEE TPAMI       & 2022  & Reconstruction-based & SR-UNet  & ScanNet           & -  & -           & 68.9    \\

\rowcolor{C4!50} PointContrast~\cite{xie2020pointcontrast} & ECCV        & 2020  & Contrastive-learning-based & SR-UNet  & ScanNet           & -  & -           & 55.8    \\ 
\rowcolor{C4!50} CSC~\cite{hou2021exploring} & CVPR        & 2021 & Contrastive-learning-based & SR-UNet & ScanNet         & -       & 63.4     & 59.4 \\
\rowcolor{C4!50} 4DContrast~\cite{chen20224dcontrast}      &ECCV      &2022       &Contrastive-learning-based     &U-Net    &ModelNet40 \& ScanNet    &-       & -    &57.6 \\

\rowcolor{C4!50} SL3D~\cite{cendra2022sl3d}      &NeuIPS      &2022       &Contrastive-learning-based     &PointNet++    & ScanNet    &-       & -    &60.2/32.9/5.8(50 pseudo classes)(Train/Val(SL3D)/Test) \\
\rowcolor{C4!50} SL3D~\cite{cendra2022sl3d}      &NeuIPS      &2022       &Contrastive-learning-based     &Transformer    & ScanNet    &-       & -    & 57.3/26.6/8.4(100 pseudo classes)(Train/Val(SL3D)/Test) \\
\rowcolor{C4!50} SL3D~\cite{cendra2022sl3d}      &NeuIPS      &2022       &Contrastive-learning-based     &PointNet++     & ScanNet    &-       & -    & 56.1/28.5/8.5(400 pseudo classes)(Train/Val(SL3D)/Test) \\\rowcolor{C4!50} SL3D~\cite{cendra2022sl3d}      &NeuIPS      &2022       &Contrastive-learning-based     &Transformer    & ScanNet    &-       & -    &  55.1/25.3/9.2(400 pseudo classes)(Train/Val(SL3D)/Test) \\\rowcolor{C4!50} SL3D~\cite{cendra2022sl3d}      &NeuIPS      &2022       &Contrastive-learning-based     &Transformer    & ScanNet    &-       & -    & 57.3/26.6/8.4(800 pseudo classes)(Train/Val(SL3D)/Test) \\

% \rowcolor{C4!50} PC-FractalDB~\cite{yamada2022point}   &CVPR    &2022   &Spatial-based     &PointNet++     &-      &-   & -           &60.8 (10 instances) \\
% \rowcolor{C4!50} PC-FractalDB~\cite{yamada2022point}   &CVPR    &2022   &Spatial-based    &PointNet++×2      &-      &-    & -         &60.6 (100 instances) \\
% \rowcolor{C4!50} PC-FractalDB~\cite{yamada2022point}   &CVPR    &2022   &Spatial-based     &SR-UNet      &-      &-       & -       &61.8 (1000 instances) \\

\rowcolor{C4!50} STRL~\cite{huang2021spatio}   &ICCV    &2021   &Spatial-temporal-based     &VoteNet      &ScanNet      &-       & 58.2       & - \\

\bottomrule[1.5pt]
\end{tabular}%
}
\end{table*}

% !TEX root = ../bare_jrnl_new_sample4.tex
% \usepackage{multirow}
% \usepackage{graphicx}
\begin{table*}[htp]
\centering
\caption{Comparisons of \textbf{3D object detection} performance on \textbf{Waymo validation} set.}
\resizebox{\textwidth}{!}{%
\begin{tabular}{lcccccc|cccc|cccc}
\toprule[1.5pt]
\multirow{2}{*}{Methods}          & \multirow{2}{*}{Publication} & \multirow{2}{*}{Year} & \multirow{2}{*}{Type of Methods}                                                       & \multirow{2}{*}{Backbone} & \multirow{2}{*}{\begin{tabular}[c]{@{}c@{}}Pre-train \\ Dataset\end{tabular}} & \multirow{2}{*}{\begin{tabular}[c]{@{}c@{}}Dataset\\ Fraction\end{tabular}} & \multicolumn{4}{c|}{L1(mAP/APH)}                                                           & \multicolumn{4}{c}{L2(mAP/APH)}                                                           \\ \cline{8-15} 
                                  &                              &                       &                                                                                        &                           &                                                                               &                                                                             & Overall              & Vehicle              & Pedestrian           & Cyclist               & Overall              & Vehicle              & Pedestrian           & Cyclist              \\ \midrule[1pt]

\rowcolor{C6!50} &           & &       & SECOND     &                 & 20$\%$                &-          & -           & -                    & -         & 60.86/57.15          & 64.38/63.85          & 60.10/50.84          & 58.11/56.76     \\\rowcolor{C6!50}  &               &                  &             & SECOND            &              & 100$\%$             & -                & -               & -           & -          & 61.03/57.30          & 64.42/63.87          & 59.97/50.65          & 58.69/57.39    \\\rowcolor{C6!50}   &              &            &                  & CenterPoint               &       & 20$\%$              & 70.00/67.19                 & -             & -             & -           & 66.70/64.25          & 64.71/64.22          & 66.21/60.59          & 69.11/67.93   \\\rowcolor{C6!50}   &               &                &                & CenterPoint            &                  & 100$\%$              & 73.50/70.90                & -              & -            & -               & 66.92/64.45          & 64.78/64.29          & 66.25/60.53          & 69.73/68.52  \\ \rowcolor{C6!50}  &               &                 &           & PV-RCNN++                 &                 & 20$\%$             & -              & -              & -           & -             & 70.45/67.96          & 69.44/69.02          & 71.14/65.21          & 70.77/69.65 \\\rowcolor{C6!50}  \multirow{-6}{*}{BEV-MAE~\cite{lin2022bev}}  &     \multirow{-6}{*}{arXiv}        &   \multirow{-6}{*}{2022}              &   \multirow{-6}{*}{Reconstruction-based}       & PV-RCNN++      &  \multirow{-6}{*}{Waymo}           & 100$\%$          & -              & -             & -            & -           & 70.54/68.11          & 69.53/69.07          & 71.50/65.69          & 70.60/69.56 \\ \hline

 \rowcolor{C6!50} &           & &  & SECOND                    &         & 20$\%$            & -                  & 71.12/70.58          & 67.21/55/68            & 57.73/56.18               &-             & 62.67/62.34          & 59.03/48.79            & 55.62/54.17 \\ \rowcolor{C6!50} 
&                    &                      &                  &  CenterPoint                &                  & 20$\%$           & -                   & 71.89/71.33                      & 73.85/67.12                 & 70.29/69.03                     & -          & 64.05/63.53          & 65.78/59.62         & 67.76/66.53 \\ \rowcolor{C6!50} 
&                    &                      &                  & PV-RCNN              &                  & 20$\%$           & -                   & 75.94/75.28                      &  74.02/63.48                 & 67.21/65.49                     & -          & 67.94/67.34          & 64.91/55.57         & 64.62/63.02 \\ \rowcolor{C6!50} \multirow{-4}{*}{Voxel-MAE~\cite{min2022voxel}}          
&   \multirow{-4}{*}{arXiv}                  &   \multirow{-4}{*}{2022}                    &    \multirow{-4}{*}{\begin{tabular}[c]{@{}c@{}}Reconstruction-based\\\end{tabular}}              &  PV-RCNN++              &      \multirow{-4}{*}{Waymo}             & 20$\%$           & -                   & 78.23/77.72                      &  79.85/73.23                 & 71.75/70.64                     & -          & 69.54/69.12          & 71.07/64.96         & 69.26/68.21 \\ \hline

\rowcolor{C6!50} &       &  &  & SECOND                    &          & 20$\%$            & -         & -         & -           & -              & 58.26/54.35        & 62.58/62.02         & 57.22/47.49       & 54.97/53.53 \\\rowcolor{C6!50} 
&                    &                      &                  &  SECOND+Voxel-MAE                &                  & 20$\%$           & -        & -        & -        & -           & 59.11/55/10      & 62.67/62.34       & 59.03/48.79       & 55.62/54.17 \\\rowcolor{C6!50} 
&                    &                      &                  &  SECOND+MAEli                &                  & 20$\%$           & -        & -        & -        & -           & 60.57/56.69        & 63.75/63.20         & 60.71/50.93       & 57.26/55.95  \\\rowcolor{C6!50} 
&                    &                      &                  &  CenterPoint                &                  & 20$\%$           & -        & -        & -        & -           & 64.51/61.92       &   63.16/62.65    & 64.27/58.23       & 66.11/64.87 \\\rowcolor{C6!50} 
&                    &                      &                  &  CenterPoint+Voxel-MAE                &                  & 20$\%$           & -        & -        & -        & -           & 65.86/63.23   & 64.05/63.53   & 65.78/59.62       & 67.76/66.53 \\\rowcolor{C6!50} 
&                    &                      &                  &  CenterPoint+MAEli                &                  & 20$\%$           & -        & -        & -        & -           & 65.60/63.00  & 64.22/63.70  & 65.93/59.79       & 66.66/65.52 \\\rowcolor{C6!50} 
&                    &                      &                  &  PV-RCNN                &                  & 20$\%$           & -        & -        & -        & -           & 64.84/60.86       & 67.44/66.80      & 63.70/53.95      &63.39/61.82 \\ \rowcolor{C6!50} 
&                    &                      &                  &  PV-RCNN+Voxel-MAE               &                  & 20$\%$           & -        & -        & -        & -           & 65.82/61.98      & 67.94/67.34      & 64.91/55.57      & 64.13/62.79 \\ \rowcolor{C6!50} \multirow{-9}{*}{MAELi-MAE~\cite{krispel2022maeli}}           
&   \multirow{-9}{*}{arXiv}                      &    \multirow{-9}{*}{2022}                  &    \multirow{-9}{*}{\begin{tabular}[c]{@{}c@{}}Reconstruction-based\\\end{tabular}}              &  PV-RCNN+MAEli               &   \multirow{-9}{*}{Waymo}               & 20$\%$           & -        & -        & -        & -        &65.72/62.15 &67.90/67.34 &65.14/56.32 &64.13/62.79 \\ \hline

\rowcolor{C6!50} &     &  & & GD-MAE         &       & 20$\%$      &-  &76.24/75.74    & 80.50/72.29  & 72.63/71.42   &70.24/67.14    &67.67/67.22    & 73.18/65.50   &69.87/68.71\\\rowcolor{C6!50} 
&                    &                      &                  &  GD-MAE                &                  & 100$\%$           & -     &77.26/76.78   &80.26/72.36    &73.12/71.94    &70.62/67.64    &68.72/68.29    &72.84/65.47    & 70.30/69.16 \\\rowcolor{C6!50} 
\multirow{-3}{*}{GD-MAE~\cite{yang2022gd}}          &   \multirow{-3}{*}{arXiv}                        &   \multirow{3}{*}{2022}                   &   \multirow{-3}{*}{\begin{tabular}[c]{@{}c@{}}Reconstruction-based\\\end{tabular}}                & GD-MAE with an extra IoU prediction head       &  \multirow{-3}{*}{Waymo}      & 100$\%$           & -               & 79.40/78.94  &82.20/75.85    &75.75/74.77   &72.90/70.43    & 70.91/70.49   &74.82/68.79    &72.98/72.03 \\\hline

\rowcolor{C5!50} &       & &  & SECOND                    &                                                        & 100$\%$                                                                       & -                    & -                    & -                    & -                     & 60.90/57.17          & 64.50/63.90          & 60.33/51.00          & 57.90/56.60          \\\rowcolor{C5!50} 
                                  &                              &                       &                                                                                        & CenterPoint               &                                                                               & 100$\%$                                                                       & -                    & -                    & -                    & -                     & 66.67/64.20          & 65.22/64.80          & 66.40/60.49          & 68.48/67.38          \\\rowcolor{C5!50} 
\multirow{-3}{*}{ProposalContrast~\cite{yin2022proposalcontrast}}                                   &    \multirow{-3}{*}{ECCV}                            &   \multirow{-3}{*}{2022}                     &     \multirow{-3}{*}{\begin{tabular}[c]{@{}c@{}}Contrastive-learning\\ -based\end{tabular}}                                                                                   & PV-RCNN++                 &                                    \multirow{-3}{*}{Waymo}                                            & 100$\%$                                                                       & -                    & -                    & -                    & -                     & 70.49/67.98          & 69.47/68.95          & 71.28/65.31          & 70.73/69.59          \\ \hline
                                  
\rowcolor{C5!50} GCC-3D~\cite{liang2021exploring}                            & ICCV                         & 2021                  & \begin{tabular}[c]{@{}c@{}}Contrastive-learning\\ -based\end{tabular}                  & CenterPoint               & Waymo                                                                         & 100$\%$                                                                       & -                    & -                    & -                    & -                     & 65.29/62.79          & 63.97/63.47          & 64.23/58.47          & 67.68/66.44          \\ \hline
			
\rowcolor{C5!50} SimIPU~\cite{li2022simipu}                           & AAAI                         & 2022                  & \begin{tabular}[c]{@{}c@{}}Contrastive-learning\\ -based\end{tabular}                  & ResNet-50               & KITTI                                                                        & -                                                                       & 66.92/63.18                    & 66.50/66.10                    & 69.40/60.50                    & 64.70/62.30                     & 63.01/59.47          & 62.40/62.00          & 64.70/56.30          & 62.30/60.00         \\

\bottomrule[1.5pt]
\end{tabular}%
}
\end{table*}
% !TEX root = ../bare_jrnl_new_sample4.tex
% \usepackage{multirow}
% \usepackage{graphicx}
\begin{table*}[htp]
\centering
\caption{Comparisons of \textbf{3D object detection} performance on \textbf{ONCE validation} set.}
\resizebox{\textwidth}{!}{%
\begin{tabular}{lcccccc|cccc}
\toprule[1.5pt]
\multirow{2}{*}{Methods} & \multirow{2}{*}{Publication} & \multirow{2}{*}{Year} & \multirow{2}{*}{Type of Methods}                                      & \multirow{2}{*}{Backbone} & \multirow{2}{*}{\begin{tabular}[c]{@{}c@{}}Pre-train \\ Dataset\end{tabular}} & \multirow{2}{*}{\begin{tabular}[c]{@{}c@{}}Dataset\\ Fraction\end{tabular}} & \multirow{2}{*}{mAP} & \multicolumn{3}{c}{AP}         \\ \cline{9-11} 
                         &                              &                       &                                                                       &                           &                                                                               &                                                                             &                      & Vehicle & Pedestrain & Cyclist \\ \midrule[1pt]

\rowcolor{C6!50} ALSO~\cite{boulch2022also}         & arXiv          & 2022        & Reconstruction-based & SECOND          & KITTI          & 100$\%$         & 52.68     & 71.73   & 28.16      & 58.13   \\ \hline

\rowcolor{C6!50} GD-MAE~\cite{yang2022gd}         & arXiv         & 2022        & Reconstruction-based & SECOND           & KITTI        & 100$\%$       & 64.92         & 76.79       &48.84       &69.14    \\ \hline

\rowcolor{C6!50} Voxel-MAE~\cite{min2022voxel}       &arXiv      &2022       &Reconstruction-based      &SECOND          &unlabeled small set (100k scenes)          &-         &52.51      &72.78      &27.49      &57.26\\ \hline

\rowcolor{C5!50} 
ProposalContrast~\cite{yin2022proposalcontrast}      & ECCV          & 2022          & \begin{tabular}[c]{@{}c@{}}Contrastive-learning\\ -based\end{tabular} & Centerpoint       & Waymo         & 100$\%$           & 66.24       & 78.00   & 52.56      & 68.17   \\ \hline

\rowcolor{C5!50} &          &       &  &       &         &        &  & 84.62(0-30m) &33.64 &68.22 \\\rowcolor{C5!50}  & & & & & & & &67.11(30-50m) &28.00 &52.89  \\\rowcolor{C5!50} & & &  & \multirow{-3}{*}{SECOND}  & & & \multirow{-3}{*}{53.28}&49.42($>$50m) &17.61 &32.92 \\\cline{8-11} \rowcolor{C5!50}  & & & & & & &&87.85 &32.75 &71.22 \\ \rowcolor{C5!50}  & & & & & & & &71.79 &26.57 &52.50 \\ \rowcolor{C5!50}  & & & & \multirow{-3}{*}{PV-RCNN} &  & &  \multirow{-3}{*}{55.17}  &57.46 &17.29 &36.20  \\\cline{8-11} \rowcolor{C5!50}  & & & & & & & &78.02 &55.09 &74.17 \\ \rowcolor{C5!50}  & & & &  & & &  &56.13 &42.34 &56.05 \\ \rowcolor{C5!50} 
\multirow{-9}{*}{CO\^{}3~\cite{chen2022co}} & \multirow{-9}{*}{arXiv}  & \multirow{-9}{*}{2022}     & \multirow{-9}{*}{\begin{tabular}[c]{@{}c@{}}Contrastive-learning\\ -based\end{tabular}} & \multirow{-3}{*}{CenterPoint} & \multirow{-9}{*}{DAIR-V2X}  &  \multirow{-9}{*}{100$\%$}    & \multirow{-3}{*}{58.50} &39.94 &27.44 &38.16       \\

\bottomrule[1.5pt]
\end{tabular}%
}
\end{table*}
% !TEX root = ../bare_jrnl_new_sample4.tex
% \usepackage{multirow}
% \usepackage{graphicx}
\begin{table*}[htp]
\centering
\caption{Comparisons of \textbf{3D object detection} performance on \textbf{nuScenes validation} set.}
\resizebox{\textwidth}{!}{%
\begin{tabular}{lcccccc|cc}
\toprule[1.5pt]
\multirow{2}{*}{Methods} & \multirow{2}{*}{Publication} & \multirow{2}{*}{Year} & \multirow{2}{*}{Type of Methods}                                      & \multirow{2}{*}{Backbone} & \multirow{2}{*}{\begin{tabular}[c]{@{}c@{}}Pre-train \\ Dataset\end{tabular}} & \multirow{2}{*}{\begin{tabular}[c]{@{}c@{}}Dataset Fraction\\ of Fine-tuning Dataset \end{tabular}} & \multirow{2}{*}{mAP} 
                         & \multirow{2}{*}{NDS}     \\ & & & & & & & & \\\midrule[1pt]
\rowcolor{C6!50}       &          &          &  &   &          & 20$\%$           & 47.35       & 59.06      \\ 
\rowcolor{C6!50}       &          &          &  &   &          & 40$\%$           & 50.02       & 61.01      \\ 
\rowcolor{C6!50}       &          &          &  &   &          & 60$\%$           & 51.00       & 61.76      \\ 
\rowcolor{C6!50}       &          &          &  &   &          & 80$\%$           & 51.67       & 62.38      \\ 
\rowcolor{C6!50} \multirow{-5}{*}{Voxel-MAE~\cite{hess2022masked}}      & \multirow{-5}{*}{arXiv}          & \multirow{-5}{*}{2022}          & \multirow{-5}{*}{{\begin{tabular}[c]{@{}c@{}}Contrastive-learning \\-based \end{tabular}}} & \multirow{-5}{*}{Single-stride Sparse Transformer (SST)}       & \multirow{-5}{*}{unlabeled nuScenes}         & 100$\%$           & 51.95       & 62.16      \\ \hline

\rowcolor{C5!50} GCC-3D~\cite{liang2021exploring}      & ICCV          & 2021          & {\begin{tabular}[c]{@{}c@{}}Contrastive-learning \\-based \end{tabular}} & CenterPoint with VoxelNet       & Waymo         & 100$\%$           & 57.26       & 65.01      \\ \hline 

\rowcolor{C2!50} &           &           &  & PointPillars       &          &            & 42.06       & 55.02 \\  \rowcolor{C2!50} 
\multirow{-2}{*}{Self-supervised Scene Flow~\cite{erccelik20223d}} & \multirow{-2}{*}{arXiv} & \multirow{-2}{*}{2022} & \multirow{-2}{*}{Flow-based} & CenterPoint & \multirow{-2}{*}{KITTI} & \multirow{-2}{*}{100$\%$} & 49.94 & 60.06      \\

\bottomrule[1.5pt]
\end{tabular}%
}
\end{table*}
% !TEX root = ../bare_jrnl_new_sample4.tex
% \usepackage{multirow}
% \usepackage{graphicx}
\begin{table*}[t]
\centering
\caption{Comparisons of \textbf{3D object detection} performance on \textbf{nuScenes test} set.}
\resizebox{\textwidth}{!}{%
\begin{tabular}{lcccccc|cc}
\toprule[1.5pt]
\multirow{2}{*}{Methods} & \multirow{2}{*}{Publication} & \multirow{2}{*}{Year} & \multirow{2}{*}{Type of Methods}                                      & \multirow{2}{*}{Backbone} & \multirow{2}{*}{\begin{tabular}[c]{@{}c@{}}Pre-train \\ Dataset\end{tabular}} & \multirow{2}{*}{\begin{tabular}[c]{@{}c@{}}Dataset\\ Fraction\end{tabular}} & \multirow{2}{*}{mAP} 
                         & \multirow{2}{*}{NDS}     \\ & & & & & & & & \\\midrule[1pt]
                         
\rowcolor{C3!50} S2M2-SSD~\cite{zheng2022boosting}      & CVPR          & 2022          & Multi-modality & CenterPoint       & nuScenes/nuImages         & 100$\%$           & 62.90       & 69.30      \\ \hline 

\rowcolor{C2!50} &           &        &  & PointPillars       &          &          & 43.63       & 56.28 \\ \rowcolor{C2!50}    \multirow{-2}{*}{Self-supervised Scene Flow~\cite{erccelik20223d}} & \multirow{-2}{*}{arXiv} & \multirow{-2}{*}{2022}  & \multirow{-2}{*}{Flow-based} & CenterPoint & \multirow{-2}{*}{KITTI} & \multirow{-2}{*}{100$\%$}   & 51.42 & 60.92      \\

\bottomrule[1.5pt]
\end{tabular}%
}
\end{table*}


% \end{IEEEbiography}


%\newpage



%\vfill


\end{document}